\documentclass[12pt]{book}

\usepackage[english]{babel}
\usepackage{cmbright}

\usepackage[a4paper, right=30mm, left=30mm, top=40mm, bottom=40mm]{geometry}

\usepackage{booktabs, tabularx, wrapfig, subcaption,  bibentry, natbib, graphicx, amsmath}
\usepackage{algorithm}
\usepackage[noend]{algorithmic}

\usepackage{hyperref}
\hypersetup{
	colorlinks = true,
	linkcolor = purple,
	urlcolor  = blue,
	citecolor = blue,
	anchorcolor = red]{hyperref}
}
\usepackage[dvipsnames,table]{xcolor}
\usepackage{multirow}
\title{Dissecting continual learning: a structural and data analysis}	% INSERT HERE THE TITLE OF YOUR THESIS
\author{Francesco Pelosin}	% INSERT HERE YOUR NAME

\usepackage{fancyhdr}

\usepackage{amsmath,amsfonts,bm}

\def\eqref#1{equation~\ref{#1}}
\def\1{\bm{1}}

\def\vx{{\bm{x}}}

\DeclareMathAlphabet{\mathsfit}{\encodingdefault}{\sfdefault}{m}{sl}
\SetMathAlphabet{\mathsfit}{bold}{\encodingdefault}{\sfdefault}{bx}{n}

\newcommand{\R}{\mathbb{R}}

\newlength{\myMheight}
\definecolor{mypurple}{HTML}{d000ff}
\begin{document}
\begin{titlepage}
    \begin{center}
    	\centering
        \includegraphics[width=3cm]{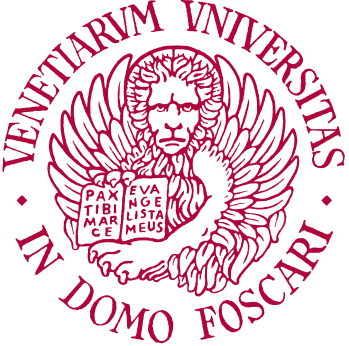}\\
        \vspace{1cm}
        \Large
        Ca' Foscari University \\
        
        Department of Environmental Sciences,\\ Informatics and  Statistics\\
        
        \vspace{1.5cm}
 
        %\framebox{
        \Huge       
        \textbf{\textcolor{black}{Dissecting Continual Learning}}\\
        \Large
        \textbf{a Structural and Data Analysis}
        %}
        
        \vspace{1.5cm}
        \Large
        Ph.D. Thesis - Computer Science\\XXXIV Cycle

        \vspace{2cm}
        \large
        Submitted by:   \\
        \textcolor{blue}{Pelosin Francesco}
        
        \vspace{1cm}
        \large
        Supervisor:  \\
        \textcolor{blue}{Prof. Torsello Andrea}   \\

        \vspace{1cm}
        \large
        Venice, Italy - \textcolor{black}{March, 2022}\\

    \end{center}
\end{titlepage}

\cleardoublepage

\begin{center}
\section*{Abstract}
\end{center}
\vspace{-1.8em}

Deep Learning aims to discover how artificial neural networks learn the rich internal representations required for difficult tasks such as recognizing objects or understanding language. This hard question is still unanswered although we are constantly improving the performance of such systems spanning from computer vision problems to natural language processing tasks. Continual Learning (CL) is a field dedicated in devising algorithms able to achieve lifelong learning by overcoming the knowledge disruption of previously acquired concepts, a phenomenon that affects deep learning architectures and that goes by the name of catastrophic forgetting. Currently, deep learning methods can achieve outstanding results when the data modeled does not undergo a considerable distribution shift in subsequent learning sessions, but as we expose the systems to such incremental setting, performance abruptly drops due to catastrophic forgetting. As the data generated in the world is continuously increasing, the demand to model such streams in a sequential fashion is increasing. As such, devising techniques to prevent knowledge corruption in neural networks is fundamental. Overcoming such limitations would allow us to build truly intelligent systems showing adaptability and human-like quality. Secondly, it would allow us to overcome the limitation, and onerous aspect, of retraining the architectures from scratch with the updated data. Such drawback comes from how deep neural networks learn, that is, they require several parameter updates to learn any given concept. This is also the exact reason why catastrophic forgetting happens, as we learn new concepts we overwrite old ones, while a truly intelligent system would show a stability-plasticity optimal trade-off. In this thesis, we first describe the background needed to understand continual learning in the computer vision realm. We do so with the introduction of a notation and a formal description of the problem. Then, we will introduce several CL setting variants and main solution categories proposed in the literature, along with an analysis of the state-of-the-art. We then first analyze one of the baseline approaches to continual learning and discover that in rehearsal-based techniques the quantity of data stored is a more important factor than the quality of memorized data. This trade-off surprisingly holds even for impressively high compression rates of the data. Secondly, this thesis proposes one of the early works on the study of incremental learning on vision transformer architectures (ViTs). In particular, we will compare functional, weight, and attention regularization approaches for the challenging rehearsal-free CL. We then propose an asymmetric loss variant inspired by PODNet, achieving good capabilities in terms of plasticity. Among these contributions, we propose a simple, but effective baseline for off-the-shelf continual learning exploiting pretrained models and discuss its extension to unsupervised continual learning, a topic that deserves further attention from the community. As the final work, we introduce a novel algorithm able to explore the environment through unsupervised visual pattern discovery. We then provide a conclusion and discuss further developments and promising paths to be followed by the CL research.
%\subsection*{Short-Abstract}

%Continual Learning (CL) is a field dedicated to devise algorithms able to achieve lifelong learning. Overcoming the knowledge disruption of previously acquired concepts, a drawback affecting deep learning models and that goes by the name of catastrophic forgetting, is a hard challenge. Currently, deep learning methods can attain impressive results when the data modeled does not undergo a considerable distributional shift in subsequent learning sessions, but whenever we expose such systems to this incremental setting, performance drop very quickly. Overcoming this limitation is fundamental as it would allow us to build truly intelligent systems showing stability and plasticity. Secondly, it would allow us to overcome the onerous limitation of retraining these architectures from scratch with the new updated data. To this end, we first analyze onetables of the most efficient baselines approaches and show that in rehearsal-based techniques the quantity of data stored in the rehearsal buffer is a more important factor over the quality. Secondly we propose one of the early works on the study of incremental learning on ViTs architectures, comparing functional, weight and attention regularization approaches. We then propose effective novel techniques based on a variation of PODNet-like loss along with a new asymmetric loss. 

\cleardoublepage

\begin{center}
\section*{Akwnowledgments}
First, I would like to express my gratitude to my supervisor Andrea Torsello, for all the deep insights and for welcoming me to pursue this research with him. 

Secondly, I would like to thank all the people that I encountered throughout these years, especially colleagues and friends that I met, a personal acknowledgment to Alessandro, Seyum, Fatima, and also the friends I met in Spain Hectór, Laura, and Albin. I would also like to say thank everyone that loved me during this period, you gave me the strength to carry on this tough journey!

Lastly, I would say that I learned a lot during these years, and the force that moved me to pursue a Ph.D., is the same force that allows us to expand and look for answers, to find meanings, and to unfold \textit{something beautiful}.

\texttt{``You're pretty good''} 
\end{center}
\clearpage

\tableofcontents
%   List up brief explanation of figures and tables.
\cleardoublepage
%\listoffigures
%\listoftables
%\clearpage
\pagestyle{fancy}

\chapter{Introduction}

\begin{center}
	\textit{``The measure of intelligence is the ability to change''\\ \hfill - Albert Einstein}
\end{center}
The interconnections among entities in our world are growing. Along with this fact, the ability to keep track and record such data has accordingly increased. The need for systems that can cope with such phenomena is essential. Deep Learning (DL) revealed itself to be a powerful weapon to model such complex streams, especially in Computer Vision and Natural Language Processing fields. The advent of DL unlocked the ability to develop outstanding technologies that can directly impact our lives. Self-driving cars are one example. Unfortunately not always the impact is positive, if not properly controlled. Therefore, the need for systems that show generalization abilities and can cope with unexpected scenarios, is nowadays essential. To this end, we also need responsive machines, that can be trained to quickly learn new concepts with low resource consumption. In fact, what happens if the stream of data encountered by a deep learning model changes its quality over time? This particular question is tackled by Continual Learning (CL) whose aim is to develop lifelong learning machines, unlocking fast adaptability to new environments. 

Modern deep learning methods for computer vision adapt themselves only to the manifold they are trained on. Instead, we need to devise models which are plastic enough to generalize to distributional shifts in the data and do not require complete retraining. This challenge would be solved if training Deep Learning models would not be such a delicate process affected by unexpected drawbacks. In fact, when we introduce the notion of learning through time and expose the system to face incremental tasks of different nature, things can get really complicated. 

One of the drawbacks of incrementally learning is the so-called catastrophic forgetting, where the system is subject to an abrupt deterioration of past knowledge whenever asked to learn new concepts. This big limitation is broadly studied in continual learning. To approach this delicate subject, in this thesis, we start by gently introducing some basic differences between artificial and natural intelligence. Here, we clarify some operative differences between artificial neural networks and some basic brain mechanisms arising from neuroscience. Then, we informally introduce the notion of continual learning and discuss the stability-plasticity dilemma along with the phenomenon of catastrophic forgetting of artificial neural networks. We proceed by introducing a more formal definition of incremental learning along with its fine-grained inclinations. Before moving to the contributions we introduce a brief overview of the state-of-the-art and define the main baselines which act as lower and upper bounds for continual learning methodologies. 

We step into the major contributions by focusing on rehearsal systems, a family of methods that exploit cache memories to replay previous knowledge. Here, we study how the compression of stored rehearsal data impacts the performance of the model. Tackling the memory side of CL, we provide a quality/quantity analysis through the usage of several compression schemes. We consider also extreme compression rates, providing some insights. On top of that, we consider continual learning under low-resource constraints through the usage of random projections and, in particular, Extreme Learning Machines.

To follow, as a second major contribution, we are among the first to investigate Vision Transformers in continual learning. In particular, we analyze several regularization schemes for ViTs, providing a first envision of rehearsal-free CL. We consider weight, functional and attentional regularizations, being the latter unexplored before, we carefully study the application of regularizations to specific parts of the self-attention mechanism. As a side contribution we introduce a new asymmetric loss variant inspired by a contemporary continual learning method (PODNet) principled by the observation that new attention should not penalize the acquisition of new knowledge. 

We then further clarify the usage of pretrained models in continual learning through an experimental segment. We compare fully pretrained CNNs and Vision Transformers in several incremental benchmarks. We provide a clear simple baseline that requires few KBytes to operate and does not perform parameter updates. Being simple and effective, we discuss its extension to the unsupervised realm. Here we consider further extensions for future works.

Along with these three contributions, we also study the ability of a system to autonomously discover new visual patterns, a notion embedded in an optimal incremental learner. We, therefore, provide a simple unsupervised pipeline able to discover semantic patterns on different visual scales. Finally, we conclude by wrapping up our perspectives on the main aforementioned challenges.

As a final note, we hope this thesis finds a meaningful purpose in the CL community, contributing to the development of Continual Learning and Computer Vision research.

\clearpage
\section*{Contibution Prefaction}

In this thesis we included some papers developed while pursuing the Ph.D.. The main contributions have been reported in Chapter \ref{chp:analyses}. The chapter holds the outcome of several collaborations and with the following list we report the names of the authors and the venues where the works have been submitted:

\begin{itemize}
	\item The work reported in Section \ref{wrk:smaller_is_better}, has been accepted as oral poster to \textit{IJCNN 2022}. The authors who contributed to the work are (in order): Francesco Pelosin and Andrea Torsello from Ca' Foscari University of Venice
	
	\item The work reported in Section \ref{wrk:cl_vit} is the outcome of the collaboration of the research period abroad and has been accepted as poster to the \textit{Continual Lerning Workshop of CVPR 2022}. The authors who contributed to the work are (in order): Francesco Pelosin, Ca' Foscari University of Venice (Equal Contrib); Saurav Jha, University of New South Wales, Australia (Equal Contrib); Andrea Torsello, Ca' Foscari University of Venice, Italy; Bogdan Raducanu and Joost van de Weijer from Computer Vision Center, Spain. 
	
	\item The work reported in Section \ref{wrk:off-the-shelf} has been accepted as poster to the \textit{Transformers for Vision Workshop of CVPR 2022} and it is single authored by Francesco Pelosin.
	
	\item The work reported in Section \ref{wrk:unsup}, has been accepted to the \textit{S+SSPR 2020}. The authors who contributed to the work are (in order): Francesco Pelosin; Andrea Gasparetto; Andrea Albarelli and  Andrea Torsello, Ca Foscari University of Venice, Italy
	
\end{itemize}

\chapter{Background and Motivation}
\section{Artificial vs Natural Intelligence}

Although the recent developments and great achievements of the field of Artificial Intelligence, the fundamental nature of Artificial Neural Networks (ANNs) might still be a coarse approximation of how our biological brains work. With the mathematical introduction by \citep{mcculloch1943logical} and the introduction of the ``Perceptron'' by \citep{rosenblatt1958perceptron}, which constitutes the smallest unit that form a ANN, we shaped our modeling of intelligence. An artificial neuron can be described as a cumulative summation of multiplications over some weights followed by a non-linear function.

Then, after the introduction of the famous Multy Layer Perceptrons (MLPs) the structure of ANNs has not changed much: we work in a connectionist paradigm where the learning happens through a distributed signal activity via connections among artificial neurons. In particular, the learning occurs by modifying connection strengths based on experience, this modification procedure has a particular name and it is the so-called \textit{backpropagation} algorithm whose discovery can be traced back to \citep{Rumelhart1986LearningRB} but with some earlier works by \citep{Linnainmaa1976TaylorEO} (as an M.Sc. Thesis) as pointed in \citep{DBLP:journals/corr/Schmidhuber14}.

The success of \textbf{connectionists models} span over different fields: Convolutional Neural Networks (CNN) for Computer Vision (CV) \citep{resnet18}, Language Models for Natural Language Processing (NLP) \citep{DBLP:conf/naacl/DevlinCLT19}, Deep Q-Learning Networks (DQN) for Reinforcement Learning \citep{DBLP:conf/icml/AgarwalS020}, Generative Audio Models for Audio \citep{DBLP:conf/ssw/OordDZSVGKSK16} and Graph Convolutional Networks (GCN) for graph data \citep{DBLP:conf/iclr/KipfW17}.

Connectionist models are a composition of several layers of artificial neurons, followed by a non-linearity. There are several types of layers each with its peculiarity. For example with the introduction of Batch Normalization \citep{DBLP:conf/icml/IoffeS15} we allowed the networks to achieve faster training. The introduction of some specialized units often allowed to excel in particular fields such as the convolutional operation \citep{mnist} for Computer Vision tasks and the Self-Attention mechanism in Natural Language Processing. \citep{DBLP:conf/nips/VaswaniSPUJGKP17}, although the attention mechanism has achieved tremendous achievements in vision tasks thanks to \citep{vit} and its introduction of Visual Transformers. Nowadays there is still no perfect mechanism/model for each scenario because we are still in the process of discovering how learning happens. For sure in the future, we might see other methodologies working in fields where they are not born from.

\begin{figure}[t]
    \centering
    \includegraphics[width=\textwidth]{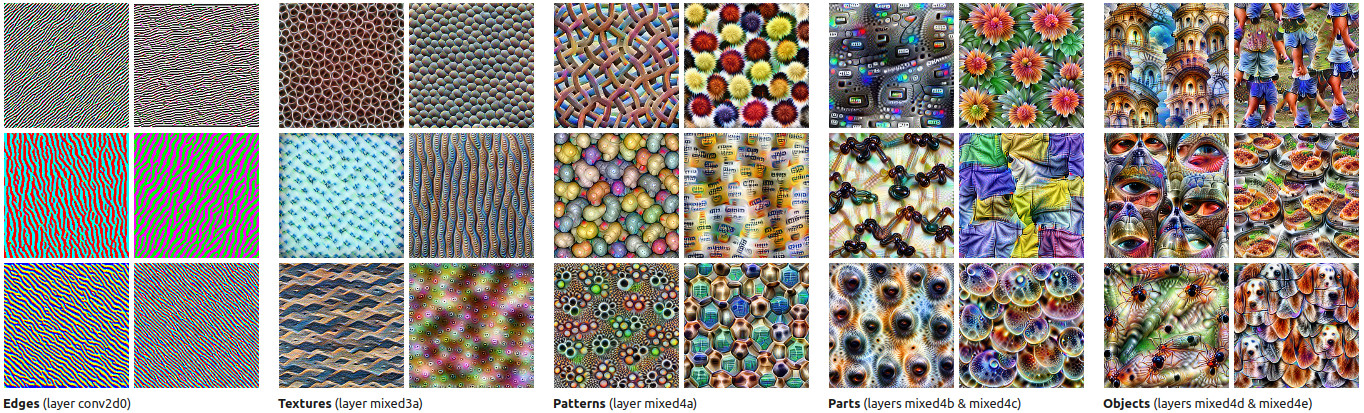}
    \caption{Feature visualization of GoogLeNet \citep{DBLP:conf/cvpr/SzegedyLJSRAEVR15}, trained on the ImageNet \citep{DBLP:journals/ijcv/RussakovskyDSKS15} dataset. Concepts in early layers are reported on left while concepts of last layers are on the right. The image is taken from \citep{olah2017feature}}
    \label{fig:01_features}
\end{figure}

While attention-based models spread the knowledge, and feature representations, uniformly across the layers \citep{Raghu2021DoVT}, in classical convolution-based models (such as ResNets \citep{resnet18}) the knowledge is constructed in a bottom-up fashion. This is a well-known fact. In particular, abstract concepts are always the result of the composition of simpler concepts. For example in early layers of CNNs for CV tasks, each neuron specializes in the detection of low-level features, while, as we move towards the head, the network learns patterns with more semantic relevance for us humans. This can be seen thanks to the beautiful visualization of \citep{olah2017feature} captured in Figure~\ref{fig:01_features}. This also reflects some neuroscientific discoveries where hierarchies of more and more abstract concepts have been demonstrated repeatedly, especially in the visual brain areas \citep{Riesenhuber1999HierarchicalMO}.

%hanks to the fact that in the past neuroscience adopted the interpretation that in the brain there are different areas dedicated to different fields, nowadays neuroscientific evidence underline the need of a more holistic model to understand our brain \todo{requires ref}.

While those resemblances are appealing to draw a connection between artificial and biological brains, the difference is still striking. For example, quite often Deep Learning models are static, that is, they are not altering their architecture over time but, in our biological brains, new connections can appear, while others can also cease to exist. This is also the so-called \textit{neuroplasticity} of our brains, whose first scientific evidence has been reported by \citep{Bennett1964ChemicalAA}. As we will see, continual learning and few other fields (e.g., dynamic routing, conditional computation, etc.) are the only ones going in this direction.

On another note, \textbf{time} seems to be a major factor in both artificial and natural learning. Our current connectionist framework does not exploit the notion of time in learning. To accommodate such a factor we would need to redefine the current learning framework because so far the models process data but without being conditioned to \textit{when} something is learned. There have been some attempts towards this direction by defining the learning as a system of differential equations taking into consideration time as a fundamental variable and also some attempts to implement it by \citep{DBLP:conf/ijcnn/BettiGMM20}, although the majority of the works still operate in the classical scenario.

Another clear distinction between artificial and biological neurons lies in how they decide to fire. The artificial neuron receives inputs and multiplies them by some weights that are adapted during learning. To fire, it uses an activation function (such as ReLu \citep{Agarap2018DeepLU}), but the reality of biological neurons is different. Each biological neuron has its threshold resultant from a complex chemical interaction. A class of models that are trying to bridge this gap is Spiking Neural Networks \citep{MAASS19971659} where the firing of the neuron is determined by a threshold on the signal received. Note that also this simplified model mimics neither the creation nor the destruction of connections (dendrites or axons) between neurons, and ignores signal timing. However, this restricted model alone is powerful enough to work with simple classification tasks. 

Another important difference is that biological circuits contain a myriad of additional details and complexity not translated to DL models, including diverse neural cell types \citep{Tasic2018SharedAD} with some recent attempts by \citep{Doty2021HeterogeneousT} to bridge this gap by changing the activation function for each artificial neuron. Another attempt to introduce more complex structures has been proposed by \citep{capsnet} with the introduction of Capsule Net models, a family of networks where the neurons are structured in hierarchies.

The most widely known neuroscientific framework for the brain is the Complementary Learning Systems (CLS) \citep{McClelland1995WhyTA}. This framework explains why the brain requires two deferentially specialized learning and memory systems, and it nicely specifies their central properties i.e., the \textit{hippocampus} as a sparse, pattern-separated system for rapidly learning episodic memories, and the \textit{neocortex} as a distributed, overlapping system that gradually integrates experienced episodes and extracts latent semantic structures. Instead, most of the proposed artificial models, are more of a well-engineered pipeline crafted to excel in a particular task such as Computer Vision, NLP, etc. and do not draw inspiration from such theories, although a very recent work prosed by \citep{Arani2022LearningFL} explored over this direction. With some recent developments in the CL field, rehearsal systems \citep{review0} (systems that replay old data through a buffer) can be recast with such a point of view. In fact, we can think of the rehearsal buffer (or the part of the CL system dedicated to storing ``old'' patterns used in replay) as a long-term memory while the other part of the architecture is the fast-paced learner of the intelligent agent i.e. the hippocampus. Perhaps the key to continual learning will be in the inspiration from neuroscientific models. Indeed recently \citep{McCaffary2021TowardsCT} proposed a systematic review of the approaches in CL along with some insights into why we should pay more attention to neuroscientific theories.
% neurosci + continual learning https://arxiv.org/pdf/2112.14146.pdf DA METTERE DOPO???/

As we saw, the gap between artificial and biological models is still relevant and the two fields, nowadays, show big differences in their understanding of intelligence. However, one striking fact is that the artificial community has achieved impressive results without \textit{directly mimicking} the current neuroscientific theories, suggesting that, perhaps, several paradigms of intelligence exist.

\section{What is Continual Learning?}
\textit{``Every machine is built to make decisions, if it does not have the faculty to learn, it will act always in conformity to a mechanical scheme.  We don't have to let the machine decide about our conduct if we first have not studied the laws that rule its behavior, and made sure that such behavior will be based on principles that we can accept!'' \hfill - Norbert Wiener}

\paragraph{Definition:} The aforementioned quote is taken from ``Introduction to Cybernetics'', and highlights the fact that the fundamental ability to\textit{ continually learn} is a very important skill that any intelligent system should possess.  Although we are now able to devise powerful artificial systems achieving superhuman performance in some tasks, we, as humans, still exhibit a core ability that would be fundamental to replicate intelligence as we know it. The ability to learn new concepts without erasing past knowledge. These two aspects are the main objectives of Continual Learning. First, exhibiting the ability to assimilate new concepts incrementally. Secondly, showing the capability of memorization i.e. not forgetting what has been previously learned. In a nutshell \textbf{Continual Learning studies how to develop systems that learn incrementally over time without forgetting previously acquired knowledge}.

\paragraph{History:} Continual Learning has drawn a lot of interest from the research community only in the later years even though the question itself is very old.  One of the early papers trying to tackle this phenomenon has been proposed by \citep{carpenter1988art} where the authors proposed a short-term and long-term memory pattern detector through the Adaptive Resonance Theory. In fact, to the best of our knowledge, this seems to be the earliest work proposed. Later, as connectionist models pave the way for modern Artificial Intelligence, other attempts and several proposals have been made. Later the work by \citep{ring1994continual} coined the term ``Continual Learning'', here the system proposed, aimed to construct hierarchies of knowledge within a neural network. Later, with the works by  \citep{DBLP:phd/dnb/Thrun95} and \citep{DBLP:journals/ras/ThrunM95} Continual Learning started to get attention especially in both the Robotic and Reinforcement Learning research community. 

\paragraph{Terms:} When we say Continual Learning we have two other equivalent terms: \textit{Incremental Learning }and\textit{ Lifelong Learning}. These terms can be used interchangeably and denote the same setting. There are no clear distinctions and probably the preference of one over another is just a matter of the research field we are in. For example, in the computer science field, it seems that continual learning and incremental learning are more common. Other terms are used but differ in the specific continual setting they study. For example: \textit{Online Learning }and \textit{Streaming Learning}. These are very similar, and there is no clear distinction yet. These terms are used to describe algorithms that learn by observing an example just one time and, sometimes, the latter can also refer to systems that can respond to queries in real-time. We will introduce a more formal definition in the next chapter.

\paragraph{Subject of CL:} As we previously discussed, the study of Continual Learning is strictly tight with the widespread usage of connectionist models. In fact, before the advent of Artificial Neural Networks (ANNs), intelligence was modeled, usually, by a mixture of expert systems and clever algorithms. Posing the same ``continual learning question'' for these systems is still an interesting challenge, but the success of ANNs shifted the focus to connectionist models.

\begin{figure}[t]
	\centering
	\includegraphics[width=\textwidth]{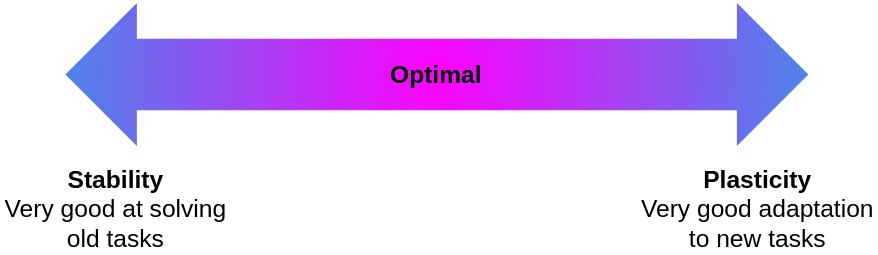}
	\caption{Continual learning spectrum. The optimal algorithm should exhibit enough plasticity to learn new tasks while retaining enough stability to not forget the acquired knowledge.}
	\label{fig:dilemma_depiction}
\end{figure}

\subsection{Stability-Plasticity Dilemma}
 
Learning incrementally (or continually) with connectionist models requires one core ability, that is, \textbf{to adapt to a changing environment}. If the environment would not change over time, and we expose a system to operate on it, we would just need to understand, model, and hard-code the environment's rules to the system and we would achieve perfect functionality. Unfortunately, the real world does not seem to behave in such a predictable way. Instead, our reality constantly changes and we need to redefine our knowledge, reshape it in light of new facts, have room to constantly learn something new, and recombine previous knowledge to understand a novel concept. This is not \textit{the only} necessary property for an intelligent system, the counterpart is also important. In fact, some things do not change in the world, old challenges might propose again, and, therefore, fundamental knowledge should not be forgotten. A truly intelligent system would \textbf{ behave consistently on past lessons}. It would be able to detect and recognize past challenges, delivering correct solutions. The researchers gave a name to this trade-off and it is called the \textit{stability-plasticity dilemma}. The long-term goal of Continual Learning is to create a system able to achieve a perfect balance between these two abilities as depicted in Figure \ref{fig:dilemma_depiction}. As we will see, it is termed a ``dilemma'' since achieving the optimal trade-off is a very hard task.

On top of these considerations dissecting new concepts and redefining them as a combination of old knowledge allows the forward transfer of intelligence. That is when we learn we sometimes can abstract the knowledge to solve a related problem. This is not uncommon it is the mechanism of \textit{analogy thinking} where an ``operational pattern'' can be used to solve problems in apparently different domains. As an example, \citep{DBLP:conf/iclr/HillSBML19} investigates such property of intelligence in artificial networks. On the other hand, continual learning should give the ability to better grasp the past knowledge improving the ability to past challenges. This is even more common and we can think of this kind of ability as the ``experience'' that an agent accumulates in a certain field or in solving a certain category of tasks. 

In a nutshell, the stability-plasticity dilemma can be considered the crux of intelligence. Showing adaptability to new environments while at the same time retaining knowledge of old environments seems to be the major qualities of an intelligent agent.

\subsection{Catastrophic Forgetting}

One core aspect of deep neural networks lies in the fact that if we do not introduce any kind of mechanism to achieve the balance between stability and plasticity, \textbf{the artificial network is naturally inclined to forget}. That is, the neural networks put much more emphasis on plasticity rather than stability. From a neuroscientific point of view, this fact does not make much sense unless we think about neural networks as systems without any form of memory. The reality is that networks do have memory, but by the nature of the learning algorithms we overwrite such memory. As the model incrementally learns, \textit{each parameter in the network is modified by the updates of the backpropagation algorithm}. The optimal continual learning method would be able to modify the parameters without altering the performances of old tasks. This seems not to happen and therefore neural networks are prone to the so-called \textit{catastrophic forgetting}, the phenomenon where old knowledge is corrupted.

\begin{figure}[t]
	\centering
	\includegraphics[width=0.9\textwidth]{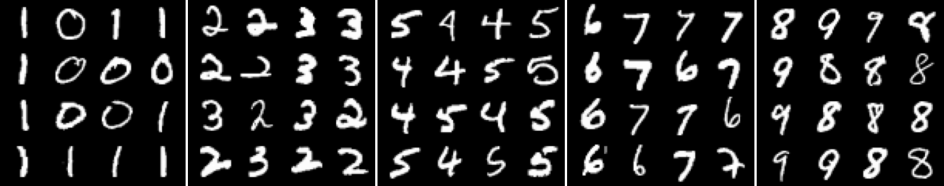}
	\caption{The original images for each task. This image shows the ground truth relative to Figure~\ref{fig:example_glued}.}
	\label{fig:example_true}
\end{figure}
 
\subsection{A Visual Example}
\label{sec:cl_visual_example}

To better grasp the phenomenon of catastrophic forgetting, we will provide a visual example in the following section. As we discussed, catastrophic forgetting happens because the parameters tuned to solve a task (usually experienced before in time), are not suited for the currently experienced task. We hope to provide a clear visual example of the effects of catastrophic forgetting in a shallow architecture. 

As the name suggests Deep Learning refers to architectures with many layers on top of each other. Because of this huge depth, computer vision (but not only this community) was able to achieve impressive results in the domain of pattern recognition. Unfortunately, we still do not \textit{fully} control how the knowledge is built inside a deep neural network and if we want to counter forgetting we would need such information. To do so, we would need to \textit{keep track of each parameter variation as we learn new concepts} in a continuous fashion, but doing so, especially in such models, is hard if not an impossible job. Said that, on a small scale, we can still show what is going on inside a network. In the following toy example \textbf{we try to track forgetting of an autoencoder by dissecting the learning process per task}. We will use a simple one-layer autoencoder model and try to incrementally learn the famous MNIST \citep{mnist} dataset, still used in the continual literature to validate the proposed methods. We will divide the dataset into 5 tasks and learn to compress and reconstruct images. By doing so we will show the corruption of old images as we learn new tasks and connect them to the network's variation of the parameters. 

The MNIST dataset is a grayscale dataset of $28 \times 28$ images of handwritten digits going from the digit 0 to the digit 9, here some examples: \includegraphics[height=\myMheight]{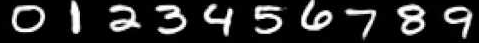} . The MNIST was constructed from NIST's Special Database 3 and Special Database 1, the first has been collected among Census Bureau employees and the second one among high school students. It has a training set of 60,000 examples, and a test set of 10,000 examples. We will divide the dataset in 5 tasks, the first 1 is composed of the digits \includegraphics[height=\myMheight]{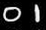}, task 2 by \includegraphics[height=\myMheight]{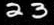}, task 3 by \includegraphics[height=\myMheight]{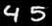}, task 4 by \includegraphics[height=\myMheight]{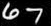} and finally task 5 by \includegraphics[height=\myMheight]{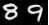}

Although the best practice to work with image data is to use CNNs, we will limit our toy example to a naive autoencoder model of linear layers. This choice allows us to better unfold and analyze the variation of the parameters due to its simplicity. The model is composed of a single layer encoder $\phi$ that encodes an image into a latent vector and a single layer decoder $\psi$ that reconstructs the image. In particular, the single-layer encoder is a linear layer $\phi: \R^{784} \rightarrow  \R^{16}$ that will receive in input a flattened ($28 \times 28 = 784$) representation of the image and compress into a latent vector of magnitude 16. The decoder, then take care of the reconstruction of the image by doing the reverse process, that is  $\psi: \R^{16} \rightarrow  \R^{784}$ i.e. given a latent vector of size 16 it decompresses it to a flattened image.  

More formally an autoencoder can be represented in the following way:
\begin{figure}[t]
	\centering
	\includegraphics[width=\textwidth]{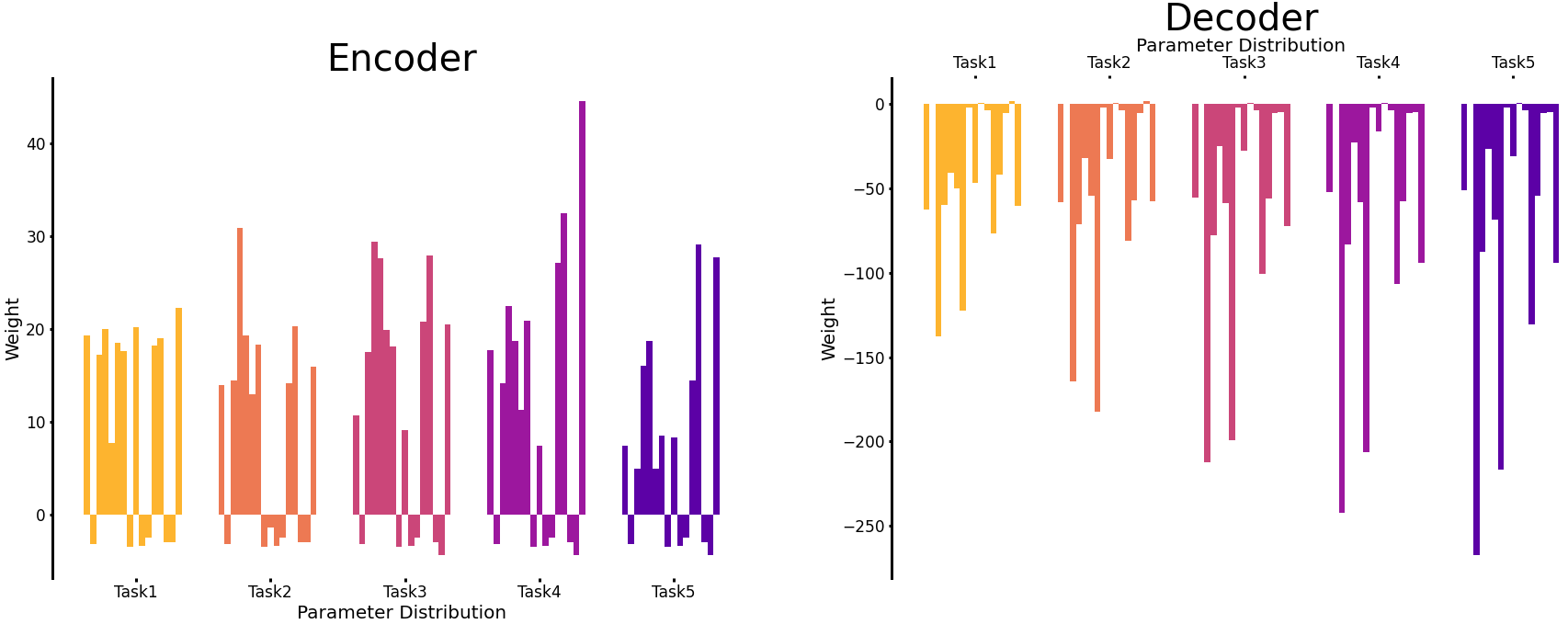}
	\caption{Variation of the parameters grouped by task. Each bar plot shows the distribution of the weights, we can see that each task modifies internal parameters. Each weight is computed as the sum of all the connections of the particular latent neuron.}
	\label{fig:example_distributions}
\end{figure}
\begin{equation*}
\hat{\vx} = \psi(\phi(\vx))
\end{equation*}
Where $\vx \in \R^{784}$ is the flattened representation of an original image coming from a task $t$, $\phi$ is the encoder network and $\psi$ is the decoder network, and $\hat{\vx} \in \R^{784}$ is the flattened representation of the reconstructed image. The objective is to optimize and, in particular, minimize the mean square error (MSE) between the original image and the encoder's reconstruction. More formally we can define the objective function as:
\begin{equation*}
\min_{\phi_\Theta, \psi_\Theta} \mathcal{L}\left(\vx, \hat{\vx}\right)= \min_{\phi_\Theta, \psi_\Theta} \left\|\vx-\hat{\vx} \right\|^{2}
\end{equation*}
%=\left\|\mathbf{x}-\sigma^{\prime}\left(\mathbf{W}^{\prime}(\sigma(\mathbf{W} \mathbf{x}+\mathbf{b}))+\mathbf{b}^{\prime}\right)\right\|^{2}
Here $\phi_\Theta$ represents the set of encoder's parameters to be optimized while we use $\psi_\Theta$ for the decoder. 

%We will use Adam \citep{adam} optimization with learning rate 0.01 and of with 10 epochs per task, batch size of 64.

\begin{figure}[th!]
	\centering
	\includegraphics[width=13cm]{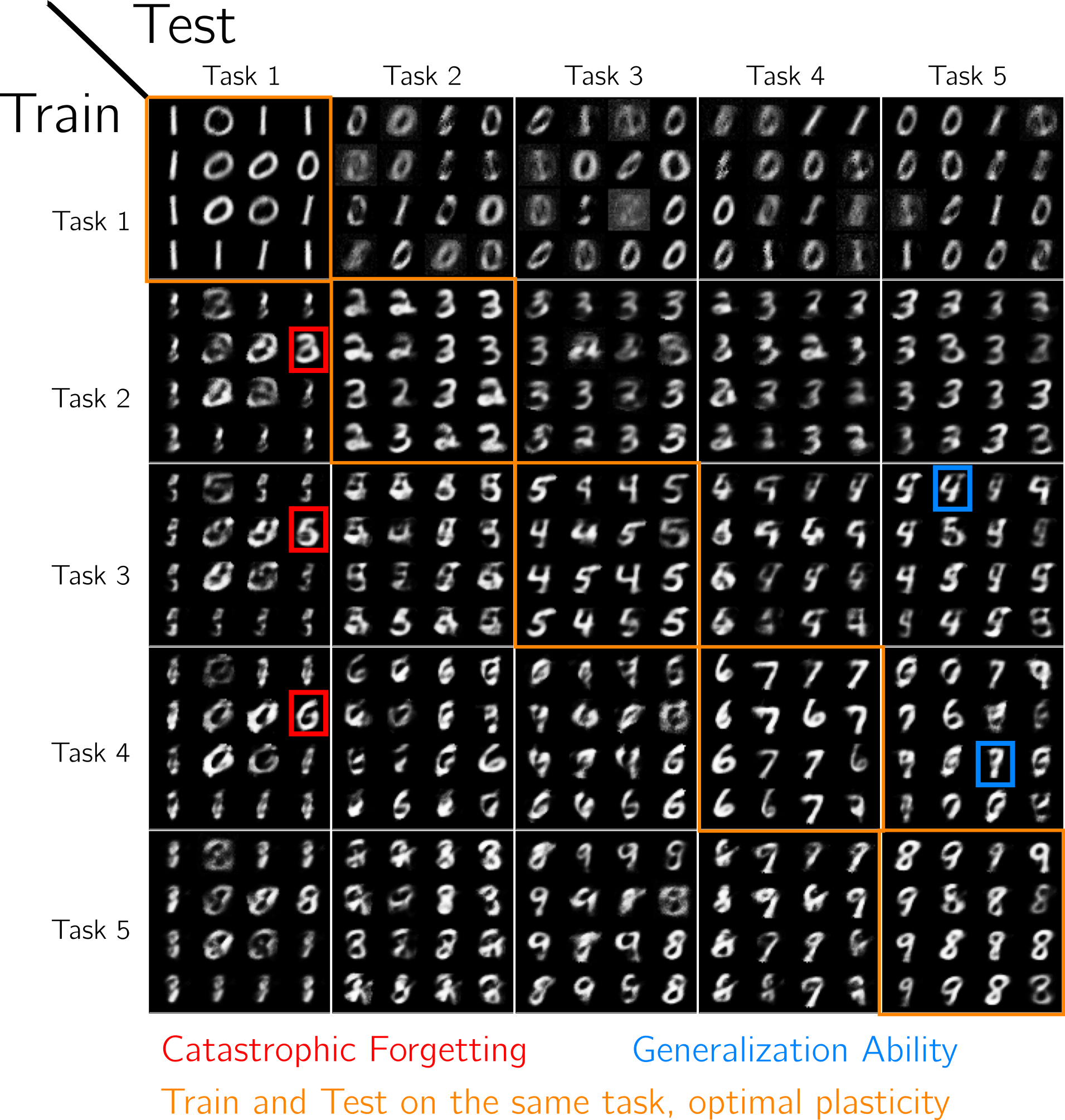}
	\caption{Results of the incremental training and test of the autoencoder model in the MNIST dataset were split into 5 tasks. Each row $i$ of the grid, reports the performance of the model when trained on task $i$ (or time $t_i$) and tested on both old (left) and future (right) tasks.  Training on previous tasks might unlock the intrinsic possibility to solve future tasks. This latest phenomenon is highlighted with the blue boxes. Ground truth in Figure~\ref{fig:example_true}.}
	\label{fig:example_glued}
\end{figure}

By \textit{incrementally learning} each task we want to show the \textbf{corruption in the ability of reconstruction of previous tasks}. The change in the parameters to accommodate the new task negatively impacts old tasks. In fact, if we try to retrieve old concepts we see catastrophic interference, that is, the network is confusing old concepts with newly learned ones. From now on let us refer to Figure~\ref{fig:example_glued}, where is depicted the complete incremental learning and its effects. The grid reported encodes the performance of the autoencoder. Each row $i$ refers to the model trained solely on data of task $i$ but tested on all the other tasks. From the experiment, we can appreciate several effects. First, if we isolate the first column of the grid, we can visualize the performance of the original first task as time passes (we can think of it as the stability of the network as we will discuss in Section \ref{sec:stab_plast_curves}). Here, one can clearly see that feeding new concepts corrupts old ones. On the other hand, if we focus on the upper triangular section of the matrix, we see the ability of the model to generalize knowledge. This stresses the fact that generalization is a key component in continual learning. Intuitively more ``general'' models might experience less forgetting (further hints on this path can be found in Section \ref{wrk:cl_vit} and Section \ref{wrk:off-the-shelf}). The connected change in the weights for each task is reported in Figure~\ref{fig:example_distributions} (for both the encoder and decoder). As we can see, even a small change in the parameters dramatically impacts the stability plasticity trade-off. As reference in Figure \ref{fig:example_true} we report the ground truths.

%But what happens if we increase/decrease the number of the neurons? This is a good question and it is worth exploring a bit on that side. In fact if we increase the number of the neurons to be 128 we can see almost no catastrophic forgetting, instead we would see perfect generalization on future tasks. This might be due to the simplicity of the dataset but also to the generalization ability of the model. What happens if we freeze the encoder and just train the decoder?

%\begin{equation}
%\hat{x} = \text{ReLU}(U^{\phi}\cdot x)^T U^{\psi}
%\end{equation}

%\begin{figure}[t]
%	\centering
%	\includegraphics[width=0.8\textwidth]{images/fgt_example/forgetting.png}
%	\caption{An example of catastrophic forgetting. We train an autoencoder on task one (digits 0 and 1 at time $t-1$) then we will test it on task two (digits 2 and 3) at time $t$). On the right we see the forgetting of the model learned at time $t$ when prompted to the data of time $t-1$. The depiction of the weights and the reconstruction examples are real and taken from the studied example.}
%	\label{fig:example_forgetting}
%\end{figure}

\chapter{Continual Learning Framework}
\section{Definition and Settings}

Being Continual Learning a relatively new discipline,\textit{ the community unfortunately still does not fully agree on a formal setting}. This is also corroborated by the fact that incremental learning is under the research light of several communities. Among the most active communities, we have NLP, Computer Vision, Reinforcement Learning, Neuroscience, and Robotics. Each of these communities has a well-established history and standard protocols, therefore, accommodating everyone in a common ground is still an ongoing process. However, in the following, we will introduce the most common definitions and settings shared in the Computer Vision literature.

There have been some attempts to formalize a setting for continual learning \citep{van2019three, core50} through the definition of learning protocols and new terminologies. We will see these different learning paradigms in the following sections, but the core feature underlying learning incrementally is that the data experiences some distributional shift, that is, the \textbf{distribution of the data changes over time}. This is sufficient to abruptly cause forgetting in connectionist models, but we can define some settings which are more prone to cause such phenomenon, while others are more simple to overcome.

\textit{The typical continual learning setting in computer vision is composed of a split dataset, where each (usually non-overlapping) split is considered an incremental task}. Therefore, each task contains data from several classes. Although this is not the only way to define a continual learning scenario, this is the most prominent one as pointed out in these surveys \citep{Mai2022OnlineCL, 9349197, review0}. Let us define a more formal definition: 

\paragraph{Formal Definition:}
Given a dataset $\mathcal{D}$ containing (in our case) images, we want to split $\mathcal{D}$ in a sequence of $n$ disjoint tasks that can be learned sequentially by our model:
\begin{equation}
\mathcal{T}=\left[ t_1, t_2, \dots, t_n \right]
\end{equation} 
where each task $t_i = (C^{i}, D^{i})$ is represented by a set of classes $C^{t}=\left\{c_{1}^{t}, c_{2}^{t} \ldots, c_{n^{t}}^{t}\right\}$ and training data $D^{t}$. We use $N^{t}$ to represent the total number of classes in all tasks up to and including task $t: N^{t}=\sum_{i=1}^{t}\left|C^{i}\right|$. As a side note, usually in literature one would use the notation $t$ to point at the current task (the task at time $t$) and $t-1$ to point to the task before the current one.

\begin{figure}[t]
	\centering
	\includegraphics[width=0.9\textwidth]{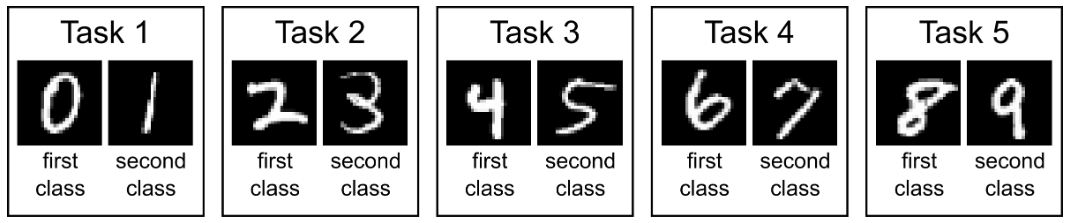}
	\caption{Schematic representation of split-MNIST task protocol.Taken from \citep{van2019three}}
	\label{fig:class_incremental}
\end{figure}

A continual learning algorithm aims to model each task sequentially as time passes exposing the model at training time to each task in a sequential fashion. Operatively: first, the algorithm is trained with mini-batches of patterns coming from task 1. Here we will record the system performance. Then, the model is exposed to task 2 data and the process continues until task $n$. One visual example can be seen in Figure~\ref{fig:class_incremental}, here the MNIST dataset is split into 5 tasks with 2 classes each \footnote{This particular setting takes the name of MNIST-split}. 

The previously defined learning scenario takes into consideration a distinct transition among tasks. In this particular case, we implicitly assume a \textit{reset signal} between two tasks. When such signal is not present, and the transition between tasks is \textit{smooth}, the complexity of the continual learning problem increases. If in this particular setting we query the system for real-time response, we are talking about streaming learning \citep{DBLP:conf/icra/HayesCK19}. This setting is more challenging because the models are allowed much less time to consolidate previously seen knowledge and therefore are more prone to experience catastrophic forgetting. Since this thesis focuses on computer vision problems, throughout the work we will stick to the introduced setting.

\paragraph{Fine-Grained} So far we limited the notion of a task as a split of a dataset, but what happens if in a new task we experience new instances of previously seen classes? To this end, more complete settings for continual learning benchmarking have been proposed. One example is constituted by \citep{core50}. Here the authors, along with a new dedicate dataset, introduce three different settings by mixing the experience of old and new data. Specifically, here we report the different scenarios:
\begin{itemize}
	\item \textbf{New Instances (NI)}: new training patterns of the same classes become available in subsequent tasks. Here the model can experience new instances of old, previously seen, classes. With the possibility of seeing the same objects in new poses and conditions (illumination, background, occlusion, etc.). Here a good model is expected to incrementally consolidate its knowledge about the known classes without compromising what it has learned before.
	
	\item \textbf{New Classes (NC)} \label{scenario_nc}: new training patterns belonging to different, never seen, classes become available in subsequent tasks. This is the classic scenario (the one we formally introduced) and a model should be able to deal with the new classes without losing accuracy on the previous ones.
	
	\item \textbf{New Instances and Classes (NIC)}: new training patterns belonging both to known and new classes become available in subsequent training tasks. A good model is expected to consolidate its knowledge about the known classes and learn the new ones. This is the most complete and difficult scenario since the addition of new classes poses the challenge of having good plasticity while the introduction of new old patterns asks for stability.
\end{itemize}
In our opinion, this categorization is preferable since it provides a more complete description of a continual learning benchmark. In fact, if we assume, as an example, that each task data is generated by an independent source, the task data will be continually augmented with new information. This scenario is captured by the NIC setting and cannot be handled by the standard definition. Unfortunately, due to the recent development of the field, we usually assume the NC scenario independently.

\subsection{Online CL vs Offline CL}
So far we introduced a basic notation, now we discuss \textit{how a model can be trained} to face a continual learning stream of tasks and introduce the name of these scenarios. The continual learning literature distinguishes two options: online training and offline training. 

\paragraph{Online} In particular, in the online continual learning protocol, the algorithm is required to have a \textit{single parameter update per pattern} (or one forward-pass). This is a very coercive setting and requires maximum performance in knowledge consolidation from the continual learner. In fact, this scenario is quite challenging because of the nature of Stochastic Gradient Descent i.e. the learning algorithm at the core connectionists models. Here the system might not have enough time to assimilate a concept, therefore weakening its understanding and subsequent stability.
%Typically the learning process requires the model to go through several passes of the data (each pass is called epoch). Such passes are required to build knowledge by approximating the gradient direction to optimize the loss function, unfortunately these can be also the cause of forgetting.

\paragraph{Offline} In the offline learning protocol, instead, we are \textit{free to perform several parameter updates per pattern} i.e. we are allowed to see an image more than once. For an incremental learner, this setting is a double edge sword, in one case it favors the consolidation of the concepts since setting a large number of epochs guarantees the correct training of a model. On the other side, if we do not introduce any forgetting prevention mechanism, this corrupts the old informational content of the network i.e. the system is more exposed to catastrophic forgetting.

In the following paragraphs, we will introduce some of the settings that are now, de facto, shared among all the research communities researching continual learning.

\subsection{Task-Incremental vs Class-Incremental}
Assuming an NC-type of task flow, two sub-settings have been widely adopted by the research community and are well-defined. The Task Incremental (TI) setting and the Class Incremental (CI) setting. 

\paragraph{Task-Incremental} In the Task Incremental scenario, which is sometimes also referred to as \textit{multi-head} scenario or \textit{task aware (TAw)} the learning happens sequentially, but at test time, the learner \textit{has also access to the task label}. This scenario is also known as multi-head because a typical learning system can potentially dedicate a particular subsystem per task, that can be specifically queried at test time through the task label knowledge. Typically the subsystem is a classifier head on top of a backbone. 

More formally we consider task-incremental classification problems where at training time the learner has access to:
\begin{equation*}
D^{t}=\left\{\left(\mathbf{x}_{1}, \mathbf{y}_{1}, \mathbf{z}_{1}\right),\left(\mathbf{x}_{2}, \mathbf{y}_{2}, \mathbf{z}_{2}\right), \ldots,\left(\mathbf{x}_{m^{t}}, \mathbf{y}_{m^{t}}, \mathbf{z}_{m^t}\right)\right\}
\end{equation*}
while at test time the learner has access to:
\begin{equation*}
D^{t}=\left\{\left(\mathbf{x}_{1}, \mathbf{z}_{1}\right),\left(\mathbf{x}_{2}, \mathbf{z}_{2}\right), \ldots,\left(\mathbf{x}_{m^{t}}, \mathbf{z}_{m^t}\right)\right\}
\end{equation*}
where $\mathbf{x}$ are input features for a training sample, and $\mathbf{y} \in\{0,1\}^{N^{t}}$ is a one-hot class ground truth label vector corresponding to $\mathbf{x}_{i}$ while $\mathbf{z}\in\{0,1\}^{|\mathcal{T}|}$ is the is a one-hot task ground truth label vector. In a nutshell, during training for task $t$, the learner only has complete access to $D^{t}$, then we assume a reset signal among tasks i.e. $C^{i} \cap C^{j}=\emptyset$ if $i \neq j$, and at test time the learner has access to patterns and their task label.

\paragraph{Class-Incremental} Instead, in class incremental scenario, also known as \textit{single-head} or \textit{Task Agnostic (TAg)} the system has both access to task and class label during training time, but at test time it only has raw data. This constitutes a harder problem, but also a more realistic scenario. 

More formally we consider class-incremental classification problems where at training time the learner has access to:
\begin{equation*}
D^{t}=\left\{\left(\mathbf{x}_{1}, \mathbf{y}_{1}, \mathbf{z}_{1}\right),\left(\mathbf{x}_{2}, \mathbf{y}_{2}, \mathbf{z}_{2}\right), \ldots,\left(\mathbf{x}_{m^{t}}, \mathbf{y}_{m^{t}}, \mathbf{z}_{m^t}\right)\right\}
\end{equation*}
while at test time the learner has access only to:
\begin{equation*}
D^{t}=\left\{\mathbf{x}_{1},\mathbf{x}_{2},\ldots,\mathbf{x}_{m^{t}}\right\}
\end{equation*}
where $\mathbf{x}$ are input features for a training sample, and $\mathbf{y} \in\{0,1\}^{N^{t}}$ is a one-hot class ground truth label vector corresponding to $\mathbf{x}_{i}$ while $\mathbf{z}\in\{0,1\}^{|\mathcal{T}|}$ is the is a one-hot task ground truth label vector, same as in TAW setting.

Although taw scenarios are more interesting from a pure machine learning perspective, the tag setting is more realistic. For example, let's draw an analogy: let us consider a baby as our incremental algorithm. We want to teach the baby to recognize elements coming from a particular environment, for example, kitchen accessories. Here the task label would be 'kitchen'. After the learning process has successfully terminated, whenever we ask the baby to recognize a fork, we do not need to provide a hint on the task (kitchen). In fact, the information of \textbf{where} he learned the concept should be irrelevant. This is also important because several objects can appear, and could be part of, several environments (tasks). For example, scissors can be found in the kitchen, but also in a studio. Therefore knowledge itself should be independent of the context where it is learned and, we think that class incremental setting provides a more useful challenge.

\section{Baselines}
In this chapter, we will see the principal naive approaches and introduce an overview of the state-of-the-art. In particular, we will introduce the cumulative and the finetuning methods which constitute, respectively, the upper and the lower bound to evaluate continual learning strategies. Moreover, we consider our model to be composed of a backbone (or a feature extractor) and a dedicated classifier (head) for each task. We do so in light of the majority of the works in continual learning and computer vision, which are composed of this very structure. 

\subsection{Cumulative}
\begin{figure}[t]
	\centering
	\includegraphics[width=\textwidth]{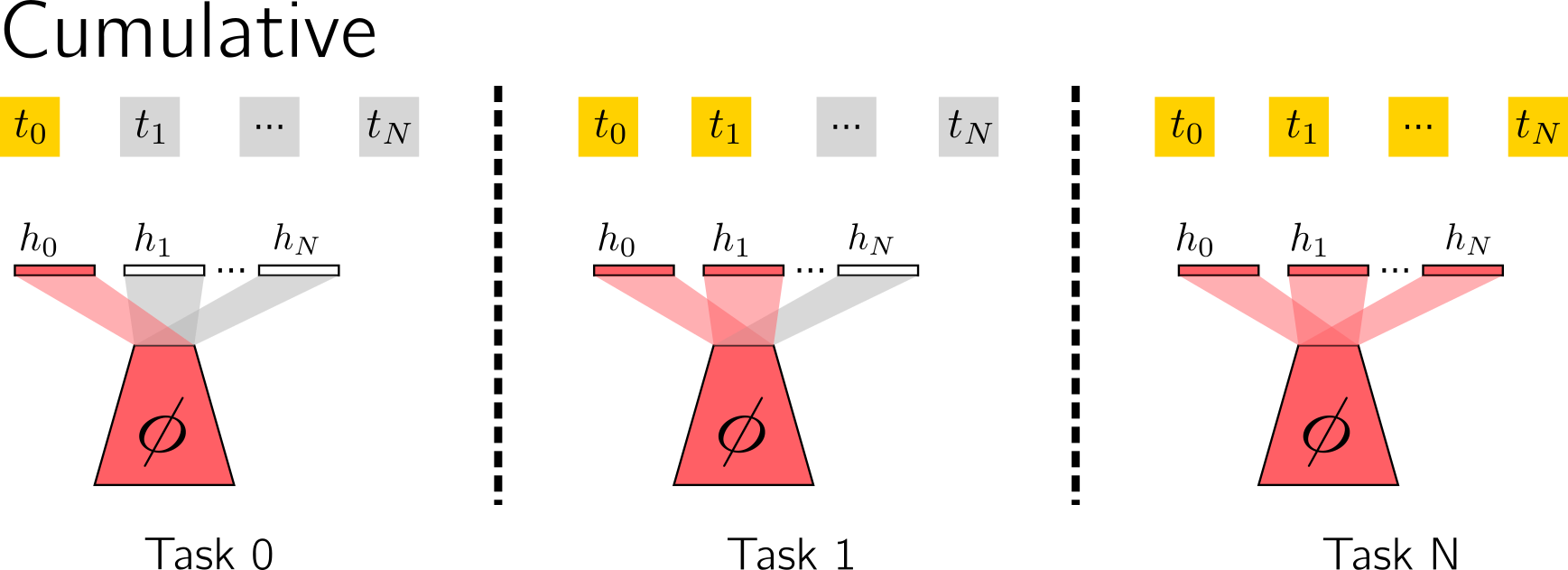}
	\caption{Depiction of the Cumulative/Joint approach for continual learning. The model is trained with all the data up to the current task $t_i$. The updates flow in the backbone and in all the heads up to $h_i$.}
	\label{fig:cumulative}
\end{figure}
To evaluate a continual learning algorithm we need an optimal method that acts as an \textbf{upper bound}. The cumulative strategy (also known as \textit{joint-training}) constitutes the optimal continual learning strategy since \textbf{mimics a learner with perfect memory}. Indeed if we have perfect memory we can recall the past and not experience forgetting, to this end a recent work from \citep{optimalcl} proved theoretically that optimal continual learning has a perfect memory and is NP-hard.

To have optimal memory of the past, an algorithm should be able to save all the data that has been seen. This is a very inconvenient requirement and it must be avoided when considering the development of real lifelong learning systems. In fact, as the pace of real-world data generation is growing, such constraints would not be satisfied. Training from scratch with all the dataset data could be an upper bound approach, but it does not break down each incremental step upper bound. To this end, the cumulative strategy accumulates all the data seen up to a certain task and trains the network from scratch, therefore, providing an incremental upper bound. More formally, for the cumulative approach, the data of task $i$ is defined to be:
\begin{equation*}
t_i = \bigcup_{j=0}^{j=i} t_j
\end{equation*}
when $i = 1 \dots n$ to complete the incremental setting. At each time $t_i$ the model is trained on the cumulative data and therefore we are able to define the upper bound performance for each task $i$. One observation is that the cumulative performance in the last task it is equivalent to the performance of the model trained with the whole data. In Figure~\ref{fig:cumulative} we depict a visual example of the cumulative approach. Here for each task, the backbone is always updated along with the heads of competence. However the updates of the heads can be also shared among all the tasks, that is, each task data alters all heads parameters. Of course, this design choice does not favor the prevention of forgetting, instead, it allows the disruption of consolidated knowledge and we won't consider this case\footnote{this is valid for finetuning too}.

\begin{figure}[t]
	\centering
	\includegraphics[width=\textwidth]{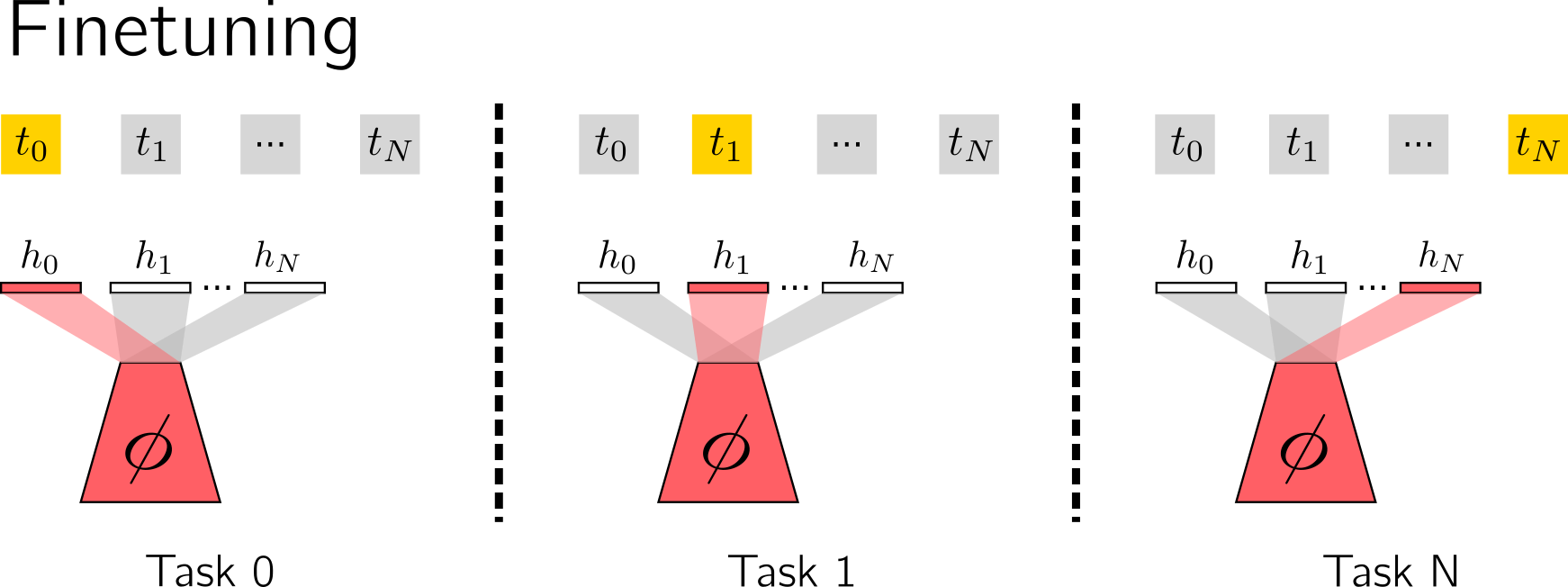}
	\caption{Depiction of the Finetuning approach for continual learning. The model is trained exclusively with the data coming from the current task $t_i$. The updates flow in the backbone and only in the $h_i$ head.}
	\label{fig:finetuning}
\end{figure}

\subsection{Finetuning}
We previously saw the upper bound for CL, that is, the optimal continual learning approach for a benchmark. Now, we introduce the finetuning approach which constitutes the \textbf{lower bound} methodology. Although we can argue that a random classifier would be the true lower bound, in practice we consider finetuning in which it is absent of any forgetting prevention mechanisms. In fact, it is equal to the practice of \textbf{transfer learning among subsequent tasks} and measures the base resilience of the model against incremental scenarios. We also can consider it as a baseline to assess the generalization capabilities of a model. 

A depiction of the method is given in Figure~\ref{fig:finetuning}. Here, the model is trained sequentially and each task head is updated with the data of its competence and, as in the cumulative approach, the backbone is always updated.

\section{State-of-the-art}
In the following sections, we will introduce the main categorizations of the approaches proposed by the community. In particular, we will explain the core mechanism and show the pros and cons of each category. Although there is no absolute preferred solution, some approaches are more explored than others and show more promising results.
%we will see the different main categories of approaches that have been widely used in continual learning and discuss the main pros and cons of each.

\subsection{Structural-based}

Structural-based approaches, also known as \textit{architectural approaches} or \textit{parameter-isolation} methods, fight forgetting by altering the structural composition of the network itself. In particular, structural approaches instantiate dedicated modules as they experience new tasks. The first work falling in this category is perhaps Progressive Neural Networks (PNN) \citep{pnn} where the network is augmented with new connections spanning both height-wise and width-wise.

\begin{figure}[h!]
	\centering
	\includegraphics[width=0.7\textwidth]{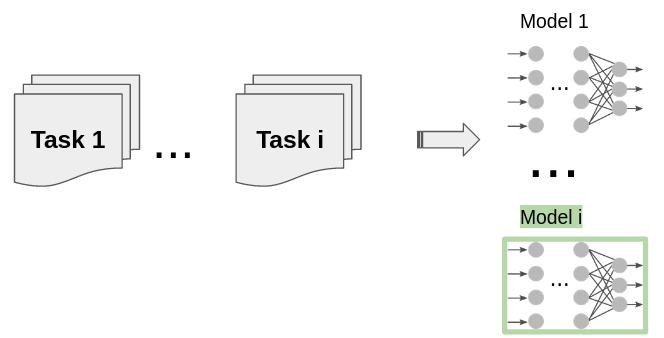}
	\caption[]{Architectural approaches for Continual Learning alter the structural properties of the network itself.}
	\label{fig:architectural_appproaches}
\end{figure}

In the task-aware setting, this approach constitutes a convenient and naive solution to fight catastrophic forgetting. In fact, having the task label at test time allows us to correctly determine a dedicated subnetwork. Instead, in task agnostic setting, we would not be able to select such a submodule. We can see very few structural-based approaches tackling class incremental setting due to the aforementioned limitation \citep{DBLP:conf/iclr/LeeHZK20, DBLP:conf/nips/RajasegaranH0K019}. That said, Structural approaches can be subdivided into Fixed Architecture (FA) and Dynamic Architecture (DA). FA only activates relevant parameters for each task without modifying the architecture \citep{packnet, ewc}, while DA adds new parameters for new tasks while keeping old parameters unchanged \citep{DBLP:conf/iclr/YoonYLH18, pnn}. Although architectural methods are very intuitive, they are bulky. In fact, the major drawbacks are in the expansion of the parameters which can result in a memory-intensive method (DA), or in the architectural limitation of the number of parameters that can be saturated (FA).

\subsection{Regularization-based}
In parameter based approaches also known as \textit{weight-regularization} or \textit{data-regularization} approaches, forgetting is handled with procedures that regularize the parameter updates. Among the most famous ones, there are Elastic Weight Consolidation (EWC) \citep{ewc} and Synaptic Intelligence (SI) \citep{si}. EWC was the first regularization-based approach using second-order information. In particular, the procedure regularizes the updates through the Fisher information which is computed at each parameter update. 
 
\begin{figure}[h!]
	\centering
	\includegraphics[width=0.6\textwidth]{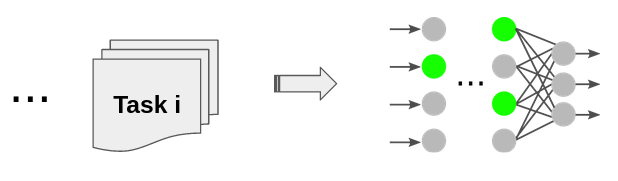}
	\caption[]{Regularization approaches for Continual Learning alter only the parameters properties of the network.}
	\label{fig:regularization_appproaches}
\end{figure}

In this category, we can also find Learning without Forgetting (LwF) \citep{DBLP:conf/eccv/LiH16}, which is one of the most influential methods in continual learning literature. LwF uses Knowledge Distillation \citep{kd} in the logits of the network. The main strength of LwF lies in the fact that it does not use previously-stored examples while still being purely data-driven. In particular by storing the old model at time $(t-1)$ the method can distillate old knowledge by forwarding to the old model the current data. Since the introduction LwF, KD has been widely adopted by the continual learning community as part of new methodologies among the works we report \citep{podnet, icarl, der, kd1, kd2, kd3, kd4, kd5, kd6, dhar2019learning}, but we are aware of many others that we do not report for brevity. The main strength of regularization-based approaches lies in their data/architecture constraint-free nature. In fact, they usually work with an underlying mathematical justification. This property surely allows a more principled continual learning strategy, but it can make the learning procedure cumbersome: computing second-order or estimating gradients directions, might slow down the learning while hindering it.

\subsection{Rehearsal-based}
\begin{figure}[h!]
	\centering
	\includegraphics[width=0.8\textwidth]{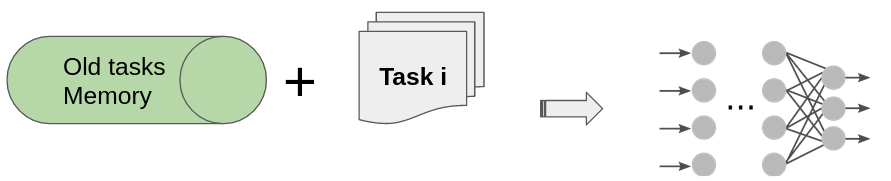}
	\caption[]{Rehearsal approaches for Continual Learning store old patterns to augment the data of the current task.}
	\label{fig:rehearsal_appproaches}
\end{figure}

In rehearsal-based approaches (or \textit{data-replay} approaches) the main mechanism exploited to overcome forgetting, lies in the usage of a \textbf{replay buffer} for old exemplars. The methods falling under this category, dedicate a memory cache to store data examples encountered during the incremental training i.e. the system samples and stores images experienced in previous tasks. We can think of the buffer as long-term memory. In fact, what typically happens is that the memory is queried to augment the task at hand, that is, we retrieve and inject old examples to the current data batch. This mechanism prevents forgetting by allowing the network to directly recall past examples, a visual depiction can be seen in Figure~\ref{fig:rehearsal_appproaches}.

Perhaps the most famous work among rehearsal-based approaches is Experience Replay (ER) \citep{er} inspired by the Reinforcement Learning community its strategy is replaying data by randomly selecting old examples. In the evolution of ER, which is Maximally Interfered Retrieval (ER-MIR) \citep{ermir}, proposed a controlled sampling of the replays. Specifically, they retrieve the samples which are most interfered with, i.e. whose prediction will be most negatively impacted by the foreseen parameters update. Another famous method is Gradient Episodic Memory (GEM) \citep{gem} in which the authors devised a system where the gradient update of the replay examples should follow the original direction.

A closely related mechanism is \textbf{generative replay} (GEN) \citep{DBLP:conf/nips/ShinLKK17, vandeVen2018GenerativeRW, Wu2018IncrementalCL}. In this approach, old data is recorded in a buffer and then compressed, after that, a generative model such as a GAN \citep{gans}, generates a \textit{synthetic version} of the old distribution and augments the data of the current task. The main disadvantages of generative replay are that it takes a long time to train and it does not constitute a viable option for more complex datasets given the current state of deep generative models. Another approach devised by \citep{liu2020generative} tries to overcome such limitations by generating intermediate features instead of the original data, trying to decrease the computational complexity of the generation procedure. 

The pros of rehearsal-based approaches are their simplicity and effectiveness. In fact, the methods with best performances in continual learning exploit exemplars as shown in this challenge review \citep{clworkshop2020} where the best approaches used exemplars. The drawback of rehearsal continual learning is the usage of a memory buffer, which can be saturated as the number of tasks to be learned grows. To overcome such drawback some methods propose the usage of representative exemplars \citep{DBLP:conf/icra/HayesCK19} and herding \citep{herding} techniques aimed to reduce the amount of memory required. Here, an interesting work (GDumb) proposed by \citep{gdumb} offers a simple baseline to rehearsal systems and questions the advancements of continual learning research itself due to its outstanding performance. Besides its performance, the system is very simple. In particular, the model samples data as experiences the stream of incoming task data. It does so until it fills a rehearsal buffer, by taking care to balance the proportion among classes. When the task data stream ends the dumb learner (a simple MLP or CNN) is trained only on the buffer data. GDumb achieves state-of-the-art performances.

\chapter{Works}
\label{chp:analyses}
\clearpage6
\section{Smaller is Better: An Analysis of Instance Quantity/Quality Trade-off in Rehearsal-based Continual Learning}
We begin our dissection by focusing on rehearsal-based methods i.e., solutions in where the learner exploits memory to revisit past \textbf{data}. Due to its prominet performance and the abrupt usage, rehearsal systems are nowadays one of the preferred countermeasures to fight catastrophic forgetting. 

So far, the focus from the community has been put into finding smart methodologies to improve the incremental performance. Instead, we ask ourselves what happens if we boost the capacity of the memory buffer. How much does impact altering the data storable in the memory? Indeed, in this study, we propose an analysis of the memory quantity/quality trade-off adopting various data reduction approaches to increase the number of instances storable in memory. By apply complex instance compression techniques to the original data, such as deep encoders, but also trivial approaches such as image resizing and linear dimensionality reduction, we offer a simple study on the trade-off. 

Then we introduce the usage of Random Projections as compression scheme and offer a simple pipeline through Extreme Learning Machines to resource-constrained continual learning, an appealing scenario where computational and memory resources are limited.

\definecolor{lightgray}{rgb}{0.83, 0.83, 0.83}
\newtheorem{theorem}{Lemma}

\newcommand{\gdumbC}{GDumb~\citep{gdumb} }
\newcommand{\gdumb}{GDumb }

\label{wrk:smaller_is_better}
Continual Learning (CL) is increasingly at the center of attention of the research community due to its promise of adapting to the  dynamically changing environment resulting from the huge increase in size and heterogeneity of data available to learning systems. It has found applications in several domains. Its prime application, and still most active field, is computer vision, and in particular object detection~\citep{gidaris2018dynamic, thrun1996learning,  review0}; however it has since found applications in several other domains such as segmentation~\citep{cermelli2020modeling, michieli2019incremental, DBLP:journals/corr/abs-2012-03362}, where each segmented class has to be learned in an incremental fashion, as well as in other fields, among which we mention Reinforcement Learning (RL)~\citep{ rl1, rl2} and Natural Language Processing (NLP)~\citep{nlp0, nlp1, nlp2}.

Ideally, the behaviour of CL systems should resemble human intelligence in its ability to incrementally learn in a dynamical environment~\citep{review1}, with minimal waste of resources, spatial or computational. The main problem encountered by these systems resides in the famous stability-plasticity dilemma of neuroscience, resulting in the so called {\em catastrophic forgetting}~\citep{MCCLOSKEY1989109}, a phenomenon where new information dislodges or corrupts previously learned knowledge, resulting in the deterioration of the ability to solve previously learned tasks.

Solutions to this problem typically incur in a increase in resource requirements \citep{clworkshop2020} both for CL's very nature (the more tasks arrive the more data the agent need to process), and for the nature of the systems that try to solve it, both in the increased complexity of the typically  deep learning  models, and in the time and space requirements of continuously learning multiple models. This problem become particularly evident in rehearsal-based methods.

Rehearsal-based methods, {\em i.e.}, approaches that leverage a memory buffer to cope with catastrophic forgetting, are emerging as the most effective methodology to tackle CL. Their performance, backed by extensive empirical evidence~\citep{clworkshop2020}, finds also a theoretical justification in Knoblauch and co-workers' finding that optimally solving CL would require perfect memory of the past~\citep{optimalcl}. In fact, if we were able to completely re-train a new system with all previous data every time a new task arrives, Continual Learning  would not appear to be any different from any other learning problem. However, this approach is both spatially and computationally infeasible for most real-world problems and we can argue it is precisely these memory and computational limitations that characterize CL and distinguish it from other learning problems.

\begin{figure*}[t]
    \centering
    \includegraphics[width=1\textwidth]{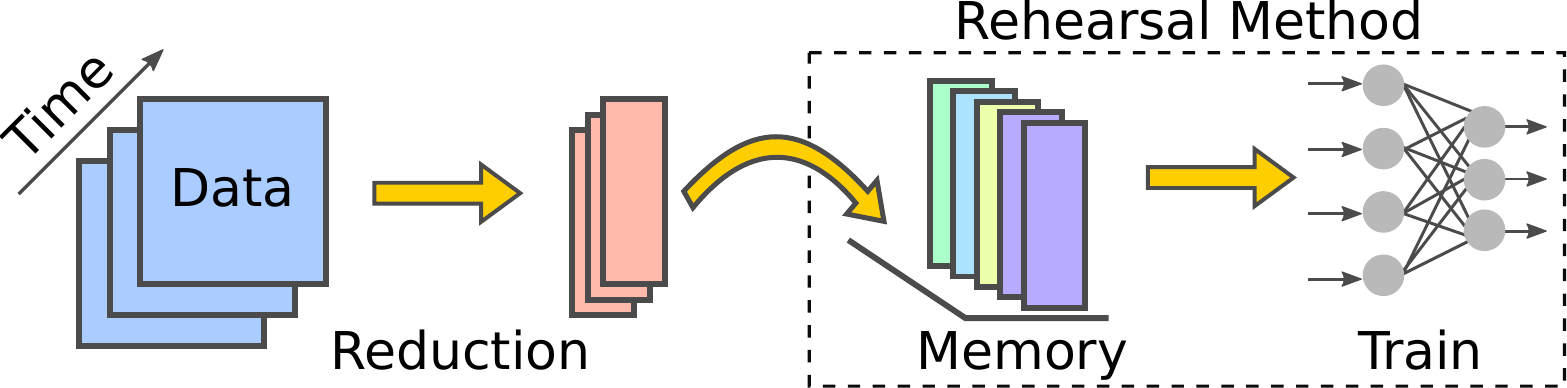}
    \caption{Our work analyzes the optimal instance quantity/quality trade-off in memory buffers of rehearsal-based Continual Learning systems. We carry out our analysis by applying several dimensionality reduction schemes to increase the quantity of storable data.}
    \label{fig:pipeline}
\end{figure*}

Our investigation aims to analyze the trade-offs on limited-memory CL systems. In particular, we focus on the quantity/quality trade-off for memory instances. We do so through the analysis of several dimensionality-reduction schemes applied to data instances that allows us to increase the number of examples storable in our fixed-capacity memory.
In particular we adopted deep learning encoders such as a variation of ResNet18~\citep{resnet18} and Variational Autoencoders (VAE)~\citep{DBLP:journals/corr/KingmaW13}, the simple yet surprisingly effective extreme resizing of image data, and, lastly, we explored Random Projections for dimensionality reduction. The latter scheme turns out to be very effective in low memory scenarios also reducing the model's parameter complexity. Indeed, we will show that a variation of Extreme Learning Machines (ELM) offers a simple yet effective solution for resources-constrained CL systems.

Our analysis will focus on computer vision tasks and use \gdumbC as a rehearsal-baseline. GDumb is a model that has been proposed to question the community's progress in CL thanks to the fact that in lieu of its outstanding simplicity, it was still able to provide state-of-the-art performance. Further, its simplicity also results in high versatility, as it proposes a general CL formulation comprising all task formulations in the literature. GDumb is fully rehearsal-based, and it is composed by a greedy sampler and a dumb learner, that is, the system does not introduce any particular strategy in the selection of replay data. Therefore, it represents the ideal candidate method to carry out our analysis. 

The experimental findings highlighted in this study are multiple: first, we show that when the memory buffer is fixed and extreme values of resizing of instance data is applied, we can easily push the state-of-the-art of CL rehearsal systems by a minimum of $+6\%$ to a maximum of $+67\%$ in terms of final accuracy. This surprising result suggests that the optimal trade-off between data quantity and quality is severely skewed toward the former and that in general the informational content required to correctly classify images in standard datasets is relatively low. Then, we analyze the consumption of resources of rehearsal CL systems as we saturate the rehearsal buffer, and show that ELM offer a clear solution on CL systems constrained by very low resources environments.

\begin{figure*}[t]
	\centering
	\includegraphics[width=0.9\textwidth]{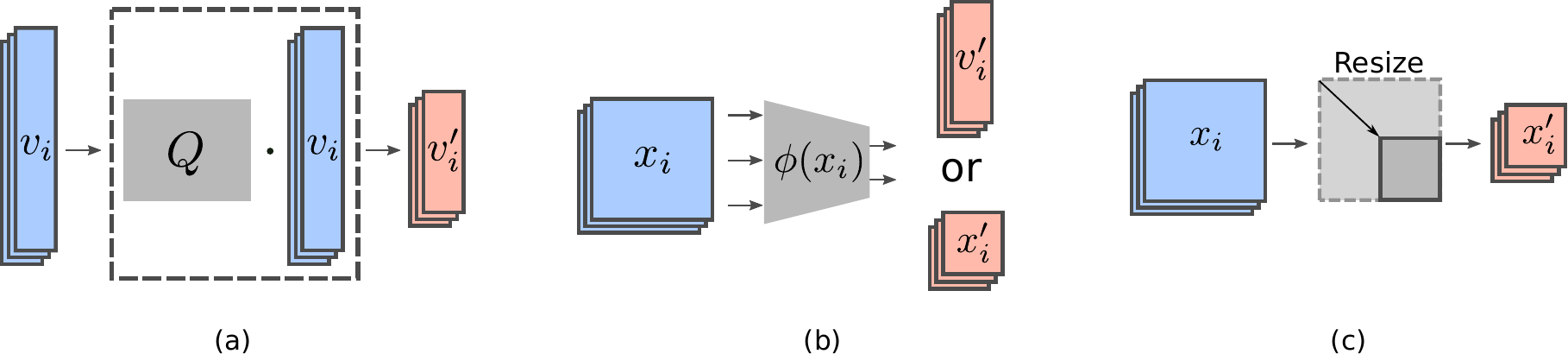}
	\caption{Depiction of the three main dimensionality reduction techniques analyzed. In  (a) random projection (RP) each image is vectorized ($v_i$) and then orthogonally-projected through a random matrix $Q$ into $v^\prime_i$. In (b), the encoder $\phi$ outputs a latent vector  $v_i^\prime$ (such as in VAEs) or a noise-free~/~shrinked image $x_i^\prime$ (as in CutR). In (c), we adopt a simple image resizing strategy through standard biliniear interpolation.}
	\label{fig:techniques}
\end{figure*}

%The field of Continual Learning is growing at fast pace, thanks to its appealing setting that aims to resemble some human learning aspects i.e. the ability to incrementally learn new tasks without forgetting past knowledge and the ability to learn in a non-stationary, dynamic environment \citep{hadsell_embracing_2020}. 

\subsection*{Related Works}

Following some recent surveys~\citep{review0, review1, review2}, we divide CL approaches into three main categories: regularization-based approaches, data rehearsal-based approaches and architectural-based approaches. Although a few novel theoretical frameworks based on meta-learning have been introduced recently~\citep{review1}, the majority still fall within these categories (or in a mixture of them).

\vspace{1em}
Regularization-based approaches address catastrophic forgetting by controlling each parameter's importance through the subsequent tasks, by means of the addition of a finely-tuned regularizing loss criterion. 
Elastic Weight Consolidation (EWC)~\citep{ewc} was the first well established approach of this class. It uses Fisher information to estimate each parameter's importance while discouraging the update for parameters with greatest task speciﬁcity. Learn without Forgetting (LwF)~\citep{DBLP:conf/eccv/LiH16} exploits the concept of ``knowledge distillation'' to preserve and regularize the output for old tasks. More recently,  Learning without Memorizing (LwM)~\citep{dhar2019learning} adds in the loss an information preserving penalty exploiting attention maps, Continual Bayesian Neural Networks (UCB)~\citep{ebrahimi2020} adapts the learning rate according to the uncertainty defined in the probability distribution of the weights in the network, while Pomponi \emph{et al.}~\citep{pomponi2020} propose a regularization of network's latent embeddings.

\paragraph{Rehearsal-based}
Rehearsal-based solutions allocate a memory buffer of a predefined size and devise some smart schemes to store previously used data to be replayed in the future, {\em i.e.}, to be added to future training samples. One of the first methodologies developed is Experience Replay (ER)~\citep{er}, which stores a small subset of previous samples and uses them to augment the incoming task-data.
Aljundi \emph{et al.}~\citep{ermir} propose an evolution of ER which takes  in consideration Maximal Interfered Retrieval (ER-MIR). Their proposal lies between rehearsal and regularization methods, its strategy is to retrieve the samples that are most interfered, \emph{i.e}. whose prediction will be most negatively impacted by the foreseen parameters update. Among other mixed approaches we have Rebuffi \emph{et al.}~\citep{icarl} that proposes a method which simultaneously learns strong classifiers and data representation (iCaRL). Gradient Episodic Memory (GEM)~\citep{gem} and its improved version Averaged-GEM (AGEM)~\citep{agem} exploits the memory buffer to constrain the parameter updates and stores the previous samples as trained points in the parameter space, while Gradient based Sample Selection (GSS) ~\citep{ermir} diversifies/prioritizes the gradient of the examples stored in the replay memory. Finally, a recent method proposed by Shim \emph{et al.}~\citep{shim2020online} scores memory data samples according to their ability to preserve latent decision boundaries (ASER). 

\paragraph{Architectural-based}
Architectural methods alter their parameter space for each task. The most influential architectural-based approach is arguably Progressive Networks (PN)~\citep{pnn}, where a dedicated network is instantiated for each task while Continual Learning with Adaptive Weights (CLAW)~\citep{adel2020} grows a network that adaptively identifies which parts to share between tasks in a data-driven approach. Note that, in general, the approaches that use incremental modules suffer the lack of task labels at test time, since there is no easy way to decide which module to adopt.

\subsection*{Method}
Before introducing the dimensionality reduction approaches adopted in our quantity/quality analysis we have to introduce the CL scenario considered and its task composition. Unfortunately the  community has not yet converged to a unique standard way to define a CL setting~\citep{van2019three}. Here we adopt GDumb's formulation which is the most general one and specifically resembles Lomonaco and Maltoni's formulation~\citep{core50}. In particular, we focus on the new class (NC)-type scenario~\citep{core50} where each task $T_i$ introduces data instances of $C_{T_i}$ new, previously unseen, classes.  More formally a dataset benchmark $\mathcal{D}$, containing examples from $C_{\mathcal{D}}$ classes, is divided into $n$ tasks. Each task, $T_{i}$ with $i=1 \dots n$, carries a set of examples $T_i = \{ \mathbf{X}_{T_i}, \mathbf{Y}_{T_i} \}$ whose class is previously unseen i.e. $\mathbf{Y}_{T_j} \cap \mathbf{Y}_{T_i} = \emptyset $ with $j = 1\dots i$ and $\mathbf{Y}_{T_i} = \{c_1 \dots c_{T_i}\}$. In other words, the model experiences a shift in the distribution of data as we train on each new task. We also consider the more realistic class incremental scenario (CI), that is, we are not allowed to know task labels at test time.

As incremental approach we use the recently proposed GDumb, which is composed of a simple learner and a greedy balancer. That is, given a fixed amount of memory $\mathcal{M}$, each instance of task data is randomly sampled in order to balance class instances in the memory, so that, at the end of the $T_i$ task experience, the memory contains an equal number of instances of all previously encountered classes i.e. each class has $\left \lfloor \frac{\mathcal{M}}{C_{\mathcal{D}} * i} \right \rfloor$ instances in memory.  

Besides providing state-of-the-art performances, GDumb has been proposed as standard baseline to question our progresses in continual learning research, since after experiencing a task, the simple learner (such as a ResNet18~\citep{resnet18} or a MLP) is trained \emph{only} with memory data, making GDumb a fully rehearsal based approach with random filtering of incoming data, and thus the ideal candidate to carry our study. In the following paragraphs, we briefly describe all the strategies adopted for dimensionality reduction.

%--------------------------------------------------
\subsubsection{Random Projections (RP)}
%\paragraph{Extreme Learning Machines (ELM)} 
Extreme Learning Machines (ELM)~\citep{huang_universal_2006} are a set of algorithms that exploit random projections as dimensionality reduction technique to preserve computational and spatial resources while learning. ELM  have been introduced in 2006 and recently have found application in neuroscience~\citep{elm_neuro0, elm_neuro2} and in other problems such as in molecular biology~\citep{elm_chem0}. The idea can be roughly described as a composition of two modules where the first one performs a random projection of the data, while the second one is a learning model. The appealing property of RP lies in the Johnson-Lindenstrauss lemma~\citep{Johnson1984ExtensionsOL} which states that given a set of points in a high dimensional plane, there is a linear map to a subspace that roughly preserves the distances between data points by some approximation factor.

%\begin{theorem}[Johnson-Lindenstrauss]
%Given an approximation factor $0<\varepsilon<1,$ a set of $K$ points living in $\mathbb{R}^{n}$, and a number $m>8 \frac{\ln (K)}{\varepsilon^{2}}$, there is a linear map $f: \mathbb{R}^{n} \rightarrow \mathbb{R}^{m}$ with $n > m$, such that:

%\begin{equation}
%(1-\varepsilon)\|u-v\|^{2} \leq\|f(u)-f(v)\|^{2} \leq(1+\varepsilon)\|u-v\|^{2}
%\end{equation}
%for all the points $u, v$ living in $\mathbb{R}^{n}$.
%\end{theorem}

The Johnson-Lindenstrauss lemma guarantees that we can obtain a low-distortion to the dimensionality reduction by multiplying each instance vector by a semi-orthogonal random matrix $Q^{m \times n}$ in the $(m,n)$ Stiefel manifold. More formally, let $x_i$ be an image of the current task of width, height and number of channels $w$, $h$, and $c$ respectively, then the size of $x_i$ is $n = h w c$.  We can consider its vectorization as $v_{i} \in \mathbb{R}^{n}$ and its compressed representation 

\begin{equation}
v_{i}^{\prime} = Q v_{i}  \hspace{15pt} \text{s.t.} \hspace{15pt} Q^{T} Q = I_{m}
\end{equation}
with $v_{i}^{\prime} \in \mathbb{R}^{m}$.

The usage of ELM unsuspectedly unlocks two main advantages: First it allows us to exploit the dimensionality reduction by \emph{increasing the number data instances} storable in the memory buffer. Secondly and, more importantly, allows us to \emph{use models with significantly fewer parameters}. On the other hand, the approach loses coordinate contiguity and, with that, shift co-variance, rendering convolutional approaches inapplicable.
%In fact, in the computer vision realm, where data points are images, through RP we can transform images into vectors. This allows us to omit the usage of all heavy architectures that rely on spatial information of data i.e. convolutional architectures, which are nowadays the standard weapon of choice for computer vision based problems, but that also require a huge amount of computational resources. 

After the random projection, data instances will be forwarded to the greedy sampler of GDumb to fill the memory $\mathcal{M}$. Then, we perform a rehearsal train with any MLP-like architecture, resulting in an order-of-magnitude reduction in the amount of parameters needed to process visual data allowing the usage of CL rehearsal based solutions in very low resource scenarios.

\begin{table}[t]
	\scriptsize
	\centering
	
\begin{tabular}{@{}cccc@{}}
\toprule
\multicolumn{4}{c}{\textbf{CIFAR10}} \\
\textit{Method} & \textit{Acc@600KiB} & \textit{Acc@1.5MiB} & \textit{Acc@3MiB} \\
\midrule
EWC \citep{ewc} & 17.9 $\pm$ 0.3 & 17.9 $\pm$ 0.3 & 17.9 $\pm$ 0.3 \\
GEM \citep{gem} & 16.8 $\pm$ 1.1 & 17.1 $\pm$ 1.0 & 17.5 $\pm$ 1.6 \\
AGEM \citep{agem}                       & 22.7 $\pm$ 1.8 & 22.7 $\pm$ 1.9 & 22.6 $\pm$ 0.7 \\ 
iCARL \citep{icarl}          & 28.6 $\pm$ 1.2 & 33.7 $\pm$ 1.6 & 32.4 $\pm$ 2.1 \\
ER \citep{er} & 27.5 $\pm$ 1.2 & 33.1 $\pm$ 1.7 & 41.3 $\pm$ 1.9 \\
ER-MIR \citep{ermir}        & 29.8 $\pm$ 1.1 & 40.0 $\pm$ 1.1 & 47.6 $\pm$ 1.1 \\
ER5    \citep{ermir}        & -              &  -             & 42.4 $\pm$ 1.1 \\
ER-MIR5 \citep{ermir}       & -              &  -             & 49.3 $\pm$ 0.1 \\
GSS \citep{gss}                         & 26.9 $\pm$ 1.2 & 30.7 $\pm$ 1.2 & 40.1 $\pm$ 1.4 \\
ASER \citep{shim2020online}             & 27.8 $\pm$ 1.0 & 36.2 $\pm$ 1.1 & 43.1 $\pm$ 1.2 \\
ASER$_\mu$ \citep{shim2020online}       & 26.4 $\pm$ 1.5 & 36.3 $\pm$ 1.2 & 43.5 $\pm$ 1.4 \\
\rowcolor{red!15}
GDumb \citep{gdumb}          & 35.0 $\pm$ 0.6 & 45.8 $\pm$ 0.9 & 61.3 $\pm$ 1.7 \\
\midrule
\rowcolor{orange!15}
Resize ($8\times8$) & \textbf{55.5 $\pm$ 0.2} & \textbf{64.5 $\pm$ 0.2} & \textbf{73.1 $\pm$ 0.2} \\
\rowcolor{blue!15}
ELM ($128$)    & \textbf{43.0 $\pm$ 0.3} & \textbf{47.1 $\pm$ 0.2} & 50.0 $\pm$ 0.2          \\
\rowcolor{green!15}
CutR ($8\times8$)   & \textbf{54.4 $\pm$ 0.2} & \textbf{60.9 $\pm$ 0.2} & \textbf{71.6 $\pm$ 0.6} \\ 
\bottomrule
\end{tabular}

	\vspace{1em}
	\caption{CIFAR10 experiments (5 runs)}
	\label{tbl:cifar10}
	
\end{table}

%--------------------------------------------------

\subsubsection{Deep Encoders} 
Deep encoders are neural models $\phi$ that take as input an image $x_{i}$ and, depending from the structure of such model, can output either a latent vectorial representation $v_{i}^{\prime}$, or a squared feature map which we consider as a noise-free shrinked image $x_i^\prime$. Figure \ref{fig:techniques} (b) reports visually the two possible encoding scenarios. In this work, we adopt a Variational AutoEncoder (VAE)~\citep{DBLP:journals/corr/KingmaW13} for the first case and a pretrained ResNet18~\citep{resnet18} cut up to a predefined block (CutR) as a prototype for the second.

\paragraph{VAE} Variational Autoencoders~\citep{DBLP:journals/corr/KingmaW13} have been introduced as an efficient approximation of the posterior for arbitrary probabilistic models. A VAE is essentially an autoencoder that is trained with a reconstruction error between the input and decoded data, with a surplus loss that constitutes a variational objective term attempting to impose a normal latent space distribution. The variational loss is typically computed through a Kullback-Leibler divergence between the latent space distribution and the standard Gaussian, the total loss can be summarized as follows:
\begin{equation}
\mathcal{L}=\mathcal{L}_{r}(x_{i},\hat{x_{i}}) +\mathcal{L}_{KL}(q(z_{i}|x_{i}), p(z_{i}))
\end{equation}

given an input data image $x_{i}$, the conditional distribution $q(z_{i}|x_{i})$ of the encoder, the standard Gaussian distribution $p(z_{i})$, and the reconstructed data $\hat{x_{i}}$. We use the encoding part of a VAE pretrained on a dataset by feeding each incoming image and retrieving the vectorial output representation $v_i^\prime$, then the data point is forwarded to GDumb's greedy sampler to feed $\mathcal{M}$. 

\paragraph{CutR} As our second encoding approach, we use a pretrained ResNet18~\citep{resnet18} cut up to a predefined block.
ResNets models are Convolutional Neural Networks (CNNs) introducing skip connections between convolutional blocks to alleviate the so called vanishing gradient~\citep{hochreiter1998vanishing} problem afflicting deep architectures. 
The idea behind it, is to use the cut ResNet18 as a \emph{filtering module} that outputs a smaller feature map, giving us $x_{i}^{\prime}$. In fact, we cut the network towards later blocks, since neurons in the last layers, encode more structured semantics with respect to the early ones~\citep{olah2017feature}. Therefore, we are able to extract semantic knowledge from unseen images leveraging transfer learning~\citep{tan2018survey}, that is, we exploit the ability of a model to generalize over unseed data. We refer to this method with the name CutR(esnet18). We use CutR instance encoding by feeding each image belonging to the current task and retrieving the shrinked output $x_i^\prime$ which is then forwarded to the greedy sampler module of GDumb to fill the memory $\mathcal{M}$.

In our analysis, we adopted the less resource-hungry VAE scheme  for datasets where shift co-variance is not as important, such as the MNIST, in which the digits are centered in the image and thus most approaches at the state-of-the-art use a MLP as classifier. In all other instances, we used the CutR scheme.
\begin{table}[t]
	\scriptsize
	\centering
	
\begin{tabular}{@{}ccccc@{}}
\toprule
\multicolumn{1}{c|}{} & \multicolumn{2}{c|}{\textbf{ImageNet100}} & \multicolumn{2}{c|}{\textbf{CIFAR100}} \\
\textit{Method} & \textit{Acc@12MiB} & \textit{Acc@24MiB} & \textit{Acc@3MiB} & \textit{Acc@6MiB} \\
\midrule
AGEM \citep{agem} & 7.0 $\pm$ 0.4 & 7.1 $\pm$ 0.5 & 9.05 $\pm$ 0.4 & 9.3 $\pm$ 0.4 \\
ER \citep{er} & 8.7 $\pm$ 0.4 & 11.8 $\pm$ 0.9 & 11.02 $\pm$ 0.4 & 14.6 $\pm$ 0.4 \\
EWC \citep{ewc} & 3.2 $\pm$ 0.3 & 3.1 $\pm$ 0.3 & 4.8 $\pm$ 0.2 & 4.8 $\pm$ 0.2 \\
GSS  \citep{gss} & 7.5 $\pm$ 0.5 & 10.7 $\pm$ 0.8 & 9.3 $\pm$ 0.2 & 10.9 $\pm$ 0.3 \\
ER-MIR \citep{ermir} & 8.1 $\pm$ 0.3 & 11.2 $\pm$ 0.7 & 11.2 $\pm$ 0.3 & 14.1 $\pm$ 0.2 \\
ASER \citep{shim2020online} & 11.7 $\pm$ 0.7 & 14.4 $\pm$ 0.4 & 12.3 $\pm$ 0.4 & 14.7 $\pm$ 0.7 \\
ASER$_{\mu}$ \citep{shim2020online} & 12.2 $\pm$ 0.8 & 14.8 $\pm$ 1.1 & 14.0 $\pm$ 0.4 & 17.2 $\pm$ 0.5 \\
\rowcolor{red!20}
GDumb \citep{gdumb} & 13.0 $\pm$ 0.3 & 21.6 $\pm$ 0.3 & 17.1 $\pm$ 0.2 & 25.7 $\pm$ 0.7 \\
\bottomrule
\rowcolor{orange!15}
Resize ($8\times8$) & \textbf{33.6 $\pm$ 0.2} & \textbf{33.6 $\pm$ 0.3} & \textbf{38.5 $\pm$ 0.4} & \textbf{45.1 $\pm$ 0.2} \\
\rowcolor{blue!15}
ELM ($128$) & \textbf{13.3 $\pm$ 0.2} & 15.4 $\pm$ 0.4 & \textbf{22.4 $\pm$ 0.3} & \textbf{25.7 $\pm$ 0.3} \\
\rowcolor{green!15}
%CutR (8x8) & \textbf{36.1 $\pm$ 0.3} & \textbf{36.0 $\pm$ 0.3} & \textbf{32.6 $\pm$ 0.6} & \textbf{37.1 $\pm$ 0.2} \\ 
CutR ($8\times8$) & \textbf{36.25 $\pm$ 0.4*} & \textbf{36.27 $\pm$ 0.5*} & \textbf{32.6 $\pm$ 0.6} & \textbf{37.1 $\pm$ 0.2} \\
\bottomrule
\end{tabular}

	\vspace{1em}
	\caption{ImageNet and CIFAR100 experiments (5 runs)}
	\label{tbl:cifarimagenet}
	
\end{table}

\subsubsection{Resizing}
We used also the simplest instance reduction approach one can think of {\em i.e.},  resizing  the images to very low resolution through standard bilinear interpolation. The resized images are then fed to the sampler of GDumb to balance the classes in $\mathcal{M}$ and all  training and prediction is performed on the  lowered resolution images.

Independently of the approach adopted, all data instances are reduced before storing them in memory $\mathcal{M}$, then we use GDumb's greedy sampler to select and balance class instances, and finally, we use a suitable learner to fit memory data and assess the performance. 
In general, following GDumb, we adopt ResNet18 for large-scale image classification tasks for all approaches that maintain shift co-variance, reverting to a simple MLP for approaches without shift co-variance like RP.

\subsection*{Experiments}
%We now introduce the experimental settings, the datasets used, and the learning scenarios for our experimental evaluation. 
%We analyzed 3 dimensionality reduction schemes that allow to increase the quantity of data instances storable in memory meanwhile measuring GDumb's . 
We performed our analysis on the following  standard  benchmarks:
\begin{itemize}
	\item MNIST~\citep{mnist}: the dataset is composed by 70000 $28\times28$ grayscale images of handwritten digits divided into 60000 training  and 10000 test images belonging to 10 classes.
	\item CIFAR10~\citep{cifar}: consists of 60000 RGB images of objects and animals. The size of each image is $32\times32$ divided in $10$ classes, with 6000 images per class. The dataset is split into 50000 training images and 10000 test images. 
	\item CIFAR100~\citep{cifar}: is composed by 60000, $32\times32$ RGB images subdivided in 100 classess with 600 images each. The dataset is split into 60000 training images and 10000 test images. 
	\item ImageNet100~\citep{imagenet}: the dataset is composed of $64\times64$ RGB images divided in 100 classes; it is composed of  60000 images split into 50000 training and 10000 test. 
	\item Core50~\citep{core50}: the dataset is composed of $128\times128$ RGB images of domestic objects divided in 50 classes. The set consists of 164866 images split into  115366  training and 49500 test. 
\end{itemize}

Following~\citep{gdumb}, we use final accuracy as the evaluation metric throughout the work.
The metric is computed \textit{at the end of all tasks} against a test set of never seen before images  composed of an equal number of instances per  class. This allows us to directly compare against the largest number of competitors in the literature.

%some settings statistics such as the maximum amount of instances of images that can be saved in the memory if no instance compression is applied, and variation as we vary the memory buffer size. Then, we also report the CL task subdivision for each dataset benchmark. 

All the experiments has been conducted with an Intel i7-4790K CPU with 32GB RAM and a 4GB GeForce GTX 980 machine running \texttt{PyTorch 1.8.1+cu102}.

\subsubsection{Parameter Sensitivity}

In the first experiment, we compared different dimensionality reduction strategies as we altered the parameters. The analysis was conducted on three different datasets: MNIST, CIFAR10 and ImageNet100.  In this evaluation we fixed the amount of memory buffer used for GDumb during rehearsal training, and we measured the final accuracy as the parameters varied for each dimensionality reduction method. In particular we subdivided both MNIST and CIFAR10 datasets into 5 tasks of 2 classes each,  with 600 KiB dedicated memory buffer, while ImageNet100 was divided into 10 tasks of 10 classes each, with 12 MiB memory buffer.  

\begin{figure*}[t]
	\centering
	\includegraphics[width=\textwidth]{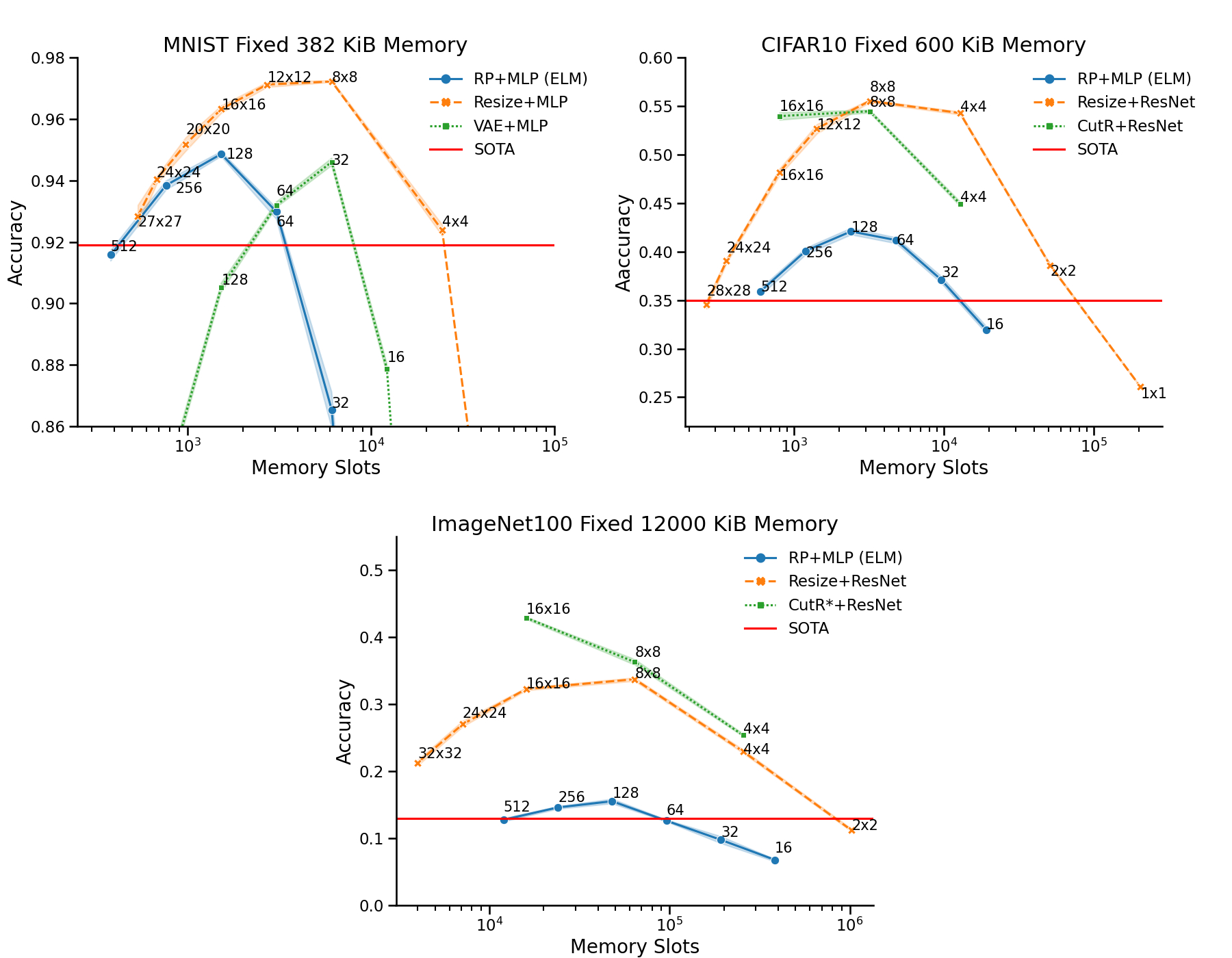}
	\caption{At top-left the accuracy analysis of the MNIST dataset. In top-right we have the analysis of CIFAR10 and at bottom we have ImageNet100. The state-of-the-art (SOTA) method is plain GDumb with an MLP as incremental learner in the MNIST experiment and Resnet18 in the others. The number of instances in memory (\emph{i.e.} the $x$ axis) is in $log$ scale. We report the results of (5 runs).}
	\label{fig:optimal}
\end{figure*}

Figure~\ref{fig:optimal} plots the performance of the various  schemes as we reduce the dimensionality of the instances and and thus increase their number in the allocated memory.  
The orange line represents the performance of the resize scheme. For the MNIST dataset, we considered eight different target sizes\footnote{throughout the work we omit to write the channel component for brevity} $x_{i}^{\prime} \in \{27\times27, 24\times24, 20\times20, 16\times16, 12\times12, 8\times8, 4\times4, 2\times2, 1\times1\}$. We performed the same resizing for CIFAR10 data. We did not report CIFAR100 analysis since the data format is the same as CIFAR10 and the result would be analogous. For ImageNet100, we resized each instance to $x_{i}^{\prime} \in \{32\times32, 24\times24, 16\times16, 6\times6, 4\times4, 2\times2\}$. 

The green line of Figure~\ref{fig:optimal} represents the  deep encoders. In particular, for MNIST  we used a VAE \citep{DBLP:journals/corr/KingmaW13} pretrained on KMNIST \citep{clanuwat2018deep} and analyzed the performance of GDumb with compressed instances as we altered the size of the latent embedding vector to  $v_{i}^{\prime} \in \{128, 64, 32, 16\}$. On the other hand, for the CIFAR10  and ImageNet100 dataset we considered different parameters for CutR. In particular, we cut the ResNet18 up to the sixth layer to get a $4\times4$ output, to the fifth to have a $8\times8$ encoding, and lastly up to the third block to get a $16\times16$ feature map. 

The CutR Resnet18 has been pretrained on the complete ImageNet, thus the results in the ImageNet100 benchmark can be biased. We denote these biased results with CutR\textbf{*}.

Lastly, the blue line of Figure~\ref{fig:optimal} reports the accuracy of  Random Projection followed by an MLP classifier. We recall that this kind of architecture is a variation of an Extreme Learning Machine (ELM), therefore we will refer to it with the term ELM. We analyzed the final accuracy as the size of the random projection changes, in particular the embedding sizes considered are $v_{i}^{\prime} \in \{512, 256, 128, 64, 32, 16\}$ for all the datasets.

For all the experiments in MNIST data, we used a 2-layer MLP with 400 hidden nodes as learning module, while we used a Resnet18~\citep{resnet18} for all the other analysis with exception of ELM scheme that maintains the 2-layer MLP model throughout. We did not perform any hyperparameter tuning on the learning module in accordance with the \gdumbC experimental protocol.   For completeness we report the learning parameters: the system uses an SGD optimizer, a fixed batch size of $16$, learning rates $[0.05,0.0005]$, an SGDR~\citep{DBLP:conf/iclr/LoshchilovH17} schedule with $T_{0}= 1$, $T_{mult}= 2$ and warm start of 1 epoch. Early stopping with patience of 1 cycle of SGDR, along with standard data augmentation is used (normalization of data). GDumb uses cutmix~\citep{DBLP:conf/iccv/YunHCOYC19} with $p=0.5$ and $\alpha=1.0$ for regularization on all datasets except MNIST.

As we can also see from Figure~\ref{fig:optimal} all the strategies considered unlock performance greatly above \gdumb, thus suggesting that the quantity/quality trade-off is severely skewed toward quantity since each dimensionality reduction technique greatly improves the amount of data instances that can be stored in the memory buffer. It is also evident that the simple resizing strategy gives the best performance improving \gdumb by $+6\%$ on MNIST and roughly by $+20\%$ on both CIFAR10 and ImageNet100 datasets.

\begin{table}[t]
	\scriptsize
	\centering
	
\begin{tabular}{@{}cc@{}}
\toprule
\multicolumn{2}{c}{\textbf{MNIST}} \\
\textit{Method} & \textit{Acc@382KiB} \\ \midrule
GEN \citep{DBLP:journals/corr/abs-1810-12488} & 75.5 $\pm$ 1.3 \\
GEN-MIR \citep{ermir} & 81.6 $\pm$ 0.9 \\
ER \citep{er} & 82.1 $\pm$ 1.5 \\
GEM \citep{gem} & 86.3 $\pm$ 1.4 \\
ER-MIR \citep{ermir} & 87.6 $\pm$ 0.7 \\
\rowcolor{red!15}
GDumb \citep{gdumb} & 91.9 $\pm$ 0.5 \\
\bottomrule
\rowcolor{orange!15}
Resize ($8\times8$) & \textbf{97.2 $\pm$ 0.1} \\
\rowcolor{blue!15}
ELM ($128$) & \textbf{95.0 $\pm$ 0.4} \\
\rowcolor{green!15}
VAE ($32$) & \textbf{94.6 $\pm$ 0.1}
\\ \bottomrule
\end{tabular}

	\vspace{1em}
	\caption{MNIST final accuracy (5 runs) analysis as we vary the memory for all schemes considered.}
	\label{tbl:mnist}
	
\end{table}

\begin{figure*}[t]
	\centering
	\includegraphics[width=\textwidth]{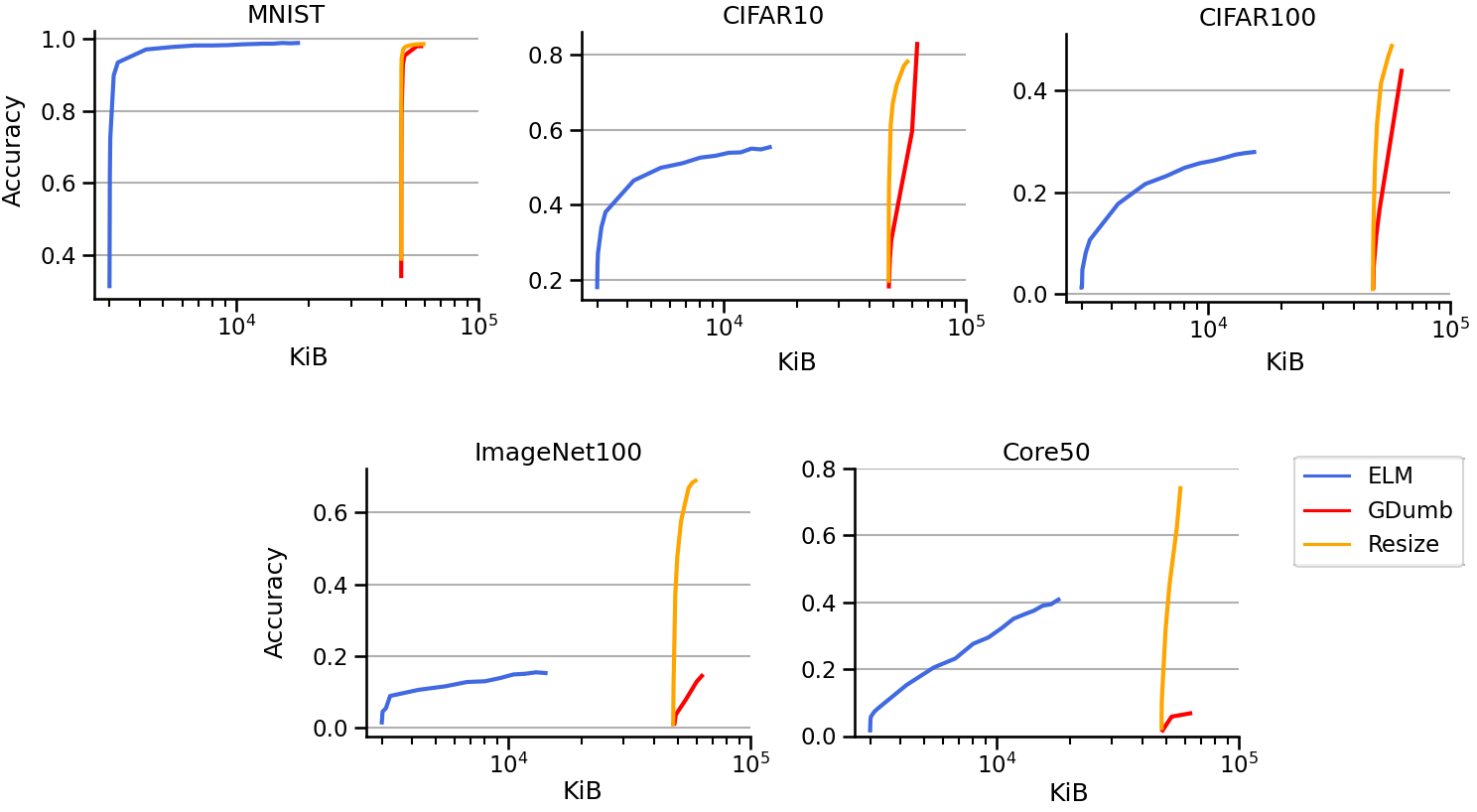} 
	\caption{We show the total amount of KiB used by the whole CL system. We measure the consumption as we saturate the rehearsal memory plus the storage of model parameters. The $x-$axis is in $log$ scale.}
	\label{fig:saturation}
\end{figure*}

Moreover, we chose to consider extreme levels of encoding. We did so to find the level of compression that irreversibly corrupts spatial information and thus makes learning impossible. Surprisingly, it turns out that a $2\times2$ resizing still works on CIFAR10 data with perfomances above \gdumb while a $1\times1$ resize is still better than a random classifier whose performance would be $20\%$ of final accuracy. 
This is a strong evidence that the amount of data storable in the memory buffer plays a central role, but also that CIFAR10 dataset constitutes an unrealistic benchmark and should not been considered to assess novel methodologies in the future.

After choosing and fixing the optimal parameters for each compression scheme, we study the performance of the rehearsal system as we alter the quantity of the memory allocated. In Tables~\ref{tbl:mnist},\ref{tbl:cifarimagenet} we compute the final accuracy for all the datasets previously considered, with the addition of CIFAR100 with an increase of 20\% in performance. The amount of dedicated memory for the rehearsal buffer, has been chosen in order to be consistent with several other methods at \gdumb, allowing us to compare GDumb's performance on optimized memory schemes against other methods. As we can see, all memory optimizations still provide huge advantages as the memory buffer varies, suggesting again, that instance quantity plays a fundamental role in rehearsal systems even with extreme encoding settings.

Finally, we note that the deep models used for classification have a large number of degrees of freedom and require a large amount of instances to be properly trained to capture the complexity of the task at hand. Simpler, lower dimensionality instances allow both for more instances and simpler classifiers with fewer parameters without losing  lot of informational content.

\subsubsection{Resource Consumption}

With the second experiment, we wanted to analyze the performance versus the total memory requirement for each approach. Here, we increased the number of instances in the memory buffer and added to the total consumption the working memory used by the classifier to store (and train) the parameters.
 
We considered three different scenarios: first we used the plain GDumb CL system without dimensionality reduction (representing \gdumb), then we used ELM (with fixed embedding size of ($v_{i}^\prime = 128$), and lastly the resizing scheme (images resized to $x_{i}^\prime = 8\times8$). We selected the best parameters resulting from the previous experiment. 

We then assessed the performance and resource usage using a new dataset, namely the Core50~\citep{core50}. The reason behind the use of Core50 to validate our findings is twofold: first, we test again whether the quantity of extremely encoded data plays a central role on our rehearsal scheme. Secondly, we measure the performance and the resource usage of a CL system on a more complex set of tasks. We divided the dataset into 10 tasks of 5 classes each.

In Figure~\ref{fig:saturation}, we report the results of this experiment. We can see that extreme levels of resizing still provide optimal results in all the datasets considered. One striking finding is that in Core50 with extreme resizing, even if the size was not optimized for the dataset, the final accuracy is increased by $+67\%$ with respect to \gdumb. Second, we note that ELM constitute a viable solution in low resources scenarios. Indeed, we can surpass the performance of \gdumb for low memory scenarios where  even just the classifier used in other approaches could not fit in the allocated memory, much less the rehearsal buffer.
This is clearly observed from the Core50 results. 
We can appreciate that by randomly projecting image data and learning in a low resource scenario provides a boost of $+34\%$ in the final accuracy.

Finally, it is worth noting there is a striking dissonance in the literature of rehersal-based method when the narrative around  buffer-memory sizes revolves around decisions among sizes of the order of 300KiB to 600KiB when then the same systems adopt complex classifiers using several megabytes of memory just for the learned parameters and in the order of gigabytes of working memory for learning. In a real constrained-memory scenario a simpler classifier with more instances offers a clear advantage.

\subsection*{Conclusion}
In this study, we analyzed the quantity/quality trade-off in rehearsal-based Continual Learning systems adopting 
several dimensionality reduction schemes to increase the number of instances in memory at the cost of possible loss in information. In particular, we used deep encoders, random projections, and a simple resizing scheme. What we found is that even simple, but extremely compressed encodings of instance data provide a notable boost in performance with respect to the state of the art, suggesting that in order to cope with catastrophic forgetting, the optimization of the memory buffer can play a central role. Notably, the performance boost of extreme instance compression suggests that the quality/quantity trade-off is severely biased toward data quantity over data quality.
We suspect that some fault might be in the overly simplistic datasets adopted by the community, but mostly the deep models used for classification are well known to be data-hungry and the instances stored are not sufficient to properly train them, but can suffice for simpler classifiers with fewer parameters working on simplified instances.

It is worth noting there is a striking dissonance in the literature of rehearsal-based method. The narrative on buffer-memory sizes revolves around decisions among sizes of the order of 300KiB to 600KiB when then the same systems adopt complex classifiers using several megabytes of memory just for the learned parameters and in the order of gigabytes of working memory for training. In a real constrained-memory scenario, a simpler classifier with more instances offers a clear advantage.

In fact, in a real low-resources scenario deep convolutional systems using several megabytes of memory for the model parameters and gigabytes of working memory for learning are not a viable solution. In this case,  a variation of Extreme Learning Machines offer a simple and effective solution.

\section*{Other Experiments}

\subsection*{Fixed Data Instances}

\begin{figure}[h]
	\centering
	\includegraphics[width=1\textwidth]{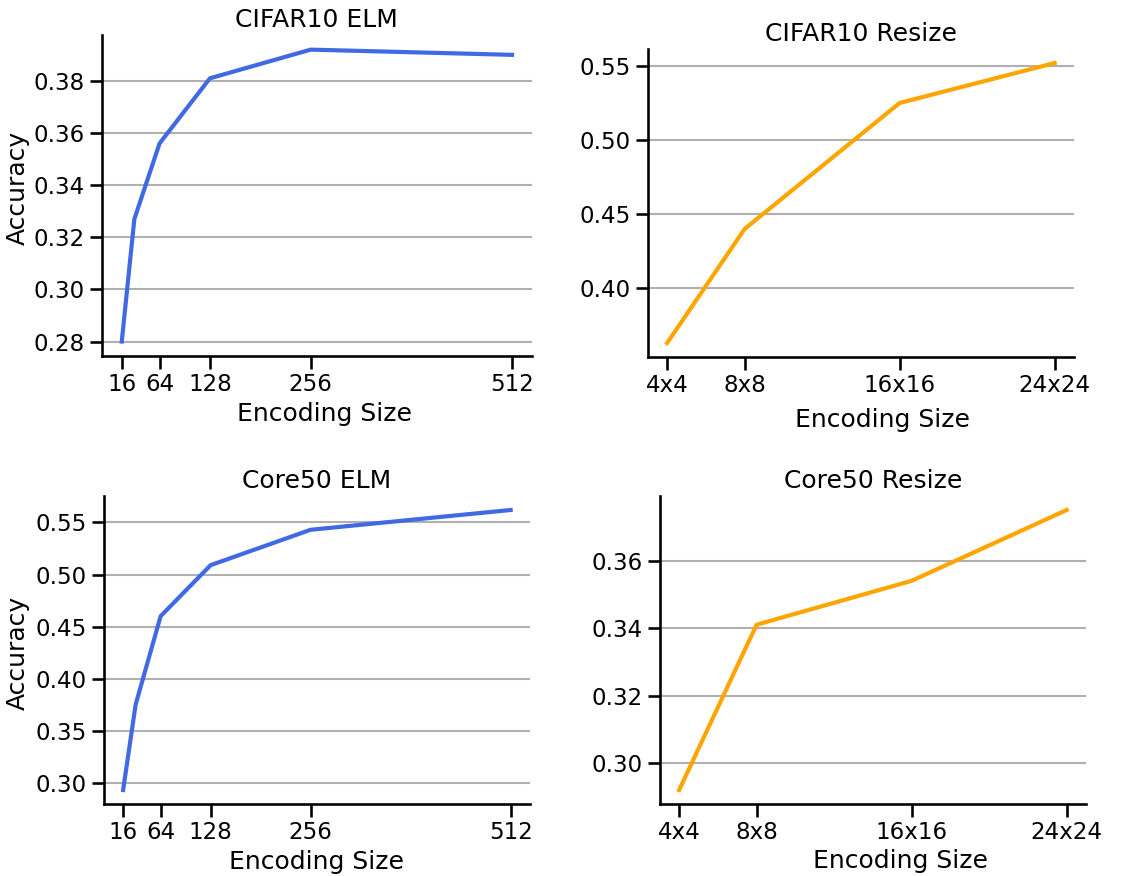}
	\caption{Performance as we vary the parameters for each scheme on CIFAR10 and Core50. In the former benchmark, the memory buffer is of 1000 fixed instances, while in the latter is of 8000.}
	\label{fig:instance}
\end{figure}

With this experiment we aim to better show that instance quantity is preferable over instance quality. We fixed the number of data slots in the memory buffer, and we analyzed the performance as we alter the encoding size. In particular, we tested two datasets namely CIFAR10 and Core50. In CIFAR10 we fixed the buffer to 1000 data slots, while in the latter benchmark we fixed it to be 8000 slots. What we can see from Figure~\ref{fig:instance} is that the improvement of performance is not given by the encoding's smoothing property, and, again, we confirm that rehearsal systems are skewed towards data quantity.

\newpage

\subsection*{ELM Width Analysis}
\begin{figure}[h]
	\centering
	\includegraphics[width=0.7\textwidth]{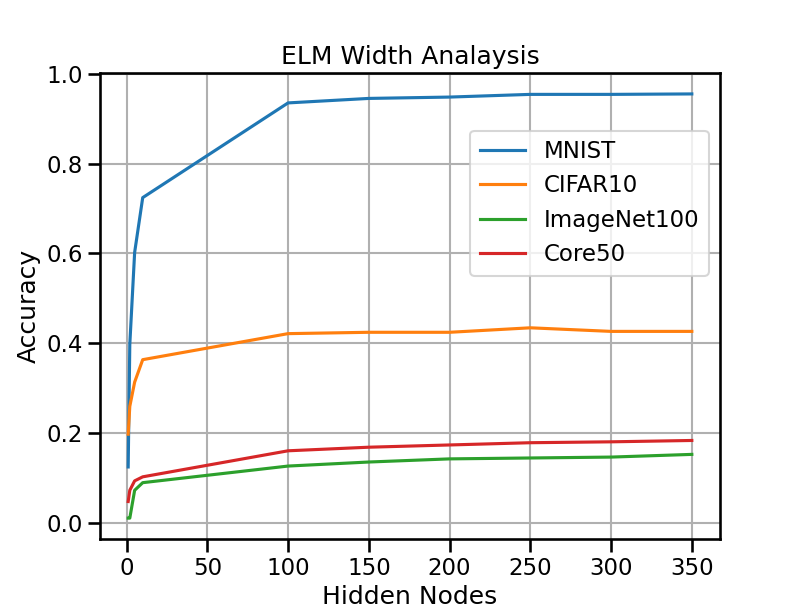}
	\caption{Analysis of final accuracy as we alter the number of hidden units in ELM.}
	\label{fig:width}
\end{figure}

% troppi pochi dati per apprendere un modello complesso
As we specified in the work, we used a variation of an Extreme Learning Machine. In particular, the architecture is composed by a random projection module and a learning module. The first is implemented through an orthogonal random matrix. While the second is a two layer MLP. Throughout the study we used 400 hidden units as last layer before the output. We choose to do so to be consistent with GDumb experimental settings. With this experiment we analyze the accuracy metric as we change the number of hidden units. We fixed the encoded size of data to be $v_{i}^{\prime}128$. As memory buffer, we used a different number of data slots for different datasets. That is, for MNIST and CIFAR10 we adopted 2400 slots (600 KiB), in ImageNet100 we used 48000 instances i.e. 12 MiB, while for Core50 we used 8000 slots (2 MiB). In Figure~\ref{fig:width} we can see that 100 hidden units are sufficient to achieve the maximum performance. This, again, shows that more deep classifiers which are common in CL rehearsal literature, might need more data to be trained properly.  

\subsection*{Experiments with other Rehearsal Systems}
Throughout our study, we used GDumb to carry out our analysis. Although we extensively motivated this choice, we also tested two different rehearsal systems. In particular we studied ER~\cite{er} and ER-MIR~\cite{ermir} performance as we adapt them to work in a low resource scenario. We simply substitute the original learner with our ELM proposal. In Table~\ref{tbl:er} we report the performance of CIFAR10 with 600 KiB buffer memory and $v_{i}^{\prime} = 128$ encoding. As validation metrics we used the final accuracy and the average forgetting~\cite{chaudhry2018riemannian} (lower is better). In order to train the systems, we used the official implementations found at \href{https://github.com/optimass/Maximally_Interfered_Retrieval}{https://github.com/optimass/Maximally\_Interfered\_Retrieval} without any alteration of training hyperparameters. 
As we can see, the results suggest again that ELMs constitute a valid solution for low resource CL systems and that rehearsal solutions are biased toward data quantity over data quality.

\begin{table}[h]
	\centering
	\scriptsize
	\begin{tabular}{@{}ccccc@{}}
\toprule
\multicolumn{5}{c}{\textbf{CIFAR10 Fixed Memory 600 KiB}}                    \\
\textit{Method}  & \textit{Accuracy (A)}  & \textit{Forgetting (F)} & \textit{ELM (A)} & \textit{ELM (F)}  \\ \midrule
ER \cite{er}  & 27.5 $\pm$ 1.20  & 48.0 $\pm$ 0.40 & \textbf{42.0 $\pm$ 0.10} &  \textbf{41.2 $\pm$ 0.16} \\
ER-MIR \cite{ermir} & 29.8 $\pm$ 1.10  & 44.6 $\pm$ 0.48 & \textbf{45.6 $\pm$ 0.10} & \textbf{31.6 $\pm$ 0.01} \\ 
\bottomrule
\end{tabular}

	\vspace{10pt}
	\caption{Experiments in CIFAR10 with two different rehearsal systems in low resource scenario. }
	\label{tbl:er}
\end{table}

%%%%%%%%%%%%%%%%%%%%%%%%%%%%%%%%%%%%%%%
%%%%%%%%%%%%%%%%%%%%%%%%%%%%%%%%%%%%%%%
\section*{Other Specifications}

\subsection*{Resource Consumption}

In Table~\ref{tab:summary} we report some summary statistics. In particular, we report GDumb's performance improvements for two encoding schemes i.e. Resize ($8 \times 8$) and ELM ($v_{i}^{\prime}= 128$). We reported only the accuracy according to optimal parameters. We also added the compression factor $\mathcal{C}$, the requirements to store model's parameters $\Theta$ and the memory buffer $\mathcal{M}$. We also report the quantity of GPU memory usage to train GDumb for each encoding scheme. We can see that there is a big gap on the training requirements and memory buffers.  

\vspace{20pt}

\begin{table}[h]
	\tiny
	\centering
% \usepackage{booktabs}

\begin{tabular}{@{}ccccccccc@{}}
\toprule
\multicolumn{1}{l}{}  & \textbf{MNIST} & \textbf{CIFAR10} & \textbf{CIFAR100} & \textbf{ImageNet100} & \textbf{Core50} & \textbf{Compression} & \textbf{Params +  $\mathcal{M}$} & \textbf{GPU Training} \\ \midrule
\textbf{Resize ($8\times8$)} & (+6\%)         & (+21\%)          & (+20\%)           & (+20\%)              & (+67\%)         & 253:1                & 60 MiB                            & 2.2 GiB           \\
\textbf{ELM ($128$)}    & (+10\%)        & (+10\%)          & (+10\%)           & (+10\%)              & (+10\%)         & 192:1                & 16 MiB                            & 0.72 GiB          \\ \bottomrule
\end{tabular}

	\vspace{10pt}
	\caption{Performance summary and memory compression}
	\label{tab:summary}
\end{table}

\subsection*{Datasets Specification}

For completeness, we report in Table~\ref{tab:memstats} some specifications for the considered datasets. In particular, we provide the task subdivision for each dataset. As we can see MNIST and CIFAR10 have been split in 5 tasks of 2 classes each. This splitting is also known in literature as Split-CIFAR10 and Split-MNIST. For CIFAR10 and ImageNet100 benchmarks we used 10 tasks of 10 classes each, meanwhile for Core50 we shuffled all scenarios and created 10 tasks of 5 classes each. The majority of the works fix the memory slots to define the memory buffer. In our case we used memory requirements expressed in KiB or MiB so that we could alter each slot consumption. We provide a correspondence between memory requirements and  memory slots in the case we consider original image sizes, we do so to ease future comparisons against our work.

\vspace{20pt}

\begin{table}[h]
	\centering
	\scriptsize
	
\begin{tabular}{@{}ccccc@{}}
\toprule
\multicolumn{5}{c}{\textbf{Experimental Settings}} \\
\textit{Dataset} & \textit{Image size} & \textit{Memory Size} & \textit{\# Instances}  & \textit{Task Composition} \\ \midrule
MNIST & 28x28x1 & 382 KiB & 500 & 5 tasks, 2 classes \\ \midrule
CIFAR10 & 32x32x3 & 600 KiB & 200 & 5 tasks, 2 classes \\
  & & 1.5 MiB & 500 & \\
  & & 3 MiB & 1000 & \\ 
  & & 6 MiB & 2000 & \\
CIFAR100 & - & - & - & 10 tasks, 10 classes \\ \midrule
ImageNet100 & 64x64x3 & 12 MiB & 1000 & 10 tasks, 10 classes\\
  & & 24 MiB & 2000 & \\ \midrule 
Core50 & 128x128x3 & 15 MiB & 312 & 10 tasks, 5 classes \\  
  
  \bottomrule
\end{tabular}

	\vspace{10pt}
	\caption{Dataset and memory statistics, in CIFAR100 row we omit the 2nd, 3rd and 4th columns since are equal to CIFAR10 row.}
	\label{tab:memstats}
\end{table}

\section{Towards Exemplar-Free Continual Learning in Vision Transformers: an Account of Attention, Functional and Weight Regularization}

While in the previous work we considered old data points as pivotal instrument to investigate catastrophic forgetting, now we focus on the \textbf{structural} properties of the model considered. In particular, we ask ourselves how some parts of a network, when properly regularized, impact to the overall performance of an incremental scenario. We decided to investigate the continual learning of Vision Transformers (ViT) for the challenging exemplar-free scenario. We opted to study ViTs since there are several works tackling CNNs while virtually no one focused to ViTs yet although they are getting consistently better at vision tasks.

This work takes an initial step towards a surgical investigation of the self attention mechanism (SAM) for designing coherent continual learning methods in ViTs. We first carry out an evaluation of established continual learning regularization techniques. We then examine the effect of regularization when applied to two key enablers of SAM: (a) the contextualized embedding layers, for their ability to capture well-scaled representations with respect to the values, and (b) the prescaled attention maps, for carrying value-independent global contextual information. We depict the perks of each distilling strategy on two image recognition benchmarks (CIFAR100 and ImageNet-32) -- while (a) leads to a better overall accuracy, (b) helps enhance the rigidity by maintaining competitive performances. Furthermore, we identify the limitation imposed by the symmetric nature of regularization losses. To alleviate this, we propose an asymmetric variant and apply it to the pooled output distillation (POD) loss adapted for ViTs. As we will see through the section, our experiments confirm that introducing asymmetry to POD boosts its plasticity while retaining stability across (a) and (b). Moreover, we acknowledge low forgetting measures for all the compared methods, indicating that ViTs might be naturally inclined continual learners.

\label{wrk:cl_vit}
Transformers have shown excellent results for a wide range of language tasks \citep{DBLP:conf/nips/BrownMRSKDNSSAA20, DBLP:journals/tacl/RoySVG21} over the course of the last couple of years. Influenced by their initial results, Dosovitskiy \textit{et al.} \citep{vit} proposed Vision Transformers (ViTs) as the first firm yet competitive application of transformers within the computer vision community.\footnote{By firmness, we refer to the non-reliance on convolutional operations.} ViTs' applications have since spanned a range of vision tasks, including, but not limited to image classification \citep{deit}, object recognition \citep{Liu2021SwinTH}, and image segmentation \citep{Wang2021EndtoEndVI}. The singlemost essential element of their architecture remains the self-attention mechanism (SAM) that allows the learning of long-range interdependence between the elements of a sequence (or patches of an image). Another  feature vital to their performance is the way they are pretrained in an often unsupervised or self-supervised manner over a large amount of data. This is then followed by the finetuning stage where they are adapted to a downstream task \citep{DBLP:conf/naacl/DevlinCLT19}. 
\begin{figure}[ht!]
	\begin{center}
		\includegraphics[width=0.65\textwidth]{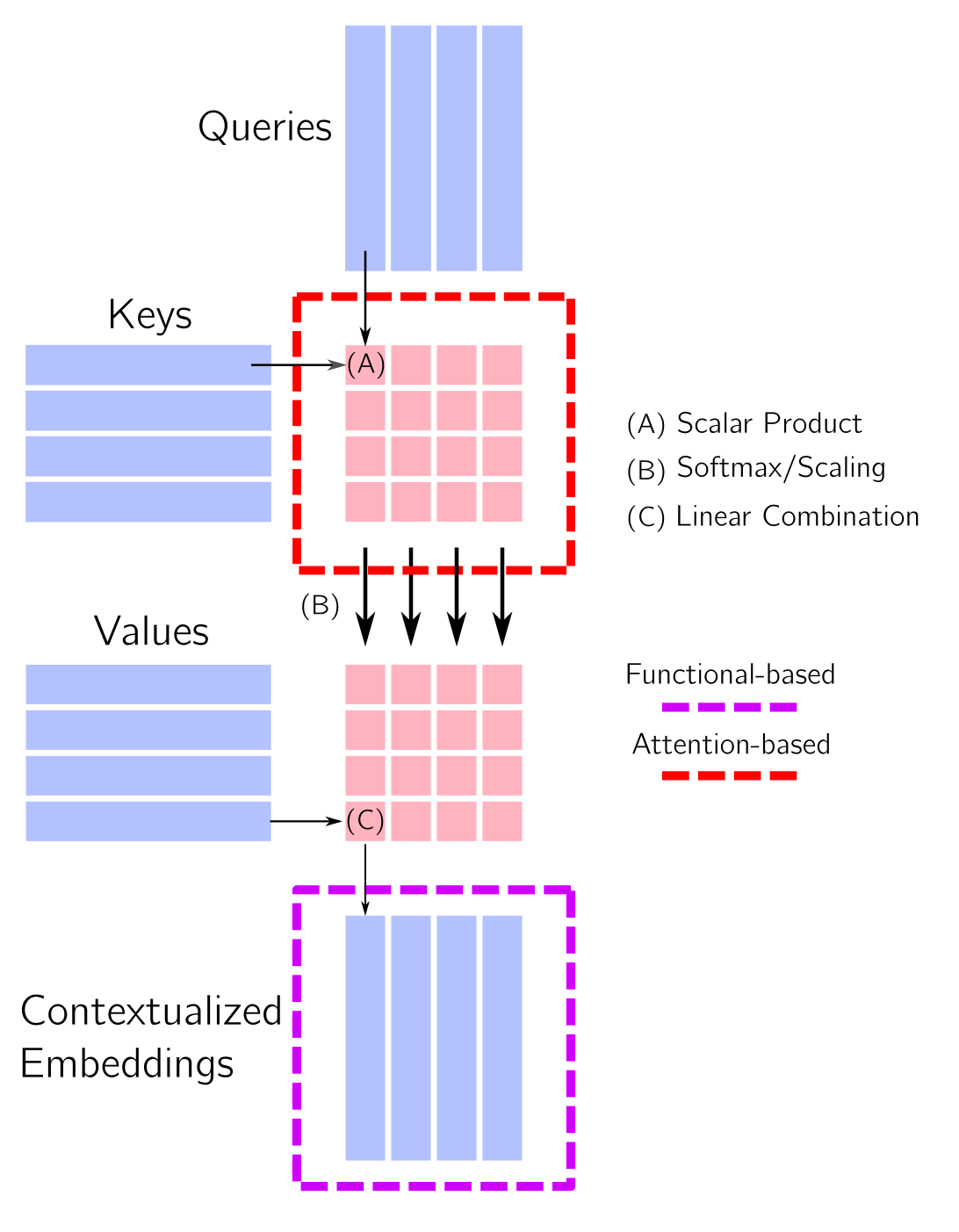}
		\caption{ Self-attention mechanism comprising a vision transformer encoder. We compare \textcolor{red}{Attention}-based approaches computed prior to the softmax operation and \textcolor{mypurple}{Functional}-based approaches computed on the contextualized embeddings.}
		\label{fig:attention_functional}
	\end{center}
\end{figure}

For ViTs to be able to operate in real-world scenarios, they must exploit streaming data, \textit{i.e.,} sequential availability of training data for each task.\footnote{A task may encompass training data of one or more classes.} Storage limitations or privacy constraints further imply the restrictions on the storage of data from previous tasks. Task-incremental continual learning (CL) seeks to find solutions to such constraints by alleviating the event of \textit{catastrophic forgetting} - a phenomena where the network has a dramatic drop in performance on data from previous tasks. Several solutions have been proposed to address forgetting, including regularization \citep{ewc, mas, si, laplace}, data replay \citep{tiny, ermir, gem} and parameter isolation \citep{packnet, pnn, aljundi2017expert, DBLP:conf/iclr/LeeHZK20}. Most works on CL \textit{de nos jours} study recurrent \citep{Sodhani2020TowardTR, DBLP:conf/nips/ChiaroTB020} and convolutional neural networks (CNNs) \citep{ewc}. However, little has been done to investigate different CL settings in the domain of ViTs. We, therefore, mark the first step for the domain by considering the further restrictive setting of \emph{exemplar-free} CL with a zero overhead of storing any data from previous tasks. We consider this restriction for its real-world aptness to scenarios involving privacy regulations and/or data security considerations.

Given that regularization-based methods form one of the main techniques for exemplar-free CL, we consider an in-depth analysis of these for ViTs. Regularization-based techniques are mainly organized along two  branches: \emph{weight regularization} methods (such as EWC~\citep{ewc}, SI~\citep{si}, MAS~\citep{mas}) and \emph{functional regularization} methods ( such as LwF~\citep{DBLP:conf/eccv/LiH16}, PODNET~\citep{podnet}). As discussed above, the architectural novelty of transformers lies in the SAM building a representation of a sequence by exhaustively learning relations among query-key pairs of its elements \citep{DBLP:conf/nips/VaswaniSPUJGKP17}. We show that for ViTs (and subsequently, all other architectures leveraging SAM), this property allows for a third form of regularization, which we coin \emph{Attention Regularization} (see Figure \ref{fig:attention_functional}). We ground our idea in the hypothesis that when learning new tasks, the attention of the new model should still remain in the neighborhood of the attention of the previous model. As another contribution, we question the temporal symmetry currently applied to regularization losses; referring to the fact that they penalize the forgetting of previous knowledge and the acquiring of new knowledge equally (see Figure \ref{fig:asym_illustrate}). With the aim of countering forgetting while mitigating the loss of plasticity, we then propose an \textit{asymmetric} regularization loss that penalizes the loss of previous knowledge but not the acquiring of new knowledge. We index the major contributions of our work below:
\begin{itemize}
	\item We are the first to investigate continual learning in vision transformers in the more challenging \textit{exemplar-free} setting. We perform a full analysis of regularization techniques to counter catastrophic forgetting. 
	\item Given the distinct role of self-attention in modeling short and long-range dependencies \citep{Yang2021FocalSF}, we propose distilling the attention-level matrices of ViTs. Our findings show that such distillation offers accuracy scores on par with that of their more common functional counterpart while offering superior plasticity and forgetting. Motivated by the work of Douillard \textit{et al.} \citep{podnet}, we pool spatiality-induced attention distillation across our network layers.
	\item  We propose an asymmetric variant of functional and attention regularization which prevents forgetting while maintaining higher plasticity. Through our extensive experiments, we show that the proposed asymmetric loss surpasses its symmetric variant across a range of task incremental settings.
\end{itemize}
	
	%%%%%%%%%%%%%%%%%%%%%%%%%%%%%%%%%%%%%%%%%%
	%%%%%%%%%%%% RELATED WORKS %%%%%%%%%%%%%%%%%%
	%%%%%%%%%%%%%%%%%%%%%%%%%%%%%%%%%%%%%%%%%%
	\section*{Related Works}
	Continual learning has been gaining contributions from the deep learning research community during the last few years. 
	In the following, we list the most prominent ones:
	
	\begin{itemize}
		\item Weight-based: these methods operate in the parameter space of the model through gradient updates. Elastic Weight Consolidation (EWC) \citep{ewc} and Synaptic Intelligence (SI) \citep{si} are two widely used methods in this family with the former being
		probably, the most well-known. EWC uses fisher information to identify the parameters important to individual tasks and penalizes their updates to preserve knowledge from older tasks. SI makes the neurons accumulate and exploit old task-specific knowledge to contrast forgetting.
		
		\item Functional-based: these methods rely upon trading the plasticity for stability by training either the current (new) model on older data or vice-versa. Learning Without Forgetting (LWF) \citep{DBLP:conf/eccv/LiH16} remains among the most widely known approaches in this family. It employs Knowledge Distillation \citep{kd} upon the logits of the network. 
		
		\item Parameter Isolation-based: also known as architectural approaches, these methods tackle CF through a dynamic expansion of the network's parameters as the number of tasks grow. Among the first widely known methods in this family remain Progressive Neural Network (PNN) \citep{pnn} followed by Dynamically Expandable Network (DEN) \citep{DBLP:conf/iclr/YoonYLH18} and Reinforced Continual Learing (RCL) \citep{rl1}. 
		%Architectural approaches have an advantage in task-aware continual learning, since we can isolate task specific parameters and query them through the task label at test time, but suffer a severe limitation in memory requirements due to the continuous growth of the model.   
	\end{itemize}
	
	% I have to fix this phrase
	The majority of the aforementioned works target CL in CNNs mainly due to their inductive bias allowing them to solve almost all problems that involve visual data. This can also be seen in several reviews \citep{Mai2022OnlineCL, biesialska-etal-2020-continual, 9349197, review0, DBLP:journals/nn/BelouadahPK21, Mai2022OnlineCL} reporting few approaches that consider architectures besides CNNs, despite the attempts to investigate CL in RNNs \citep{Sodhani2020TowardTR, DBLP:conf/nips/ChiaroTB020}.

	Only recently have some works analyzed catastrophic forgetting in transformers. Among the earliest to do so remains that of Li \textit{et al.} \citep{li2022technical} proposing the continual learning with transformers (COLT) framework for object detection in autonomous driving scenarios. Using the Swin Transformer \citep{Liu2021SwinTH} as the backbone for a CascadeRCNN detector, the authors show that the extracted features generalize better to unseen domains hence achieving lesser
	forgetting rates compared to ResNet50 and ResNet101 \citep{resnet18} backbones. In case of ViTs,  Yu \textit{et al.} \citep{Yu2021ImprovingVT} show that their vanilla counterparts are more prone to forgetting when trained from scratch. Alongside heavy augmentations, they employ a set of techniques to mitigate forgetting: (a) knowledge distillation, (b) balanced re-training of the head on exemplars (inspired by LUCIR \citep{Hou_2019_CVPR}), and (c) prepending a convolutional stem to improve  low-level feature extraction of ViTs.

	In their work studying the impact of model architectures in CL, Mirzadeh \textit{et al.} \citep{mirzadeh2022architecture} also experiment with ViTs in brief (with the rest of the work focusing mainly on CNNs). While they vary the number of attention heads of ViTs to show that this has little effect on the accuracy and forgetting scores, they further conclude that ViTs do offer more robustness to forgetting arising from distributional shifts when compared with their CNN-based counterparts with an equivalent number of parameters. The conclusion remains in line with previous works \citep{paul2021vision}.  Finally, \citep{DBLP:journals/corr/abs-2111-11326} attempt to overcome forgetting in ViTs through a parameter-isolation approach which dynamically expands the tokens processed by the last layer. For each task, they learn a new task-specific token per head. They then couple such approach through the usage of exemplars and knowledge distillation on backbone features. It is worth noting that these works rely either on pretrained feature extractors \citep{li2022technical} or rehearsal \citep{Yu2021ImprovingVT, DBLP:journals/corr/abs-2111-11326} to defy forgetting. Thus the challenging scenario of \textit{exemplar-free} CL in ViTs remains unmarked.

	%%%%%%%%%%%%%%%%%%%%%%%%%%%%%%%%%%%%%%%%%%
	%%%%%%%%%%%% METHOD %%%%%%%%%%%%%%%%%%
	%%%%%%%%%%%%%%%%%%%%%%%%%%%%%%%%%%%%%%%%%%
	\section*{Methodology}
	
	We start by shortly describing the two main existing regularization techniques for continual learning. We then propose attention regularization as an alternative approach tailored for ViTs. Lastly, we put forward an adaptation for functional and attention regularization which is designed to elevate plasticity while retaining stability levels.

	\subsection*{Functional and Weight Regularization}%%%%%%%%%%%%%%%%%%%%%%%%%%%%%
	
	\paragraph{Functional Regularization:} We include LwF \citep{DBLP:conf/eccv/LiH16} in this component since it constitutes one of the most prominent, and perhaps the most widely used regularization method acting on data. The appealing property of LwF lies in the fact it is exemplar-free, \textit{i.e.}, it uses only the data of the current task and maintains only the model at task $t-1$ to exploit Knowledge Distillation \citep{kd}. Formally, LwF can be defined as:
	\begin{equation}
	\mathcal{L}_{\text{LwF}}(\theta)=\lambda_{o} \mathcal{L}_{\mathrm{KD}}\left(Y_{o}, \hat{Y}_{o}\right)+\mathcal{L}_{\mathrm{CE}}\left(Y_{n}, \hat{Y}_{n}\right)+\mathcal{R}(\theta)
	\end{equation}
	where $\mathcal{L}_{\mathrm{KD}}$ is the knowledge distillation loss incorporated to impose stability on the outputs, $\hat{Y}_{o}$ the predictions on the current task data from the old model and $\hat{Y}_{o}$ the ground truth of such data. $\lambda_{o}$ remains the temperature annealing factor for softmax logits while $\mathcal{L}_{\mathrm{CE}}$ is the standard cross entropy loss calculated upon the new task examples.
	
	%\JW{We should also put the equation for feature distilation. SO we can later extend it to assymetric}
	
	\paragraph{Weight Regularization:} These methods encourage the network to adapt to the current task data mainly by using those parameters of the network that are not considered important for previous tasks. As representative method we select EWC \citep{ewc}. EWC exploits second-order information to estimate the importance of parameters for the current task. The importance is approximated by the diagonal of the Fisher Information Matrix $F$:
	\begin{equation}
	\mathcal{L}_{\text{EWC}}(\theta)=\mathcal{L}_{X}(\theta)+\sum_{j} \frac{\lambda}{2} F_{j}\left(\theta_{j}-\theta_{Y, j}^{*}\right)^{2}
	\end{equation}
	where $\mathcal{L}_{X}(\theta)$ is the loss for task X, $\lambda$ the regularization strength, and $\theta_{Y, j}^{*}$ the optimal value of $j^{t h}$ parameter after having learned task Y. 
	
	\subsection*{Attention Regularization}%%%%%%%%%%%%%%%%%%%%%%%%%%%%%
	\label{sec:att_reg}
	\paragraph{Self-Attention Mechanism:} 
	
	The self-attention mechanism (SAM) \citep{DBLP:conf/nips/VaswaniSPUJGKP17} forms the core of Transformer-based models and can be defined as:
	
	\begin{figure}[t!]
		\begin{center}
			\includegraphics[width=\textwidth]{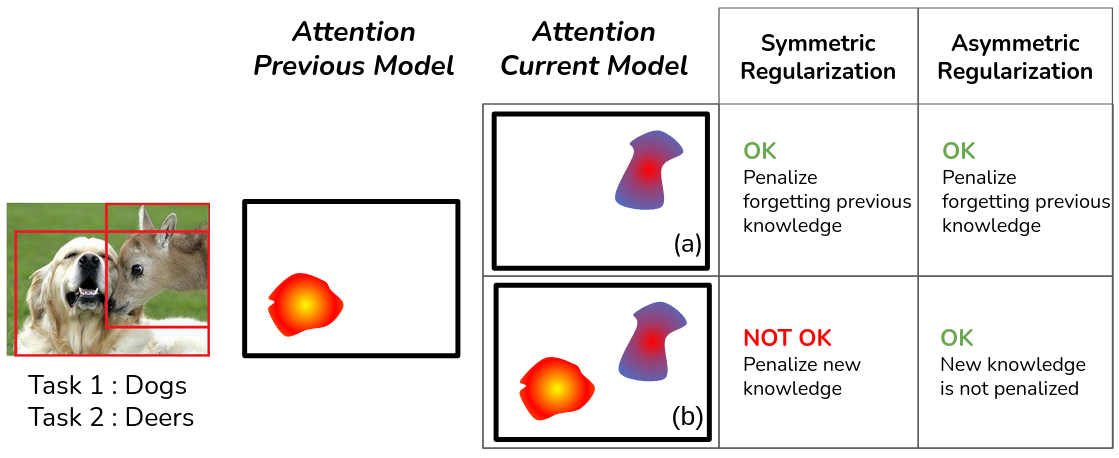}
			\caption{Visual illustration of the asymmetric loss. The image considers two generated attention maps (a) and (b) while training task 2. In case (a), when previous knowledge is lost, both the symmetric and assymetric regularization work correctly. However, in case (b), when new knowledge is acquired, this is penalized by the symmetric loss but not by the assymetric loss. The idea is that the assymetric loss leads to higher plasticity without hurting stability.}
			\label{fig:asym_illustrate}
		\end{center}
	\end{figure}
	\begin{equation}
	\mathbf{z} = \mathbf{softmax}\left ( \frac{\mathbf{QK}^T}{\sqrt{d_e}} \right )\mathbf{V}
	\end{equation}
	where $\mathbf{Q,K}$, and $\mathbf{V}$ are respectively the projections of the Query, Key, and Values of the $\mathbb{R}^{d_e}$ input embeddings while $\mathbf{z}$ constitutes the new contextualized embeddings. Our novel attention-based regularization intervenes prior to the computation of the softmax operation of the standard self-attention mechanism as illustrated in Figure~\ref{fig:attention_functional}. 
	
	In particular, given a ViT model at incremental step $t$ and  an SAM head $k$ of layer $l$, we define the prescaled attention matrix $\mathbf{A}^{t}_{k^{l}}$ prior to the softmax operation as:
	\begin{equation}
	\mathbf{A}^{t}_{k^{l}} = \frac{\mathbf{QK}^T}{\sqrt{d_e}}
	\end{equation} 
	We denote the attention matrix corresponding to the model at time step $(t-1)$ computed in a similar way as $\mathbf{A}^{t-1}_{k^{l}}$. We employ this predecessor in the calculation of knowledge distillation in what follows.

	%%%%%%%%%%%%%%%%%%%%%%%%%%%%%%%%%%%%%%%%%%
	%%%%%%%%%%%%%%%%%%%%%%%%%%%%%%%%%%%%%%%%%%
	%I would rewrite the Pooled Attention Distillation.
	
	\paragraph{Pooled Attention Distillation:}
	Functional approaches leverage network's submodules typically to apply knowledge distillation \citep{kd}. When the regularization takes place in intermediate layers, the model can experience excessive stability, therefore loosing in plasticity abilities~\citep{podnet, liu2020generative, yu2020semantic}. Amongst these methods, PODNet \citep{podnet} clearly identifies the problem of excessive stability. We devise a regularization approach which instead of regularizing functional submodules targets attention maps, the core mechanisms of SAMs.
	
	More formally, given the attention maps at steps $t$ and $(t-1)$, we define $\mathcal{L}_{\text {PAD}}\left(\mathbf{A}^{t-1}_{k^l}, \mathbf{A}^{t}_{k^l}\right)$ \citep{podnet} to be:
	
	\begin{equation}
	\label{eqn:spatial}
	\mathcal{L}_{\text {PAD-width}}\left(\mathbf{A}^{t-1}_{k^l}, \mathbf{A}^{t}_{k^l}\right) + \mathcal{L}_{\text {PAD-height}}\left(\mathbf{A}^{t-1}_{k^l}, \mathbf{A}^{t}_{k^l}\right)
	\end{equation}
	\vskip -0.2in
	\begin{align}
	\label{eqn:pod_width_height_sym}
	\begin{split}
	\text{where } \mathcal{L}_{\text{PAD-width}}\left(\mathbf{A}^{t-1}_{k^l}, \mathbf{A}^{t}_{k^l}\right) = \sum_{h=1}^{H} \mathcal{D}_W\left(\mathbf{A}^{t-1}_{k^l}, \mathbf{A}^{t}_{k^l}\right),
	\\
	\mathcal{L}_{\text{PAD-height}}\left(\mathbf{A}^{t-1}_{k^l}, \mathbf{A}^{t}_{k^l}\right) = \sum_{w=1}^{W} \mathcal{D}_H\left(\mathbf{A}^{t-1}_{k^l}, \mathbf{A}^{t}_{k^l}\right),\end{split}
	\end{align}
	\vskip -0.2in
	\begin{equation}
	\label{eqn:symdist}
	\mathcal{D}_X\left(\mathbf{A}^{t-1}_{k^l}, \mathbf{A}^{t}_{k^l}\right) =  \left\|\sum_{x=1}^{X} \mathbf{A}^{t-1}_{k^l, w, h}-\sum_{x=1}^{X} \mathbf{A}^{t}_{k^l, w, h} \right\|^{2}
	\end{equation}
	where, $W$ and $H$ indicate the width and height dimensions of the attention maps, and $\mathcal{D}_X(a,b)$ is the sum total of the distance measure between maps \textit{a} and \textit{b} along \textit{X}-th dimension. As shown in equation \ref{eqn:symdist}, the standard $\mathcal{L}_{\text{PAD}}$ uses the difference operator as the choice for $\mathcal{D}$. We now point out the limitation of such  symmetric $\mathcal{D}$ and introduce in the next section the notion of \textit{asymmetry} into our distance measure.
	
	As previously mentioned, Douilllard \textit{el al.} \citep{podnet} propose the pooled outputs distillation PODNet loss which leverages the symmetric Euclidean distance between the L2-normalized outputs of the convolutional layers of models at $t$ and $(t-1)$ after pooling them along specific dimension(s). They achieve the best results upon combining the pooling along the spatial width and  height axes which they term as the POD-spatial loss. Given the generic correspondence among the various pooling variants in their paper, our work is particularly influenced by POD-spatial as we pool attention maps of ViTs along two dimensions. In fact, throughout the experiments, we analyze this formulation when applied to the contextualized embeddings $\mathbf{z}$ resulting from a SAM operation. 
	We would like to highlight that PAD differs from PODNet in two important factors: its applied to the attention and not directly on the layer output, and secondly, its marginalization is not on the spatial dimensions due to the fact that $z$ does not encode the spatial dimension.
	
	%We highlight the fact that, even if the application of the loss to a functional submodule of a ViT would seem PAD to be equal to PODNet, this is not true. Indeed, the extension is not straightforward due to the fact that $z$ does not encode the spatial dimension directly in width and height unlike its CNNs feature maps counterpart. In fact, in contextualized embeddings, these dimensions acquire a different meaning which is not directly interpretable. 
	
	\begin{figure}[t!]
		\begin{center}
			\includegraphics[width=1\textwidth]{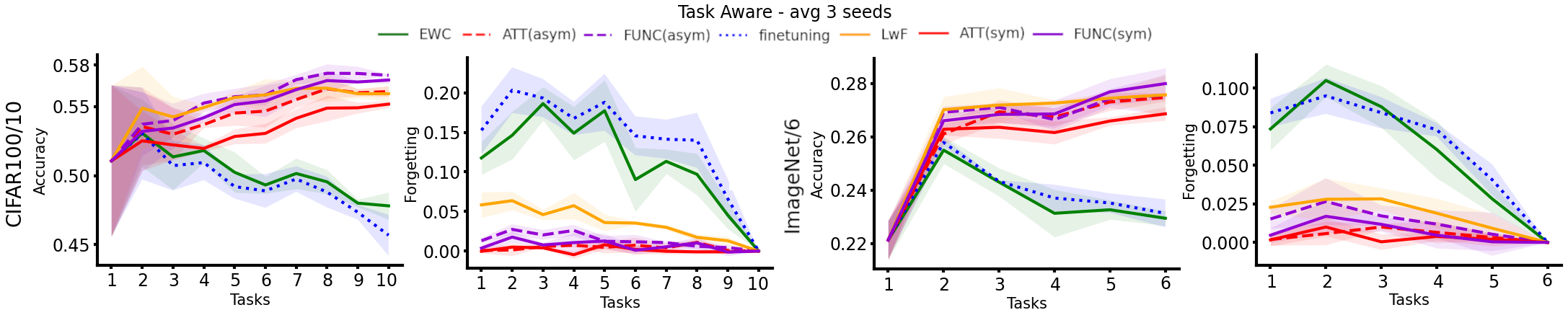}
			\caption{Mean and standard deviation of task-aware accuracy and forgetting scores for CIFAR100/10 and ImageNet/6 settings (over 3 random runs). Asymmetric approaches depict higher accuracy with respect to their symmetric counterparts. The low forgetting scores across all methods suggest an intrinsic forgetting resilience in vision transformer architectures. }
			\label{fig:acc-forgetting}
		\end{center}
	\end{figure}
	
	\subsection*{Asymmetric Regularization}%
	\label{sec:asym_reg}
	The proposed attention regularization prevents forgetting of previous task by ensuring  
	that the old attention maps be retained while the model learns to attend to new regions over tasks. However, the symmetric nature of $\mathcal{D}_X$  (with respect to the two attention maps) means that any differences between the older and the newly learned attention maps lead to increased loss values (see Equation \ref{eqn:pod_width_height_sym}). 
	We agree that penalizing a loss in attention with respect to  previous knowledge is crucial in addressing forgetting. However, also penalizing a gain in attention for newly learned knowledge is undesirable and may actually hurt the performance over subsequently learned tasks. In other words, punishing additional attention can be counterproductive. As a result, we propose using an asymmetric variant of $\mathcal{D}_X$ that can better retain previous knowledge: 
	\begin{equation}
	\label{eqn:asym_dist}
	\mathcal{D}_X\left(\mathbf{A}^{t-1}_{k^l}, \mathbf{A}^{t}_{k^l}\right) =  \left\| \mathcal{F}_{\text {asym}}\left(\sum_{x=1}^{X} \mathbf{A}^{t-1}_{k^l, w, h}-\sum_{x=1}^{X} \mathbf{A}^{t}_{k^l, w, h} \right) \right\|^{2}
	% \sum_{h=1}^{H}\left\|\text{ReLU} \left (\sum_{w=1}^{W} \mathbf{A}^{t-1}_{k^l, w, h}-\sum_{w=1}^{W} \mathbf{A}^{t}_{k^l, w, h}\right ) \right\|^{2}
	\end{equation}
	where, $\mathcal{F}_{\text{asym}}$ is as asymmetric function. We experimented with ReLU \citep{Nair2010RectifiedLU}, ELU \citep{DBLP:journals/corr/ClevertUH15} and Leaky ReLU \citep{Maas13rectifiernonlinearities} as choices for $\mathcal{F}_{\text{asym}}$ and found that in general, ReLU performed the best across our settings. By introducing the ReLU function, new attention generated by the current model at task $t$ is not penalized. Attention present at task $t-1$ but missing in the current model $t$ is penalized. An illustration of the functioning of the new loss is provided in Figure~\ref{fig:asym_illustrate}.
	
	Based on our choice for $\mathcal{D}_X$ from equations \ref{eqn:symdist} and \ref{eqn:asym_dist}, we classify our final PAD loss as symmetric $\mathcal{L}_{\text{PAD-sym}}$ or asymmetric $\mathcal{L}_{\text{PAD-asym}}$, respectively. Each of these losses are computed separately for each of the SAM head and model layer. The final asymmetric variant can thus be stated as:
	\begin{equation}
	\label{eqn:asym_final}
	\begin{split}
	\mathcal{L}_{\text {PAD-asym}}(\mathbf{A}^{t-1}_{k^l}, \mathbf{A}^{t}_{k^l}) =\\  \frac{1}{L}\sum_{1}^{L} \frac{1}{K} \sum_{1}^{K}\mathcal{L}_{\text {PAD }}(\mathbf{A}^{t-1}_{k^l}, \mathbf{A}^{t}_{k^l})
	\end{split}
	\end{equation}
	where, $K$ is the total number of heads per layer and $L$ is the total number of layers of the model. Note that equation \ref{eqn:asym_final} can be adapted for $\mathcal{L}_{\text{PAD-sym}}$ without loss of generality.
	
	\paragraph{Overall loss:} We augment the asymmetric and symmetric PAD losses from equation \ref{eqn:asym_final} with knowledge distillation loss $\mathcal{L}_{LwF}$ \citep{DBLP:conf/eccv/LiH16} and standard cross entropy loss $\mathcal{L}_{CE}$. The overall loss term takes the form:
	
	\begin{figure}[t!]
		\begin{center}
			\includegraphics[width=1\textwidth]{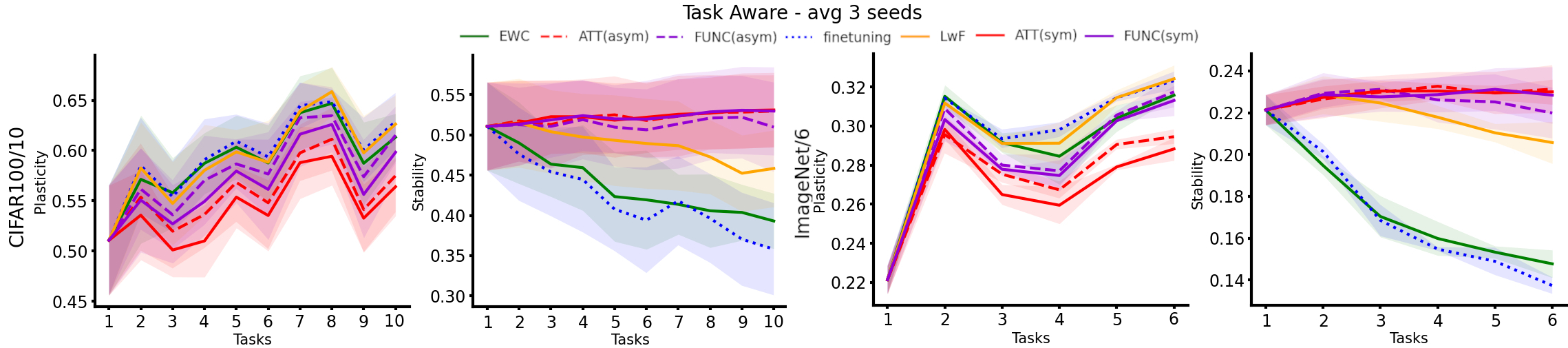}
			\caption{Mean and standard deviation of task-aware plasticity-stability scores for CIFAR100/10 and ImageNet/6 settings (over 3 random runs). Asymmetric approaches are more plastic compared to their symmetric counterparts while retaining competitive stability.}
			\label{fig:cifar_stab_plast}
		\end{center}
	\end{figure}
	
	\begin{equation}
	\label{eqn:total_loss}
	\mathcal{L} = \mu \mathcal{L}_{\text {PAD-(a)sym}} + \lambda \mathcal{L}_\mathrm{LwF} + \mathcal{L}_{CE} 
	\end{equation}
	where $\mu$, $\lambda \in [0,1]$ are two hyperparameters regulating the respective contributions. Note that when $\mu = 0$, $\mathcal{L}$ degenerates to baseline finetuning for $\lambda = 0$ and to LwF for $\lambda = 1$.
	
	\paragraph{Stability-Plasticity Curves:}%%%%%%%%%%%%%%%%%%%%%%%%%%%%%
	\label{sec:stab_plast_curves}
	Several measures have been proposed in the CL literature to assess the performance of an incremental learner. Besides the standard incremental accuracy, Lopez-Paz et al \citep{gem} introduce the notion of Backward Transfert (BWT) and Forward Transfert (FWT). BWT measures the ability of a system to propagate knowledge to past tasks, while FWT assesses the ability to generalize to future tasks. The CL community, however, still lacks consensus on a specific definition of the stability-plasticity dilemma. An elemental formulation for such quantification is thus desirable for allowing us to better grasp the balancing capabilities of an incremental learner at acquiring new knowledge without discarding previous concepts. To this end, we introduce \emph{stability-plasticity curves} computed using task accuracy matrices. 
	
\begin{figure}[t!]
		\begin{center}
			\includegraphics[width=0.4\columnwidth]{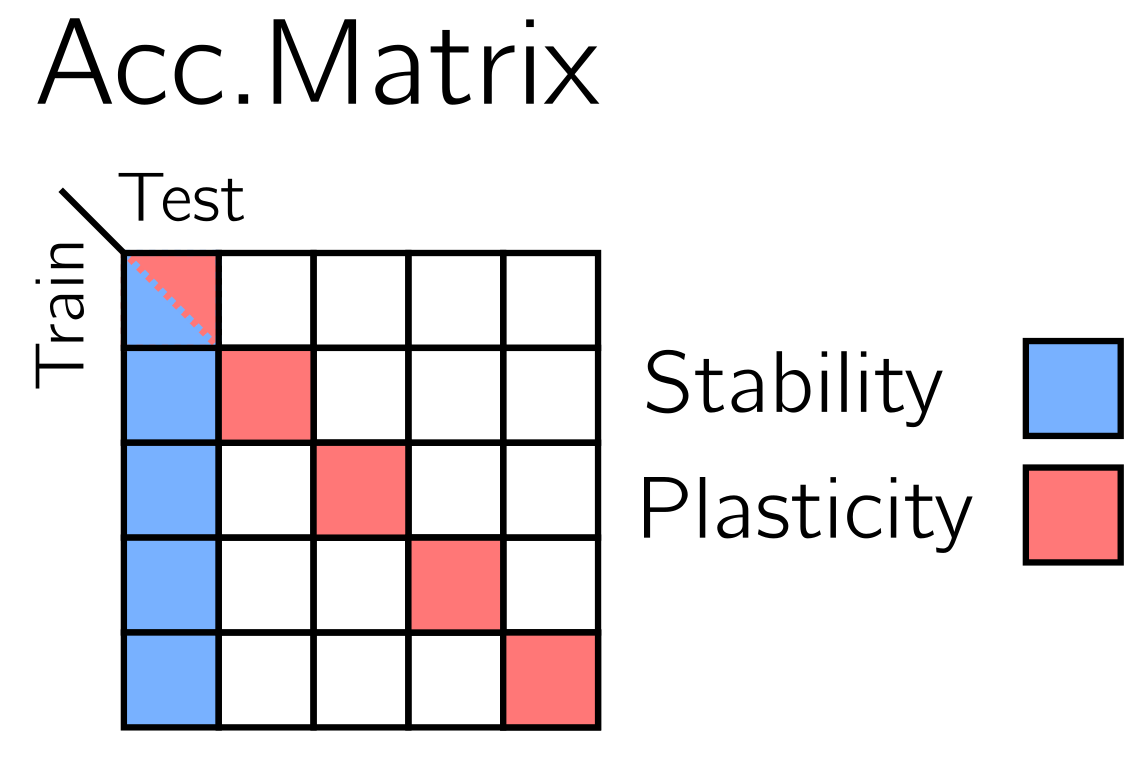}
			\caption{Illustration of a task accuracy matrix: we fix \textcolor{blue}{stability} to be the performance of the first task across time steps while we define \textcolor{red}{plasticity} to be the performance at the current step.}
			\label{fig:acc_mat}
		\end{center}
	\end{figure}

	A task accuracy matrix $\mathbf{M}$ for an incremental learning setting composed of $T$ tasks is defined to be a $[0,1]^{T \times T}$ matrix, whose entries are the accuracies computed at each incremental step.\footnote{This calls for $\mathbf{M}$ to be lower trapezoidal.} For instance, $\mathbf{M}_{i,j}$ would constitute the test accuracy of task $j$ when the system is learning task $i$. Subsequently, the diagonal entries of  $\mathbf{M}_i,i$ give us the accuracies at the respective current tasks while the entries below the diagonal, \textit{i.e.,} $j<i$, give the performance of the model on past tasks. A visual depiction can be seen in Figure~\ref{fig:acc_mat}. 

	We define the stability to be the performance on the first experienced task at any given time and plasticity to be the ability of the model to adapt to the current task. Namely, these constitute the first column $\mathbf{M}_{:,0}$ and the diagonal of the matrix $diag(\mathbf{M})$. We employ the curves dervied from these definitions to better dissect the stability-plasticity dilemma of the methods analyzed in our work.

	%%%%%%%%%%%%%%%%%%%%%%%%%%%%%%%%%%%%%%%%%%
	%%%%%%%%%%%% EXPERIMENTS %%%%%%%%%%%%%%%%%%
	%%%%%%%%%%%%%%%%%%%%%%%%%%%%%%%%%%%%%%%%%%

	\section*{Experiments}
	
	In this section, we compare regularization-based methods for \textit{exemplar-free} continual learning. We evaluate the newly proposed attention regularization and compare it with the existing functional (\emph{LwF}) and feature regularization methods. We then ablate the usefulness of the newly proposed asymmetric loss as well as the importance of pooling before applying the regularization.

	\subsection*{Experimental Setup}
	
	\paragraph{Setting:}
	
	For our experiments, we adopt the variation of ViTs introduced by Xiao \textit{et al.} \citep{early_conv}.  Here, the standard linear embedder of a ViT model is replaced by a smaller convolutional stem which helps build more resilient low-level features. Convolutional stems have previously been shown to improve performance and convergence speed in incremental learning settings \citep{Yu2021ImprovingVT}. We therefore define our architecture to be a lightweight variation of a ViT-Base by setting $L=12$ layers, $K=12$ heads per layer and a $d_e = 192$ embedding size. The choice of a small embedding size has been made to speed up the training procedure and unlock the ability to handle larger batch sizes (1024 for our work). 
	
	We analyze our task-incremental setting on two widely used image recognition datasets - namely CIFAR100 and ImageNet-32 with 100, and 300 classes each. Both datasets host $32\times32$ images. On CIFAR100, we consider a split of 10 tasks (denoted as CIFAR100/10 setting) where each incremental task is composed of 10 disjoint set of classes. On ImageNet-32, we split 6 tasks with 50 disjoint set of classes each (denoted as ImageNet/6).\footnote{Refer to Section \ref{tag_appendix} for experiments on additional settings.} 
	
	Our total training epochs remain 200 (per task) for CIFAR100 and 100 for ImageNet32 with an initial learning rate of 0.01 and patience set of 20 epochs. We report our scores averaged over 3 random runs.
	We apply a constant padding of size 4 across all our datasets. The train images are augmented using random crops of sizes $32\times32$ and random horizontal flips with a flipping probability of 50\%. For test images, we only apply center crops of sizes $32\times32$. 
	
	We compare the attentional and functional symmetric and asymmetric versions of $\mathcal{L}_{\text {PAD-(a)sym}}$. We use LwF \citep{DBLP:conf/eccv/LiH16} and EWC \citep{ewc} as our basic functional and weight regularization approaches. For all our experiments relying on PAD losses, we performed a hyperparemeter search (using equation \ref{eqn:total_loss}) for $\mu$ and $\lambda$ by varying each in the range $[0.5, 1.0]$ and found $\mu = \lambda = 1.0$ to perform reasonably well. We thus stick to these values unless otherwise specified. For the sake of brevity, we indicate $\mathcal{L}_{\text{PAD-asym}}$ with Asym\_att and  $\mathcal{L}_{\text{PAD-sym}}$ with Sym\_att. Note that these are both variations of equation \ref{eqn:total_loss}. The functional approaches are analogous to their attentional counterparts except for the fact that they rely on the regularization of the contextualized embeddings rather than the attention matrix (see Figure~\ref{fig:attention_functional}). The latter correspond to Asym\_func and Sym\_func accordingly.

	\subsection*{Results}%%%%%%%%%%%%%%%%%%%%%%%%%%%%%
	\label{sec:main_results}
	We report accuracy as well as forgetting \citep{chaudhry2018riemannian} scores in task aware (taw) setting\footnote{The corresponding task agnostic scores can be found in Figure \ref{fig:cifar100_tag_settings}, Section \ref{tag_appendix}.}. We further report taw plasticity-stability curves (based on Figure \ref{fig:acc_mat}) to provide insights upon how well the different models handle the trade-off.

	\paragraph{Accuracy and Forgetting:}%%%%%%%%%%%%%%%%%%%%%%%%%%%%%
	As seen in  Figure~\ref{fig:acc-forgetting}, all asymmetric approaches show better performances  with respect to their symmetric counterparts on CIFAR100/10 with Asym\_att offering the best accuracy of 57.3\% on the last task. The trend continues for ImageNet/6 with an exception of asymmetric functional approach with an accuracy of 27.55\% falling behind its symmetric counterpart by 0.44\%. 
	In general, the asymmetric and symmetric losses lead to improved accuracy scores with respect to other methods. Moreover, we observe that all the methods depict good forgetting resilience with their forgetting scores running around $\approx$0.01\%) except for EWC. This suggests us that vision transformers are \textit{better incremental learners} but require more training  and tuning efforts to achieve reasonable accuracies. This remark remains in accordance with prior studies \citep{mirzadeh2022architecture, paul2021vision}. In the particular case of EWC, we observe poor compatibility in terms of accuracy as well as forgetting -- with the scores falling behind finetuning at times. We suspect that the method might not be less suited for ViTs due to its reliance on exhaustive fisher information estimation.

	\paragraph{Plasticity-stability tradeoff:}%%%%%%%%%%%%%%%%%%%%%%%%%%%%%
	We compare the dilemma for various methods in Figure~\ref{fig:cifar_stab_plast}. With no distillation, finetuning is prone to the worst trading of plasticity for stability. Meanwhile, our asymmetric losses can be seen to be more plastic with respect to their symmetric counterparts while depicting comparable stability scores. This confirms our hypothesis regarding the nature of the asymmetry keeping it from discarding older attention while favoring the integration of new attention at the same time. Although, LwF with a last task score of 47.74\% on CIFAR100/10 and 32.0\% on ImageNet/6, reports the best plasticity among our approaches, it clearly lags behind the pooling-based approaches at retaining stability. On the contrary, the (a)symmetric attention losses and the symmetric functional loss perform similar with a last task stability score of $\approx0.23\%$ on ImageNet/6 and $\approx53\%$ on CIFAR100/10. EWC shows good plasticity but virtually zero stability. This trend is in line with our previous comment on the limitation of EWC in Figure~\ref{fig:acc-forgetting}.

	%On the other hand, Sym\_attentional and PODNet with respective plasticity scores of 27.47 and 21.37 perform the worst on CIFAR100/10.  PODNet and sym\_attentional with average scores of 34.67 and 23.2 offer the best stability over CIFAR100/10 while EWC and finetuning with respective averaged scores of 0.08 and 0.74 remain the worst.
	
	%On ImageNet/6, we observe a similar trend in plasticity where finetuning and LwF again remain the most plastic with average scores of 22.64 and 22.28. Also, PODNet and sym\_attentional with average scores of 16.81 and 13.69 over 10 tasks perform the worst. In terms of stability, PODNet and sym\_attentional remain the best with their respective scores averaging to 14.98 and 11.42 while EWC and finetuning with their average scores of 0.062 and 0.057 perform the worst. On both the datasets, we observe that the attentional methods remain more stable than their functional counterparts. 
	
	\begin{table}[]
		\centering
			\small
			\begin{tabular}{@{}cccccc@{}}
				\multicolumn{6}{c}{\textbf{CIFAR100/10  (taw)}}                                                                                                                                                                                                                                                                                                                                              \\ \midrule
				\textbf{}                                                                    & \textit{\begin{tabular}[c]{@{}c@{}}Asym\_Func \\ Spatial\end{tabular}} & \textit{\begin{tabular}[c]{@{}c@{}}Sym\_Func \\ Spatial\end{tabular}} & \textit{\begin{tabular}[c]{@{}c@{}}Asym\_Func \\ Intact\end{tabular}} & \textit{\begin{tabular}[c]{@{}c@{}}Sym\_Func \\ Intact\end{tabular}} & \textit{LwF} \\
				\textit{\begin{tabular}[c]{@{}c@{}}Average Incr.\\  Accuracy\end{tabular}} & \textbf{56.18\%}                                                               & 55.67\%                                                              & 54.43\%                                                             & 53.12\%                                                            & 55.11\%      \\
				\textit{\begin{tabular}[c]{@{}c@{}}Last Task \\ Accuracy\end{tabular}}           & \textbf{57.26\%}                                                               & 56.92\%                                                              & 56.04\%                                                             & 54.59\%                                                            & 55.93\%      \\ \bottomrule
			\end{tabular}

		\caption{Comparison of intact (no pooling), spatial (pooling along width and height), and LwF.}
		\label{tbl:loss_ablation}
	\end{table}
	
	\subsection*{Ablation study}
	
	Towards the end goal of evaluating the effectiveness of PAD losses, we ablate the contribution of pooling on the CIFAR100/10 setting.
	In particular, we consider distilling the attention maps when these are: (a) pooled along both dimensions, \textit{i.e.,}(A)sym\_Func Spatial (see Equation \ref{eqn:spatial}), and (b) not pooled at all, \textit{i.e.,} (A)sym\_Func Intact. Distilling the intact maps of the latter setting imply enhanced stability over their pooled counterparts. Our standard accuracy and plasticity-stability measures across tasks can therefore be deemed redundant in this setting. As a consequence, we choose to compare the task-aware average incremental accuracy \citep{icarl} and the last task accuracy across (a) and (b) while contrasting these with LwF as a strong baseline. For further crisper observations, we limit our comparisons to the functional setting. As shown in Table \ref{tbl:loss_ablation}, we find that Asym\_Func Spatial consistently performs the best across both the metrics (with a gain of  $>2\%$ over Sym\_Func Intact in either metric). In general, distilling the intact attention maps can be seen to be hurting the performance of the models as their accuracy drop below that of the baseline LwF.
	
	% \begin{figure}[t]
	% \centering
	% \centerline{\includegraphics[width=1\columnwidth]{figs/ablation_less_data.png}}
	% \caption{Ablation Less Data}
	% \label{fig:ablation_less_data}
	% \end{figure}
	
	% \paragraph{Robustness of pooling: } As a measure of the robustness of our proposed PAD losses, we adopt a data-scarce CL setting that can better reflect real-world scenarios with swift distribution shifts. Namely, we preserve the number of examples in the base task classes while randomly dropping $90\%$ of data from the remaining incremental classes.  Given that the incremental models now have a mere $10\%$ of the original data to learn from, they are obvious to acquire moderate plasticity. A rather desirable factor here is that the models preserve their learned knowledge over tasks. We therefore opt for comparing the stability of the methods as a more informative metric in such a setting. We also show their forgetting scores across tasks as a complement to the aforesaid argument. The comparison of finetuning and EWC are excluded due to their inadequate performances. A comparison of different CL methods in this setting is depicted in Figure \ref{fig:ablation_less_data}. We observe LwF
	
		\begin{figure}[t!]
		\centering
		\centerline{\includegraphics[width=0.8\textwidth]{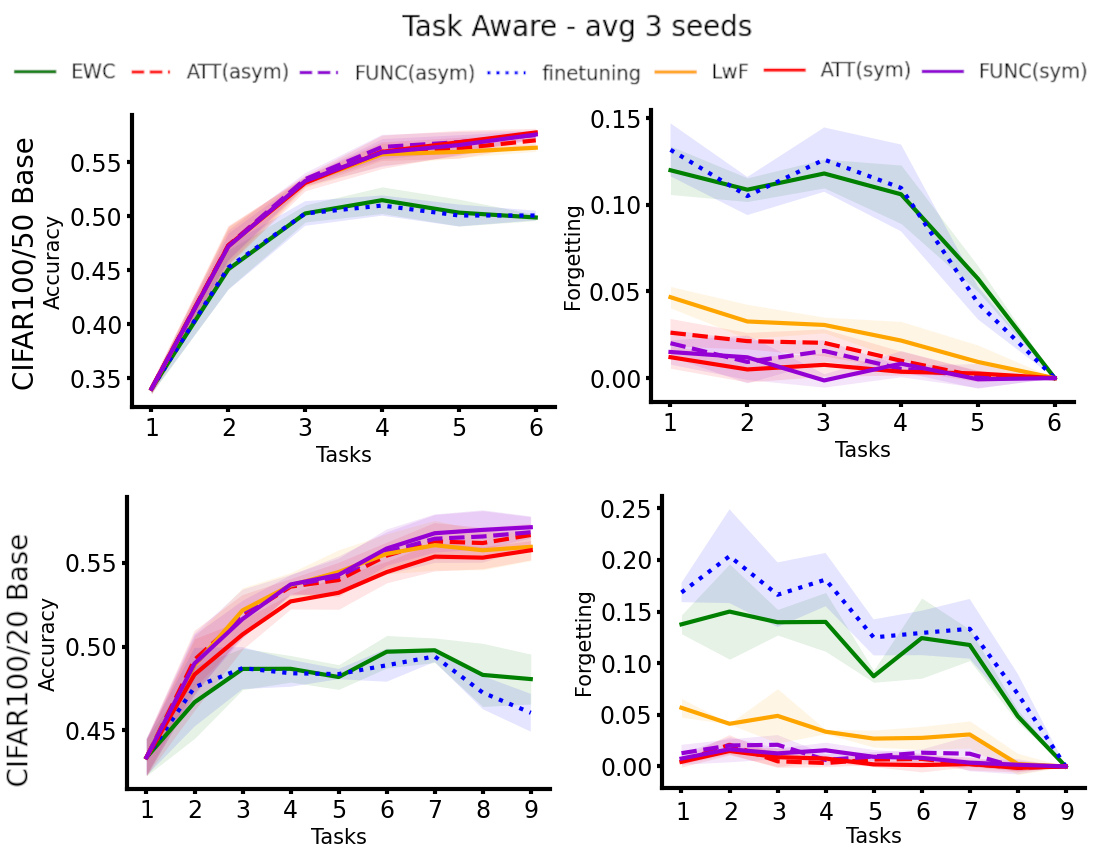}}
		\caption{Mean and standard deviation of task-aware accuracy and forgetting scores for the additional CIFAR100/20 and CIFAR100/50 settings (over 3 random runs).}
		\label{fig:cifar100_ttaw_50_20}
	\end{figure}

	%%%%%%%%%%%%%%%%%%%%%%%%%%%%%%%%%%%%%%%%%%
	%%%%%%%%%%%% CONCLUSION %%%%%%%%%%%%%%%%%%
	%%%%%%%%%%%%%%%%%%%%%%%%%%%%%%%%%%%%%%%%%%
	\section*{Conclusion}
	
	In this work, we adapted and analyzed several continual learning methods to counter forgetting in Vision Transformers mainly with the help of regularization. We then introduced a novel PODNet-inspired regularization, based on the attention maps of self-attention mechanisms which we termed as Pooled Attention Distillation (PAD). Shedding light on its limitation at learning new attention, we devised its asymmetric version that avoids penalizing the addition of new knowledge in the model. We validated the superior plasticity of the asymmetric loss on several benchmarks. 
	
	Besides the meticulous comparison of a range of regularization approaches, \textit{i.e.}, functional (LwF), weight (EWC), and the proposed attention-based regularization, we extended the application of PAD to the functional submodules of ViTs. To this end, we investigated regularization in the contextualized embeddings of ViTs.  The latter exploration led us to discover that the regularization of functional submodules can help achieve the best overall performances while the regularization of their attentional counterparts endow CL models with superior stability. Finally, we remarked the low forgetting scores of vision transformers across the incremental tasks and concluded that their enhanced generalization capabilities may endow them with a natural inclination for incremental learning.  By making our code open-source, we hope to open the doors for future research along the direction of efficient continual learning with transformer-based architectures.

	%%%%%%%%%%%%%%%%%%%%%%%%%%%%%%%%%%%%%%%%%%
	%%%%%%%%%%%% APPENDIX %%%%%%%%%%%%%%%%%%
	%%%%%%%%%%%%%%%%%%%%%%%%%%%%%%%%%%%%%%%%%%
		
	\section*{Additional Settings}
	\label{tag_appendix}
	
	We experiment on two further CIFAR100 settings with distinct cardinality of base task classes:
	\begin{itemize}
		\item CIFAR100/20 Base, with 20 base task classes followed by 8 incremental tasks with 10 classes each,
		\item CIFAR100/50 Base, with 50 base task classes followed by 5 incremental tasks with 10 classes each.
	\end{itemize}

	\begin{figure}[h!]
		\begin{center}
			\centerline{\includegraphics[width=0.8\textwidth]{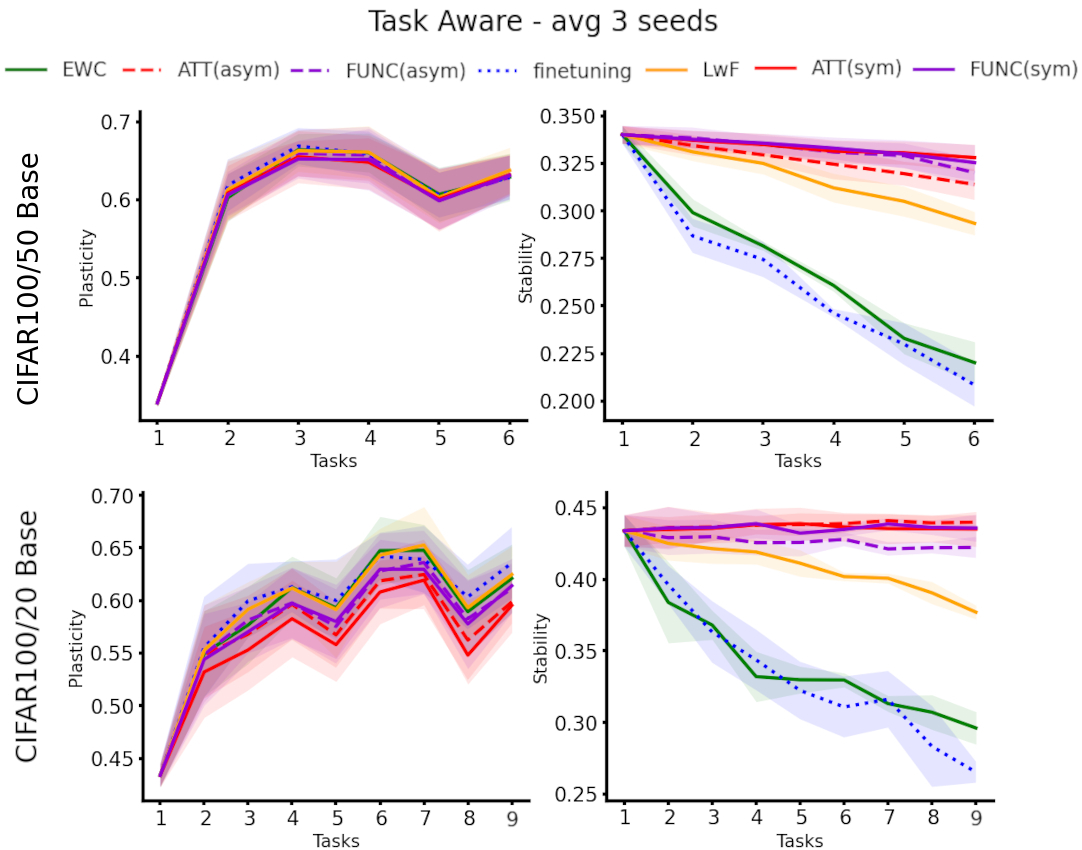}}
			\caption{Mean and standard deviation of task-aware plasticity-stability scores for the additional CIFAR100/20 and CIFAR100/50 settings (over 3 random runs).}
			\label{fig:cifar100_other_taw}
		\end{center}
	\end{figure}
	
	The task aware accuracy and forgetting scores on these are shown in Figure \ref{fig:cifar100_ttaw_50_20}. We find the PAD-based losses to consistently outperform other regularization approaches with LwF being the closest tie. Along the direction of plasticity-stability tradeoff (see Figure \ref{fig:cifar100_other_taw}), we observe that: (a) the attentional PAD losses retain better rigidity than their functional counterparts, and (b) the asymmetric variants of PAD losses are more plastic than their symmetric counterparts across these settings. These trends further validate our hypotheses in sections \ref{sec:att_reg} and \ref{sec:asym_reg}, respectively.
	
	\section*{Task Agnostic Results}
	
\begin{figure}[ht!]
		\centering
		\includegraphics[width=1\textwidth]{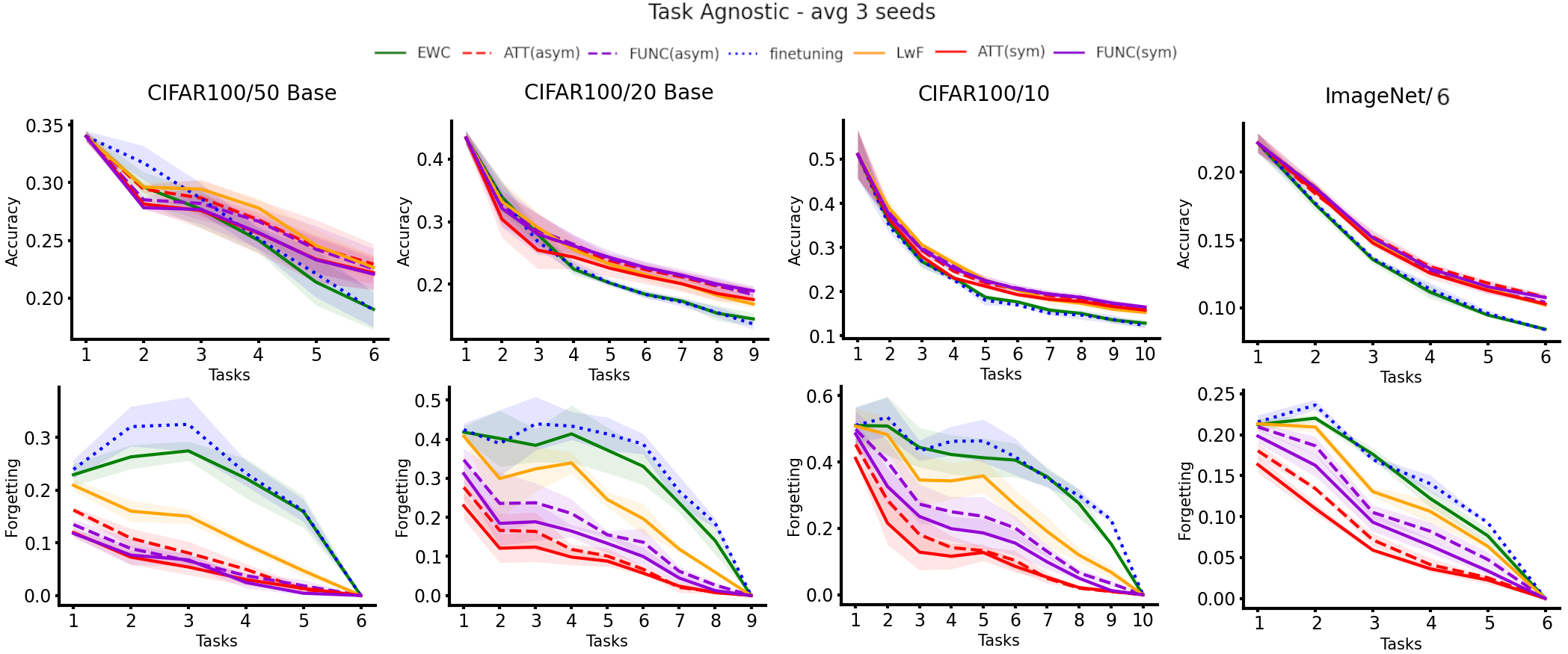}
		\caption{Mean and standard deviation of task-agnostic accuracy and forgetting scores for CIFAR100/10, CIFAR100/20, CIFAR100/50, and ImageNet/6 settings (over 3 random runs). The larger proportion of base task classes (for example, CIFAR100/50) gives rise to higher variations of accuracies and lower variation of forgetting scores across methods -- with the latter indicating the inclination of ViTs towards better generalization and preservation of knowledge.}
		\label{fig:cifar100_tag_settings}
	\end{figure}

	Figure \ref{fig:cifar100_tag_settings} depicts the task-agnostic accuracy and forgetting scores for the settings mentioned in the main section as well as in Section \ref{tag_appendix}. Given the contradictory terms of resource-scarce \textit{exemplar-free} CL and data-hungry ViTs, task-agnostic evaluations can be seen to be particularly challenging. The further avoidance of heavier data augmentations in our training settings can be seen to give rise to two major repercussions across the task-agnostic accuracies: (a) the scores remain consistently low, and (b) the models show smaller yet consistent variations in performances across all settings. 
	
	That said, we find functional PAD losses to be performing the best on all but CIFAR100/50 setting. The larger proportion of base task classes in the latter setting can be seen to be greatly benefiting the learning of LwF (the least parameterized loss term). Further on the note of class proportions, we observe that an equal spread of classes across the tasks can be seen to have a smoothing effect on the variations of scores across different methods.
	
	On the contrary, the CIFAR100/50 setting leads to low variability of task-agnostic forgetting scores across the methods. This can again be attributed to the fact that a very large first task better leverages the generalization capabilities of ViTs thus making them better at avoiding forgetting  over the subsequent incremental steps. This further adds to our reasoning regarding the natural resilience of ViTs to incremental learning settings. When compared across methods, the attentional variants of PAD losses can be seen to display the least amount of forgetting followed by their functional counterparts.
	
\section{Simpler is Better: off-the-shelf Continual Learning through Pretrained Backbones}
In this section we propose a simple baseline for continual learning that leverages pretrained backbones. The approach devised is fast, since requires \textit{no parameters updates} and has \textit{minimal memory requirements} (order of KBytes). By providing such a simple baseline, and achieving strong performance on all the major benchmarks used in literature, we follow the concerns raised in Section \ref{wrk:smaller_is_better} on the simplicity of the benchmarks used. Secondly, we show that pretraining cause the network to generalize at a point where the incremental learning of new tasks is very simple.

In particular, the "training" phase reorders data and exploit the power of pretrained models to compute a class prototype and fill a memory bank. At inference time we match the closest prototype through a knn-like approach, providing us the prediction. We will see how this naive solution can act as an off-the-shelf continual learning system. In order to better consolidate our results, and merge the above two works, we use the devised pipeline with CNN and Vision Transformers. We will discover that thew latter have the ability to produce features of higher quality. As a side note we discuss some extension to the unsupervised realm.

In a nutshell, this simple pipeline raises the same questions raised by previous works such as \cite{gdumb} on the effective progresses made by the CL community especially in the dataset considered and the usage of pretrained models. 

\label{wrk:off-the-shelf}

Until now, the CL community mainly focused in the analysis of catastrophic forgetting in Convolutional Neural Networks (CNN) models. But, as can be seen by some recent works, Vision Transformers (ViT) are asserting themselves as a valuable alternative to CNNs for computer vision tasks, sometimes, achieving better performances with respect to CNNs \cite{Chen2021WhenVT}. The power of ViTs lies in their less inductive bias \cite{DBLP:journals/corr/abs-2106-13122} and in their subsequent better generalization ability. Thanks to this ability ViTs are naturally inclined continual learners, as pointed in Section \ref{wrk:cl_vit}.

\begin{figure}[t]
	\begin{center}
		\centerline{\includegraphics[width=0.8\textwidth]{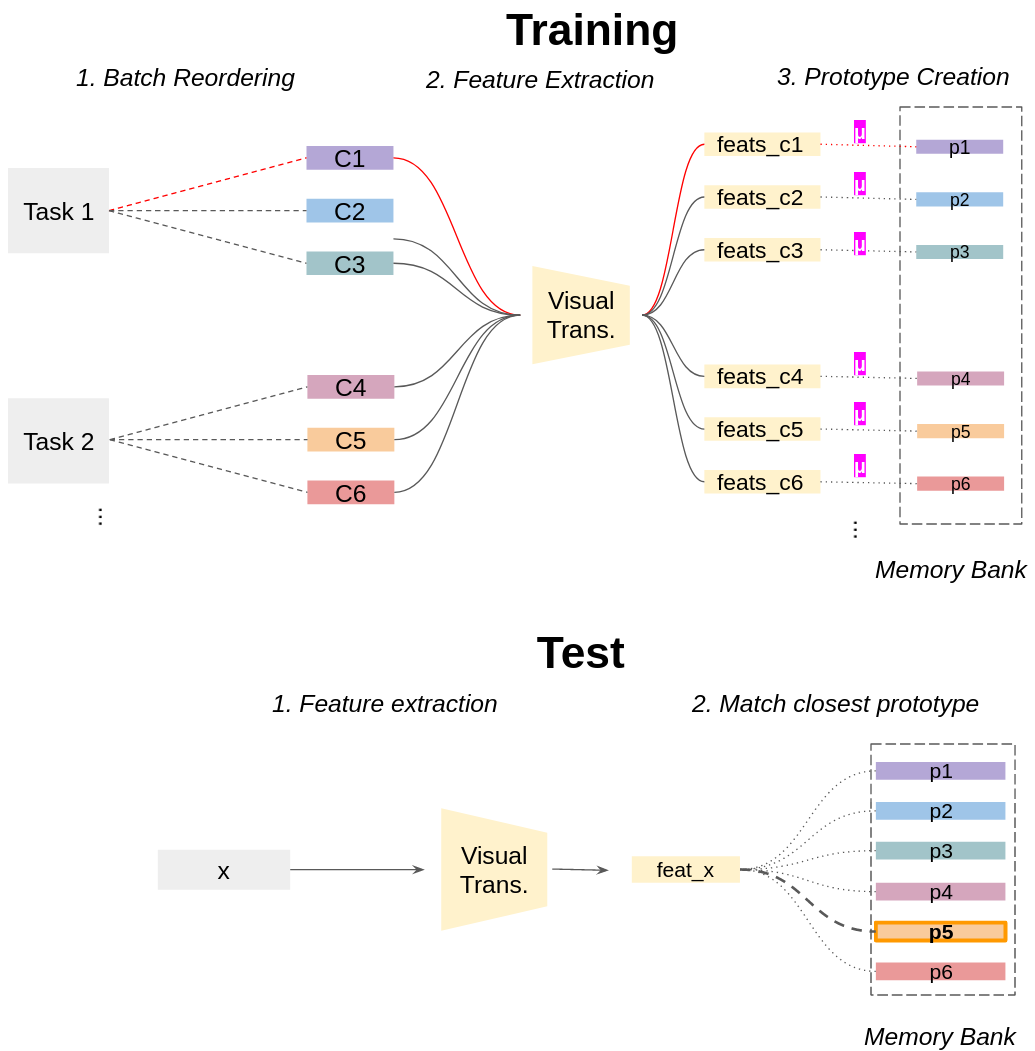}}
		\caption{Depiction of our simple baseline. Our pipeline does not perform parameters updates and consumes few KBytes as memory bank.}
		\label{fig:off-the-shelf}
	\end{center}
\end{figure}

In transformer literature, the usage of \textit{pretrained backbones} is becoming a must, in fact, training such systems requires extensive amount of data and careful hyperparameters optimization. Using pretrained backbones is common also in Computer Vision communities where CNNs are the main player. In CL literature, the pretraining is frequent, but not constant. It is typically carried on half of the analyzed dataset or through a big initial task that has the objective of facilitating the learning of low level features. The very best results, however, have been achieved when we do not skip pretraining. This can be confirmed by the CVPR 2020 Continual Learning Challenge summary report \cite{clworkshop2020}, where the authors noted that \textit{all the methods} proposed solutions leveraging pretrained backbones.

On top of that, simple baselines sometimes provide better results with respect to overly engineered CL solutions, GDumb \cite{gdumb} is such an example. In the work, the authors showed superior performance against several methods at the state-of-the-art through a system composed just by a memory random sampler and a simple learner (CNN or MLP). From a practical point of view, these methods often constitute a simple, clear, fast, intuitive and efficient solution. 

Following these lines, we explore a knn-like method to perform \textit{off-the-shelf online continual learning} leveraging the power of pretrained vision transformers. Our system constitutes a simple and memory-friendly architecture requiring zero parameters updates. Being our work one of the first using ViTs in CL, we propose a robust baseline for future works and provide an extensive comparison against CNNs.

In brevity, the contributions are the following:

\begin{itemize}
	\item We devise a \textit{simple pipeline} composed by a pretrained feature extractor and an incremental prototype bank. The latter is updated as new data is experienced. The overall cost of the method is in the storage of a pretrained backbone and few Kbytes for the memory bank.
	\item We devise a \textit{baseline for future CL methodologies} that will exploit pretrained Vision Transformers or Resnets. The baseline is fast and does not require any parameter update, yet achieving robust results in 200 lines of Python, unlocking reproducibility too.
	\item We provide a \textit{comparison for our pipeline between Resnets and Visual Transformers}. We discover that Vision Transformers produce more discriminative features, appealing also for the CL setting.
	\item In light of such results, \textit{we arise the same questions, as GDumb \cite{gdumb} does, in the progresses made by the CL community so far specifically in the quality of the datasets and in the usage of pretrained backbones.}
\end{itemize}

\begin{algorithm}
	\caption{Off-the-shelf CL. ``Training''}\label{alg:ots-tr}
\begin{algorithmic}[H]
	\REQUIRE $t_i, \phi, \mathcal{M}$
	\FOR{$t_i \in \mathcal{T}$}
	\STATE $\mathcal{G} = \texttt{GroupByClass}(t_i)$
	\FOR{$g \in \mathcal{G}$}
	\STATE $f = \phi(g)$ \hfill Extract features
	\STATE $p = \mu(f)$  \hfill Compute mean feature
	\STATE $\mathcal{M} \gets p$ \hfill Store prototype in memory
	\ENDFOR
	\ENDFOR
	\STATE \RETURN $\mathcal{M}$
\end{algorithmic}
\end{algorithm}

\subsection*{Related Works}

Only recently few works considered self-attention models in continual learning. Li et al. \cite{li2022technical} proposed a framework for object detection exploiting Swin Transformer \cite{Liu2021SwinTH} as pretrained backbone for a CascadeRCNN detector, the authors show that the extracted features generalize better to unseen domains hence achieving lesser forgetting rates compared
to ResNet50 \cite{resnet18} backbones. This also follows the conclusions made by Paul and Chen \cite{paul2021vision} on the fact that vision transformers are more robust learners with respect to CNNs. 

Several methods in CL use pretrained backbones as feature extractors such as in Hayes et al \cite{lda} or \cite{DBLP:conf/cvpr/AljundiKT19, 9206766} and sometimes the pretraining is carried on half (or a big portion) the dataset considered, as in PODNet \cite{podnet} or in Yu et al. \cite{Yu2021ImprovingVT}. For a more complete review on CL methodologies we point out these recent surveys \cite{review0, review1, review2}.

A similar study on pretraining for CL has been conducted by Mehta et al. \cite{Mehta2021AnEI}. In particular, they study the impact on catastrophic forgetting that a linear layer might accuse while using a pretrained backbone. Their study focuses only on Resnet18 for vision tasks, but they also include NLP tasks.

\subsection*{Method}
\paragraph{Setting}
Continual Learning characterizes the learning by introducing the notion of subsequent tasks. In particular, the learning happens in an incremental fashion, that is, the model incrementally experiences different training sessions as time advances. Practically, a learning dataset is split in chunks where each split is considered an incremental task containing data. CL being a relatively new field, the community is still converging to a common setting notation, but we focus on an online, task-agnostic NC-type scenario. Tat is, the model forwards a pattern just once and does not have the task label at test time. As a more fine grained specific we follow \cite{core50} categorization and use a NC-type scenario where each task contains a disjoint group of classes.

More formally, given a dataset $\mathcal{D}$ and a set of $n$ disjoint tasks $\mathcal{T}$ that will be experienced sequentially:
\begin{equation}
	\mathcal{T} = \left[t_1 , t_2 , \dots , t_n \right ]
\end{equation}

each task $t_i = (C_i , D_i )$ represented by a set of classes $C t = c_1^t , c_2^t \dots , c_{n^t}^t$ and training data $D_t$ (images). We assume that the classes of each task do not overlap i.e. $C^i \bigcap C^j = \emptyset$ if $i \neq j$

\paragraph{``Training'' Phase}
In the training phase, given a task $t_i \in \mathcal{T}$, a feature extractor $\phi$ and a memory bank as a dictionary $\mathcal{M}$, the procedure does the following:

\begin{enumerate}
	\item First it performs \textit{batch reordering}, that is, it groups the images of a given task by their class
	\item After grouping, it forwards each new subset to the feature extractor $\phi$
	\item Given the feature representations of a group, it computes the mean of the features to create a \textit{class prototype}
	\item Updates the memory bank $\mathcal{M}$ by storing the each computed prototype 
\end{enumerate}

At the end of the training procedure for a given task $t_i$, we would have a representative prototype vector \textit{for each class} contained in $t_i$. As we said, the prototype vector is computed as the mean feature representation of the patterns of the same class. A depiction of the ``training'' phase is reported in Figure~\ref{fig:off-the-shelf}, we also provide a pseudocode in Algorithm \ref{alg:ots-tr}. 
We also point out that there is not formal ``training'' of the network, in fact \textbf{we do not perform any parameter update}, we simply exploit the pretrained models and construct a knn-like memory system. 
\begin{table*}[ht!]
	\centering
	\scriptsize
	\begin{tabular}{@{}rr|c|ccccc@{}}
\toprule
\textbf{\begin{tabular}[c]{@{}r@{}}Memory\\ KiB class\end{tabular}} & \textbf{Params} & \textbf{Model}     & \textbf{CIFAR100} & \textbf{CIFAR10} & \textbf{Core50} & \textbf{\begin{tabular}[c]{@{}c@{}}Oxford\\ Flowers102\end{tabular} } & \textbf{\begin{tabular}[c]{@{}c@{}}Tiny\\ ImgNet200\end{tabular} } \\ \midrule
2 KiB                                                               & 11.7M           & \textit{resnet18}  & 0.53              & 0.76             & 0.72            & 0.73                                                                 & 0.55                                                              \\
2 KiB                                                               & \textit{21.8M}  & \textit{resnet34}  & 0.55              & 0.81             & 0.74            & 0.67                                                                 & 0.62                                                              \\
8 KiB                                                               & \textit{25.5M}  & \textit{resnet50}  & 0.59              & 0.80             & 0.71            & 0.70                                                                 & 0.63                                                              \\
8 KiB                                                               & \textit{60.1M}  & \textit{resnet152} & 0.67              & 0.89             & 0.72            & 0.66                                                                 & 0.76                                                              \\ \midrule
0.75 KiB                                                            & \textit{5.6M}   & \textit{ViT-T/16}  & 0.36              & 0.63             & 0.49            & 0.54                                                                 & 0.24                                                              \\
3 KiB                                                               & \textit{86.4M}  & \textit{ViT-B/16}  & 0.64              & 0.87             & 0.74            & \textbf{0.95}                                                        & 0.63                                                              \\
0.75 KiB                                                            & \textit{5.6M}   & \textit{DeiT-T/16} & 0.57              & 0.80             & 0.73            & 0.68                                                                 & 0.64                                                              \\
3 KiB                                                               & \textit{86.4M}  & \textit{DeiT-B/16} & \textbf{0.68}     & \textbf{0.90}    & \textbf{0.80}   & 0.74                                                                 & \textbf{0.79}                                                     \\ \cmidrule(lr){3-3}
\end{tabular}

	\caption{Off-the-shelf accuracy performance on different dataset benchmarks, we both analyzed a CNN model and a ViT pretrained models.}
	\label{tbl:off-the-shelf}
\end{table*}

\paragraph{Test Phase}
After completing the training phase for a task $t_i$ the memory bank $\mathcal{M}$ will be populated by the prototypes of the classes encoundered so far. During this test phase, we simply use a \textit{knn-like approach}. Given an image $x$, the updated memory bank $\mathcal{M}$ and the feature extractor $\phi$ we devise the test phase as follows:

\begin{enumerate}
	\item Forward the test image $x$ to the feature extractor $\phi$
	\item Compute a distance between the feature representation of the image and all prototypes contained in $\mathcal{M}$
	\item We match the prototype with minimum distance and return its class
\end{enumerate}

In a nutshell, we perform k-nn with k=1 over the feature representation of an image, matching the class of the closes prototype in the bank. If the class selected is the same of the test example we would have a hit, a miss otherwise. Figure \ref{fig:off-the-shelf} reports a visual depiction of the test procedure. As distance we use a simple $l^2$, but several tests have been made with cosine similarity. Although the results with the cosine similarity are better, we opt for the $l^2$ since provides the best speedup in the implementation through Pytorch.

\subsection*{Experiments} 
\label{sec:off-the-shelf_exp}
%We note that the generalization capabilities of a network directly impact incremental learning performances as can be seen in the visual example provided in Section \ref{sec:cl_visual_example}. Indeed, if we have network which exhibits great generalization capabilities, solving subsequent tasks is easier. 

It is suspected that Visual Transformers generalize better with respect to CNN models. To this end, we compare CNNs models and ViTs models as feature extractors. We selected four CNN models to compare against four attention-based models. In particular, we selected DeiT-Base/15, DeiT-Tiny/15 \cite{deit}, ViT-Base/16 and ViT-Tiny/16 \cite{vit} as visual transformers. While we opted for Resnet18/34/50/152 \cite{resnet18} as CNN models. We used the \texttt{timm} \cite{timm} library to fetch the pretrained models where all the models have been trained on ImageNet \cite{imagenet} and the \texttt{continuum} \cite{continuum} library to create the incremental setting for 5 datasets, namely CIFAR10/100, Core50, OxfordFlowers102 and TinyImageNet200.

In all dataset benchmarks, we upscaled the images to $224 \times 224$ pixels in order to accommodate visual transformers which needs such imput dimension. We apply such transformation to resnet data too for a fair comparison. In order to match the closes prototype at test time, we used $l^2$ as preferred measure. 

The main results are reported in Table \ref{tbl:off-the-shelf}. The pipeline is extremely simple, yet it achieves \textit{impressive performance} as an off-the-shelf method, at cost of a very small overhead to store the prototype memory. In fact, at the end of the training phase, the memory bank translates only into \textbf{few KBytes} of storage. Although this preliminary work only consider task-agnostic setting, we remind that if at test time we are given the task label of the data, we can recast the method to work in task-aware setting. In this case, performing the test phase would be easier since the comparison of the test data will be carried only on a subset of the prototypes. On the same line, one can see that in Table \ref{tbl:off-the-shelf} we do not report each dataset task split. In fact, our method works for \textbf{any dataset split} since it just need any partition of the datasets that respect a NC protocol i.e. as long as tasks are formed by images that can be grouped in classes. We can also appreciate that transformer architectures work best in all benchmarks, suggesting direct \textit{superior generalization capabilities} with respect to CNNs or, at least, more discriminative features.
\begin{figure*}[t]
	\begin{center}
		\centerline{\includegraphics[width=1\textwidth]{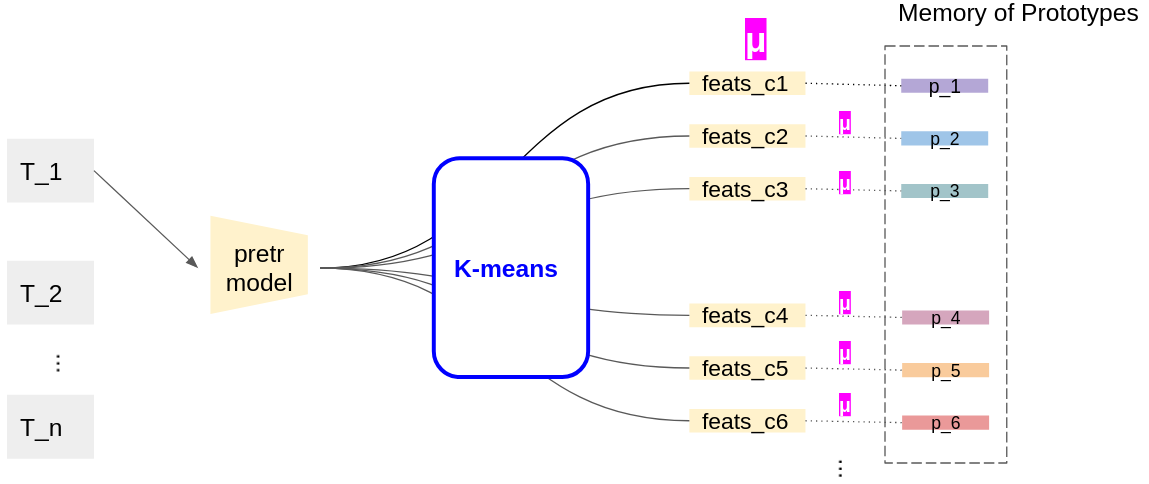}}
		\caption{Direct off-the-shelf extension of the baseline proposed to tackle unspervised continual learning.}
		\label{fig:off-the-shelf_unsup}
	\end{center}
\end{figure*}

%\begin{figure*}[t]
%	\begin{center}
%		\centerline{\includegraphics[width=1\textwidth]{images/unsup_cl.png}}
%		\caption{Direct off-the-shelf extension of the baseline proposed to tackle unspervised continual learning.}
%		\label{fig:off-the-shelf_unsup}
%	\end{center}
%\end{figure*}
\subsection*{Discussion} 
In light of these results, we think that this work may be extended to be considered as a baseline to assess the performance continual learning methodologies using pretrained networks as feature extractors. In particular, a thorough investigation should be carried by substituting the k-nn approach with a linear classifier, this would allow also a better comparison between resnets and visual transformers. However, we think that these preliminary results are of interest to the Vision Transformer and CL research community.

We then raise some concerns with respect to the procedure and the benchmarks used to assess new CL methodologies. As we can see, through a pretrained model, we can achieve impressive results with respect to the current CL state-of-the-art \cite{review0, review1, review2}. This point have been also raised by GDumb \cite{gdumb} where the authors questioned the progresses by providing a very simple baseline.

Moreover, we can further extend this simple pipeline to be used in \textit{unsupervised continual learning}. Actually, the extension is straightforward. In an unsupervised scenario the batch reordering step cannot be performed since we are not allowed to know each data class label. To cope with this lack of information one can substitute the step with any clustering algorithm such as K-means (we tried it but with no luck) or a more sophisticated approach such as autoencoders, self-organizing maps etc.. The test phase of the unsupervised extension would be analogous to the supervised counterpart.

\subsection*{Conclusion}
In this short ex[erimental segment we proposed a baseline for continual learning methodologies that exploit pretrained  Vision Transformers and Resnets. We tackle online NC-type class-incremental learning scenario, the most common one, even though, our pipeline can be extended to different scenarios. Our off-the-shelf method is conceptually simple yet gives strong results and can be implemented in 200 lines of Python therefore enhancing reproducibility. To assess the performance of different backbones our pipeline we compared Resnets models against Vision Transformers feature extractors pretrained on the same dataset, and show that vision transformers provide more powerful features. This suggests that Vision Transformers ability to encode knowledge is is broader. Then we raise some questions about CL research progress and note that with a pretrained model and a simple pipeline one can achieve strong results and, therefore, new methodologies should drop the usage of pretrained backbones when testing on such dataset benchmarks.

\section{Unsupervised Semantic Discovery through Visual Patterns detection}
So far, we directly investigated the impact of performance by altering structural and data properties of object recognition frameworks. If we step back a bit and consider a more broader vision about continual learning, we understand that, in order to adapt to a changing environment, an artificial agent should manifest also the ability to continuously discover new patterns, in our case visual patterns.

We propose a smart pipeline that it is able to discover repetitive patterns in an image, by means of a threshold parameter. That is, if we alter this specific parameter, we are able to discover new semantic levels in a scene. This work goes a bit in another direction from the \textit{dissection} of current continual learning methodologies treated in this thesis. Instead, it is a step towards the ability to build a system able to incrementally explore. 

To this end, we propose a new fast fully unsupervised method to discover semantic patterns. Our algorithm is able to hierarchically find visual categories and produce a segmentation mask. Through the modeling of what is a visual pattern in an image, we introduce the notion of “semantic levels” and devise a conceptual framework along with measures and a dedicated benchmark dataset for future comparisons. Our algorithm is composed by two phases. A filtering phase, which selects semantical hotsposts by means of an accumulator space, then a clustering phase which propagates the semantic properties of the hotspots on a superpixels basis. We provide both qualitative and quantitative experimental validation, achieving optimal results.

\label{wrk:unsup}
\label{sec:introduction}

While the vast majority of supervised object detection and segmentation approaches leverage rich datasets with semantically labelled categories, unsupervised methods cannot rely on such a luxury.
Indeed they are expected to infer from the image content itself what is a relevant object and which are its boundaries. This is a daunting task, as relevance is totally domain-specific and also highly subjective, especially when taking in account human judgement, which exploits a lot of out-of-band information that cannot be found in the sheer image data.

As a matter of fact, little effort have been put to investigate unsupervised automatic approaches to detect and segment semantically relevant objects without any additional information than the image or any \textit{a priori} knowledge of the context. This is due to the fact that a unique definition of what is a relevant object (or, how we prefer to call it, a \emph{visual category}) does not actually exist.

This is especially true if we are seeking to set a formal definition that can be adopted across all the domains in a consistent manner with respect to human judgement.  

Within this section, we try to address this problem by considering a visual category each pattern which appearance is consistent enough across the image. In other words, we consider something to be a relevant object if it appears more than once, exhibiting consistent visual features in different parts of the scene.

From a cognitive and perceptual point of view this makes a lot of sense. In fact, it is easy to observe that if a human is presented with images representing several different but recurring objects, even in a cluttered scene, he does not need to know what the objects actually are representing in order to be able to assign semantically-consistent labels to each of them. He would even be able to label each pixel, defining the boundaries of the objects.

As an example, if someone takes a look at a large bin of different (but to some extent repeated) mechanical parts he never saw before, he is still able to tell one part from the other by exploiting their coherent visual and structural appearance.
This ability is also preserved with slight changes in scale, orientation or partial occlusion of the objects.

Since this automatic assignment to a visual category of recurrent object is both well-defined and quite natural in humans, it is a very good candidate as a rule for automatically detecting relevant objects in an unsupervised manner that has good chances of being coherent with human judgement applied to the same image. 

To be fair, we must also underline the fact that, in order to define the boundaries of a visual category and thus obtain a meaningful segmentation, also the level of detail must be taken into account.
\begin{figure}[t]
	\centering
	\includegraphics[width=0.8\linewidth]{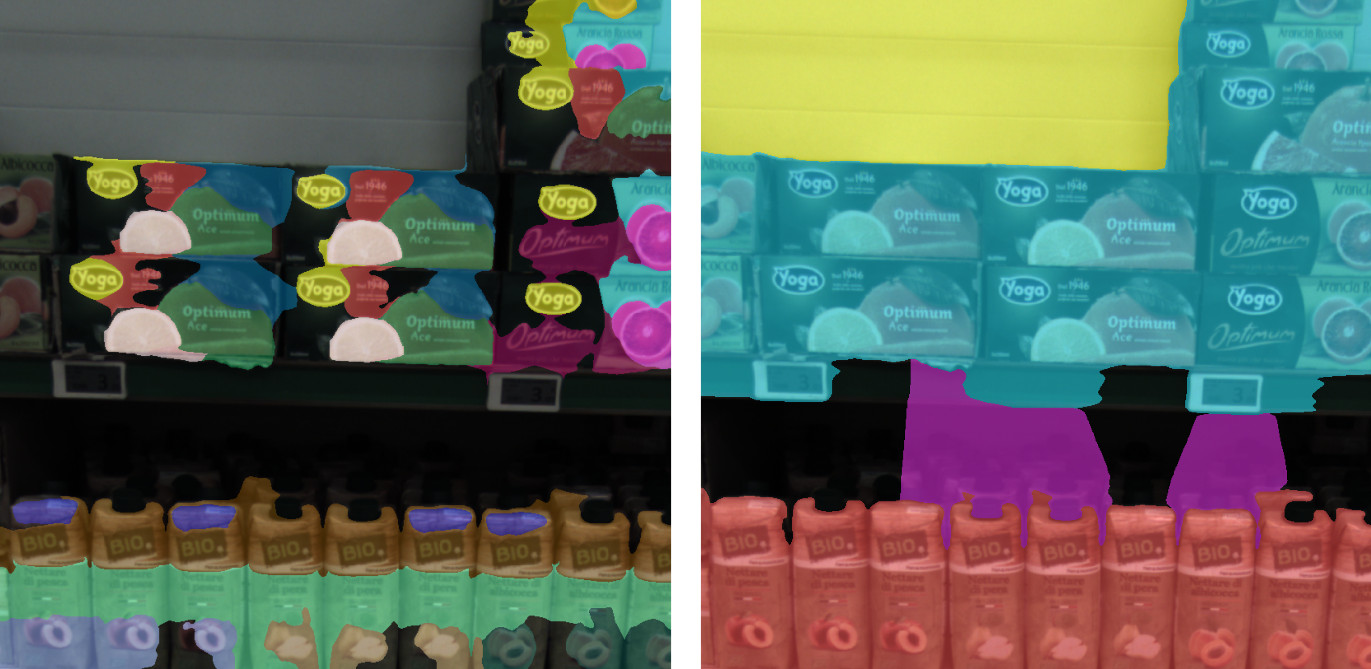}
	\caption{A real world example of unsupervised segmentation of a grocery shelf. Our method can automatically discover both low-level coherent patterns (brands, flavor images and logos) and high-level compound objects (multi-packs and bricks) by controlling the semantical level of the detection and segmentation process.}
	\label{fig:frontpage}
\end{figure}
As an example, if we present to a human an image of a crowded road captured from a side, and we ask him to segment visual categories according to recurrent patterns, we could get slightly different results from different people depending on their attention to details. Some people will segment cars and trees. Other could consider the car body to be a different object from the wheels ad branches from the tree trunk. The most picky could even separate tires from wheel rims and segment out each single leaf. In practice semantic consistency can happen at different scale when dealing with compound objects presenting themselves internal self repetitions or made up of single parts that are also present in other objects.

To address this aspect we also have to design a proper strategy to perform visual category detection and interpretation at a particular scale, according to the level of detail we want to express during the segmentation process. We define this level of detail as \emph{semantical level}. Semantical levels, of course, do not map directly on specific high level concepts, such as whole objects, large parts or minute components. Rather the semantic level will act as a coarse degree of granularity of the segmentation process that will result in a hierarchical split of segments as it changes.

These two definitions of \emph{visual categories} and \emph{semantical levels}, that will be developed throughout the remainder of the work, are the two key concepts driving our novel segmentation method.

The ability of our approach to leverage repetitions to capture the internal representation in the real world and then extrapolates visual categories at a specific semantical level is actually achieved through the combination of a couple of standard techniques, slightly modified for the specific task, and of a few key steps specifically crafted to make the process work in a consistent way with respect to the cognitive process adopted by humans. This happens, for instance, by seeking for highly relevant repetitive structural patterns, called \emph{semantical hotspots}, characterized by a novel feature descriptor, called \emph{splash}. We do this through a scale-invariant method and with no continuous geometrical constraints on the visual pattern disposition. 

We also do not constrain ourselves to find only one visual pattern, which is another very common assumption with other approaches in literature. Rather our technique is designed from the start to be able to detect more patterns at once, being able to assign to each of them a different visual category label, corresponding to a different real world object or object part, according to the selected semantical level.

Overall, with this study, we are offering to the community the following contributions: 

\begin{itemize}
    \item A new pipeline, including the definition of a specially crafted feature descriptor, to capture semantical categories with the ability to hierarchically span over semantical levels;
    \item A specially crafted conceptual framework to evaluate unsupervised semantic-driven segmentation methods through the introduction of the semantical levels notion along with a new metric;
    \item A new dataset consisting of a few hundredths labelled images that can be used as a benchmark for visual repetition detection in general. 
\end{itemize}

The remainder of the section is organized as follows. Section \ref{sec:relatedworks} describes the related works with respect to feature extraction and automatic visual patterns detection. Section \ref{sec:methoddescription} introduces our method, giving details on the overall pipeline and on the implementation details. Section \ref{sec:experimental} presents an experimental evaluation and comparison with similar approaches. Finally, the conclusions are found in Section \ref{sec:conclusions}. 

Code, dataset and notebooks used in this study will be made available for public use.

%In supervised methods, features and mere sintactical properties do not hold any valuable knowledge alone, they acquire significance when hardwired inside the model through the training procedure. 

%What makes our approach particularly appealing is the ability to span over semantical levels, this allows us to get closer on how humans percieve world, in attention-based semantical levls vaben ti ga capio

%%%%%%%%%%%%%%%%%%%%%%%%%%%%%%%%%%%%%%%%%%%%%%%%%%%%%%%%%%%%%%%%%%%%%%%%%%%%%%%%%%5
%                            RELATED WORKS
%%%%%%%%%%%%%%%%%%%%%%%%%%%%%%%%%%%%%%%%%%%%%%%%%%%%%%%%%%%%%%%%%%%%%%%%%%%%%%%%%%5

\section*{Related Works} 
\label{sec:relatedworks}

Several works have been proposed to tackle visual pattern discovery and detection. While the paper by Leung and Malik \citep{DBLP:conf/eccv/LeungM96} could be considered seminal, many other works build on their basic approach, working by detecting contiguous structures of similar patches by knowing the window size enclosing the distinctive pattern. 

One common procedure in order to describe what a pattern is, consists to first extract descriptive features such as SIFT to perform a clustering in the feature space and then model the group disposition over the image by exploiting geometrical constraints, as in \citep{DBLP:conf/cvpr/PrittsCM14} and \citep{DBLP:conf/cvpr/ChumM10}, or by relying only on appearance, as in \citep{DBLP:conf/icpr/DoubekMPC10,DBLP:conf/cvpr/LiuL13,DBLP:journals/pami/ToriiSOP15}.

The geometrical modeling of the repetitions usually is done by fitting a planar 2-D lattice, or a deformation of it \citep{DBLP:journals/pami/ParkBCL09}, through RANSAC procedures as in \citep{DBLP:conf/bmvc/SchaffalitzkyZ98} \citep{DBLP:conf/cvpr/PrittsCM14} or even by exploiting the mathematical theory of crystallographic groups as in \citep{DBLP:journals/pami/LiuCT03}. Shechtman and Irani  \citep{DBLP:conf/cvpr/ShechtmanI07}, also exploited an active learning environment to detect visual patterns in a semi-supervised fashion. For example Cheng et al. \citep{DBLP:journals/tog/ChengZMHH10} use input scribbles performed by a human to guide detection and extraction of such repeated elements, while Huberman  and Fattal \citep{DBLP:conf/cvpr/HubermanF16} ask the user to detect an object instance and then the detection is performed by exploiting correlation of patches near the input area.

Recently, as a result of the new wave of AI-driven Computer Vision, a number of Deep Leaning based approaches emerged, in particular Lettry et al. \citep{DBLP:conf/wacv/LettryPVG17} argued that filter activation in a model such as AlexNet can be exploited in order to find regions of repeated elements over the image, thanks to the fact that filters over different layers show regularity in the activations when convolved with the repeated elements of the image. On top of the latter work, Rodríguez-Pardo et al. \citep{DBLP:journals/cg/Rodriguez-Pardo19} proposed a modification to perform the texture synthesis step. 

A brief survey of visual pattern discovery in both video and image data, up to 2013, is given by Wang et al. \citep{DBLP:journals/widm/WangZY14}, unfortunately after that it seems that the computer vision community lost interest in this challenging problem. We point out that all the aforementioned methods look for \textit{only one} particular visual repetition except for \citep{DBLP:conf/cvpr/LiuL13} that can be considered the most direct competitor and the main benchmark against which to compare our results.

\section*{Method Description}
\label{sec:methoddescription}

%%%%%%%%%%%%%%%%%%%%%%%%%%%%%%%%%%%%%%%%%%%%%%%%%%%%%%%%%%%%%%%%%%%%%%%%%%%%%%%%%%5
\subsection*{Features Localization and Extraction}
\label{sec:featureslocalizationextraction}
%%%%%%%%%%%%%%%%%%%%%%%%%%%%%%%%%%%%%%%%%%%%%%%%%%%%%%%%%%%%%%%%%%%%%%%%%%%%%%%%%%5

% Consider an RGB image $\mathcal{I} \in [0,1]^{n\times m \times 3}$,
We observe that any visual pattern is delimited by its contours. The first step of our algorithm, in fact, consists in the extraction of a set $\mathcal{C}$ of contour \emph{keypoints} indicating a position $\vec{c}_{j}$ in the image. To extract keypoints, we opted for the Canny algorithm, for its simplicity and efficiency, although more recent and better edge extractor could be used \citep{liu2019richer} to have a better overall procedure.

A descriptor $d_{j}$ is then computed for each selected $\vec{c}_{j} \in \mathcal{C}$ thus obtaining a \emph{descriptor set} $\mathcal{D}$. In particular, we adopted the DAISY algorithm because of its appealing dense matching properties that nicely fit our scenario. Again, here we can replace this module of the pipeline with something more advanced such as \citep{DBLP:conf/nips/OnoTFY18} at the cost of some computational time.

%%%%%%%%%%%%%%%%%%%%%%%%%%%%%%%%%%%%%%%%%%%%%%%%%%%%%%%%%%%%%%%%%%%%%%%%%%%%%%%%%%5
\subsection*{Semantic Hot Spots Detection}
\label{sec:semantichotspotdetection}
%%%%%%%%%%%%%%%%%%%%%%%%%%%%%%%%%%%%%%%%%%%%%%%%%%%%%%%%%%%%%%%%%%%%%%%%%%%%%%%%%%5

\begin{figure*}[t!]
	\centering
	\includegraphics[width=1\linewidth]{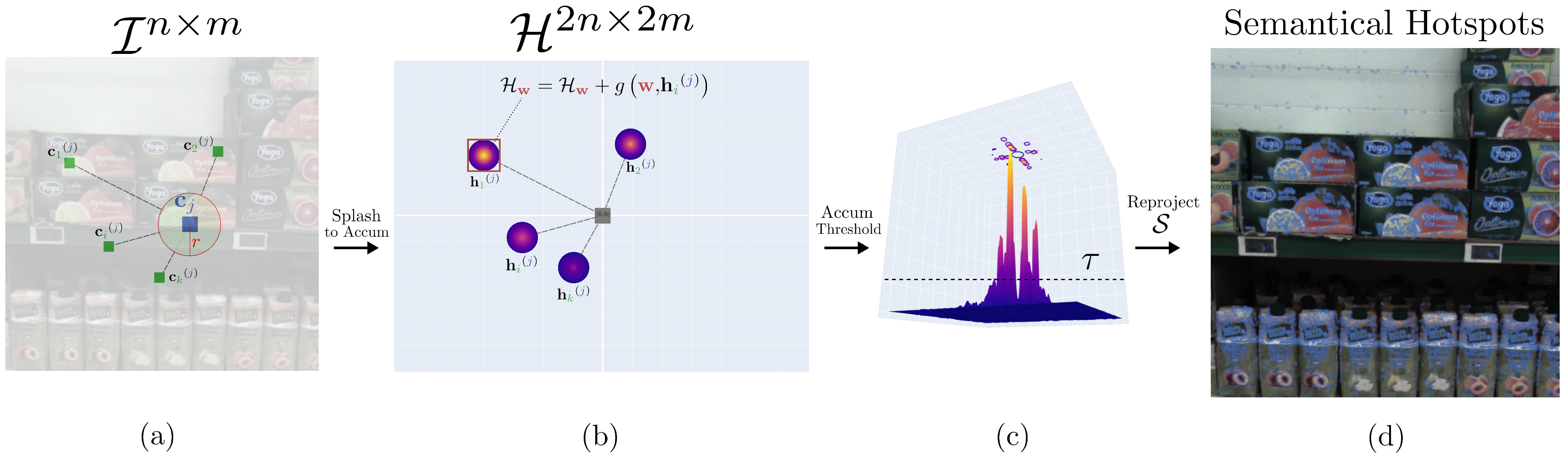}
	\caption{(a) A splash in the image space with center in the keypoint $\vec{c}_j$. (b) $\mathcal{H}$, with the superimposed splash at the center, you can note the different levels of the vote ordered by endpoint importance i.e. descriptor similarity. (c) 3D projection showing the gaussian-like formations and the thresholding procedure of $\mathcal{H}$. (d) Backprojection through the set $\mathcal{S}$.}
	\label{fig:houghaccumulator}
\end{figure*}

In order to detect self-similar patterns in the image we start by associating the $k$ most similar descriptors for each descriptor $\vec{d}_j$. We can visualize this data structure as a star subgraph with $k$ endpoints called \emph{splash} ``centered'' on descriptor $\vec{d}_{j}$. Figure \ref{fig:houghaccumulator} (a) shows one.

Splashes potentially encode repeated patterns in the image and similar patterns are then represented by similar splashes.
The next step consists in separating these splashes from those that encode noise only, this is accomplished through an accumulator space.

In particular, we consider a $2$-D \emph{accumulator space} $\mathcal{H}$ of size double the image. We then superimpose each splash on the space $\mathcal{H}$ and cast $k$ votes as shown in Figure \ref{fig:houghaccumulator} (b). In order to take into account the noise present in the splashes, we adopt a gaussian vote-casting procedure $g(\cdot)$. Similar superimposed splashes contribute to similar locations on the accumulator space, resulting in peak formations (Figure \ref{fig:houghaccumulator} (c)). We summarize the voting procedure as follows:

\begin{equation}
\mathcal{H}_{\vec{w}} = \mathcal{H}_{\vec{w}} + g(\vec{w}, \vec{h}^{(j)}_{i})
\end{equation}
where $\vec{h}^{(j)}_{i}$ is the $i$-th splash endpoint of descriptor $\vec{d}_j$ in accumulator coordinates and $\vec{w}$ is the size of the gaussian vote. We filter all the regions in $\mathcal{H}$ which are above a certain \emph{threshold} $\tau$, to get a set $\mathcal{S}$ of the locations corresponding to the peaks in $\mathcal{H}$. The $\tau$ parameter acts as a coarse filter and is not a critical parameter to the overall pipeline. A sufficient value is to set it to $0.05 \cdot max(\mathcal{H})$. 
Lastly, in order to visualize the semantic hotspots in the image plane we map splash locations between $\mathcal{H}$ and the image plane by means of a \emph{backtracking structure} $\mathcal{V}$.

In summary, the key insight here is that similar visual regions share similar splashes, we discern noisy splashes from representative splashes through an auxiliary structure, namely an accumulator. We then identify and backtrack in the image plane the semantic hotspots that are candidate points part of a visual repetition.

%%%%%%%%%%%%%%%%%%%%%%%%%%%%%%%%%%%%%%%%%%%%%%%%%%%%%%%%%%%%%%%%%%%%%%%%%%%%%%%%%%5
\subsection*{Semantic Categories Definition and Extraction}
\label{sec:graph}
%%%%%%%%%%%%%%%%%%%%%%%%%%%%%%%%%%%%%%%%%%%%%%%%%%%%%%%%%%%%%%%%%%%%%%%%%%%%%%%%%%5

While the first part previously described acts as a filter for noisy keypoints allowing to obtain a good pool of candidates, we now transform the problem of finding visual categories in a problem of dense subgraphs extraction.

We enclose semantic hotspots in superpixels, this extends the semantic significance of such identified points to a broader, but coherent, area. To do so we use the SLIC \citep{DBLP:journals/pami/AchantaSSLFS12} algorithm which is a simple and one of the fastest approaches to extract superpixels as pointed out in this recent survey \citep{DBLP:journals/cviu/StutzHL18}. Then we choose the cardinality of the \emph{superpixels} $\mathcal{P}$ to extract. This is the second and most fundamental parameter that will allow us to span over different semantic levels.

Once the superpixels have been extracted, let $\mathcal{G}$ be an \emph{undirected weighted graph} where each node correspond to a superpixel $p \in \mathcal{P}$. In order to put edges between graph nodes (i.e. two superpixels), we exploit the splashes origin and endpoints. In particular the strength of the connection between two vertices in $\mathcal{G}$ is calculated with the number of splashes endpoints falling between the two in a mutual coherent way. So to put a weight of 1 between two nodes we need exactly 2 splashes endpoints falling with both origin and end point in the two candidate superpixels.

With this construction scheme, the graph has clear dense subraphs formations. Therefore, the last part simply computes a partition of $\mathcal{G}$ where each connected component correspond to a cluster of similar superpixels. In order to achieve such objective we optimize a function that is maximized when we partition the graph to represent so. To this end we define the following \textit{density score} that given $G$ and a set $K$ of connected components captures the optimality of the clustering:

\begin{equation}
s(G, K) = \sum_{k \in K} \mu(k) - \alpha \left | K \right |
\end{equation}
where $\mu(k)$ is a function that computes the average edge weight in a undirected weighted graph.

\begin{algorithm}[t]
	\caption{Semantic categories extraction algorithm}
	\vspace{2mm}
	\begin{algorithmic}
		\REQUIRE $G$ weighted undirected graph
		\STATE $i=0$
		\STATE $s^{*}=-\inf$
		\STATE $K^{*}= \emptyset$
		\WHILE{$G_{i}$ is not fully disconnected}
		\STATE $i = i + 1$
		\STATE Compute $G_{i}$ by corroding each edge with the minimum edge weight
		\STATE Extract the set $K_{i}$ of all connected components in $G_{i}$
		\STATE $s(G_{i}, K_{i}) = \sum_{k \in K_{i}} \mu(k) - \alpha \left | K_{i} \right |$
		\IF{$s(G_{i}, K_{i}) > s^{*}$}
		\STATE $s^{*} = s(G_{i}, K_{i})$
		\STATE $K^{*} = K_{i}$
		\ENDIF
		\ENDWHILE
		\RETURN $s^{*}, K^{*}$ 
	\end{algorithmic}
	\label{alg:graphcorrosion}
\end{algorithm}

The first term, in the score function, assign a high vote if each connected component is dense. While the second term acts as a regulator for the number of connected components. We also added a weighting factor $\alpha$ to better adjust the procedure. As a proxy to maximize this function we devised an \emph{iterative algorithm} reported in Algorithm \ref{alg:graphcorrosion} based on graph corrosion and with temporal complexity of $O(\left | E \right |^{2} + \left | E \right | \left | V \right |)$. At each step the procedure corrupts the graph edges by the minimum edge weight of $G$. For each corroded version of the graph that we call \emph{partition}, we compute $s$ to capture the density. Finally the algorithm selects the corroded graph partition which maximizes the $s$ and subsequently extracts the node groups.

In brevity we first enclose semantic hotspots in superpixels and consider each one as a node of a weighted graph. We then put edges with weight proportional to the number of splashes falling between two superpixels. This results in a graph with clear dense subgraphs formations that correspond to superpixels clusters i.e. \textit{semantic categories}. The semantic categories detection translates in the extraction of dense subgraphs. To this end we devised an iterative algorithm based on graph corrosion where we let the procedure select the corroded graph partition that filters noisy edges and let dense subgraphs emerge. We do so by maximizing score that captures the density of each connected component.

%%%%%%%%%%%%%%%%%%%%%%%%%%%%%%%%%
\section*{Experiments}
\label{sec:experimental}
%%%%%%%%%%%%%%%%%%%%%%%%%%%%%%%%%

\begin{figure}[t!]
	\centering
	\includegraphics[width=\linewidth]{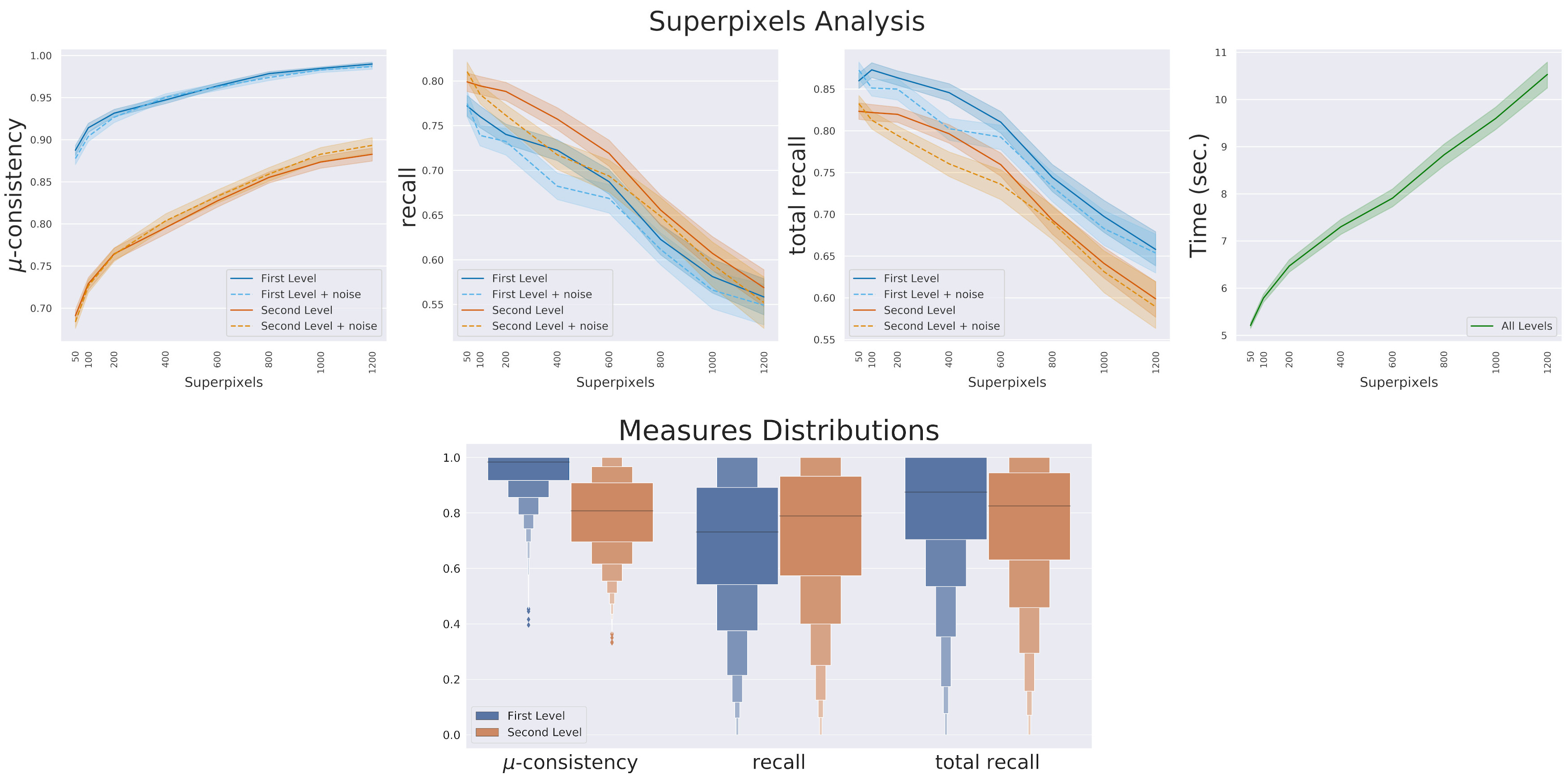}
	\caption{(top) Analysis of measures as the number of superpixels $|\mathcal{P}|$ retrieved varies. The rightmost figure shows the running time of the algorithm.  We repeated the experiments with the noisy version of the dataset but report only the mean since variation is almost equal to the original one. (bottom) Distributions of the measures for the two semantic levels, by varying the two main parameters $r$ and $|\mathcal{P}|$.}
	\label{fig:superpixels}
\end{figure}

\begin{figure}[t]
	\centering
	\includegraphics[width=1\linewidth]{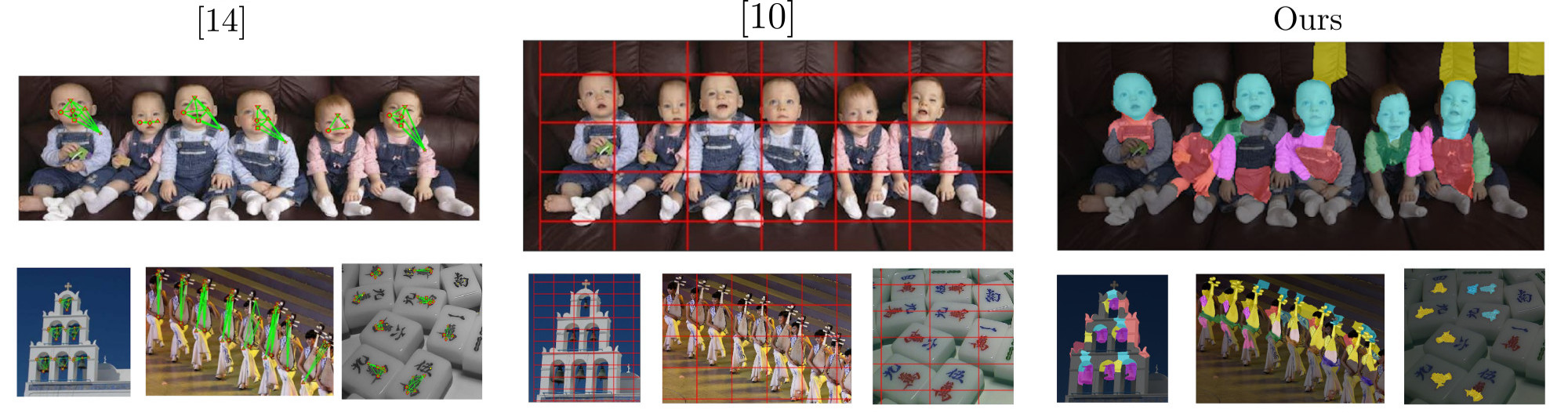}
	\caption{Qualitative comparison between \citep{DBLP:conf/cvpr/LiuL13} [14], \citep{DBLP:conf/wacv/LettryPVG17} [10] and our algorithm. Our method detects and segments more than one pattern and does not constrain itself to a particular geometrical disposition.}
	\label{fig:distributions}
\end{figure}

\subsection*{Dataset}
As we introduced in Section \ref{sec:introduction} one of the aims of this work is to provide a better comparative framework for visual pattern detection. To do so we created a public dataset by taking 104 pictures of store shelves. Each picture has been took with a 5mpx camera with approximatively the same visual conditions. We also rectified the images to eliminate visual distortions. 

We manually segmented and labeled each repeating product in two different semantic levels. In the \textbf{first semantic level} \emph{products made by the same company} share the same label. In the \textbf{second semantic level} visual repetitions consist in the \emph{exact identical products}. In total the dataset is composed by 208 ground truth images, half in the first level and the rest for the second one.

\subsection*{$\mu$-consistency}
We devised a new measure that captures the semantic consistency of a detected pattern that is a proxy of the average precision of detection.

In fact, we want to be sure that all pattern instances fall on similar ground truth objects. First we introduce the concept of semantic consistency for a particular pattern $\vec{p}$. Let $\vec{P}$ be the set of patterns discovered by the algorithm. Each pattern $\vec{p}$ contains several instances $\vec{p}_{i}$. $\vec{L}$ is the set of ground truth categories, each ground truth category $\vec{l}$ contain several objects instances $\vec{l}_{i}$. Let us define $\vec{t}_{p}$ as the vector of ground truth labels touched by all instances of $\vec{p}$. We say that $\vec{p}$ is consistent if all its instances $\vec{p}_{i}, i=0\dots |\vec{p}|$ fall on ground truth regions sharing the same label. In this case $\vec{t}_{p}$ would be uniform and we consider $\vec{p}$ a good detection. The worst scenario is when given a pattern $\vec{p}$ every $\vec{p}_{i}$ falls on objects with different label $\vec{l}$ i.e. all the values in $\vec{t}_{p}$ are different. 

To get an estimate of the overall consistency of the proposed detection, we average the consistency for each $\vec{p} \in \vec{P}$ giving us: 

\begin{equation}
\text{$\mu$-consistency} =  \frac{1}{\left | \vec{P} \right |} \sum_{\vec{p} \in \vec{P}} \frac{\left| \operatorname{mode}\left(\vec{t}_{p}\right)\right|}{\left|\vec{t}_{p}\right|}
\end{equation}

\subsection*{Recall}
The second measure is the classical recall over the objects retrieved by the algorithm. Since our object detector outputs more than one pattern we average the recall for each ground truth label by taking the best fitting pattern.

\begin{equation}
\frac{1}{\left | \vec{L} \right |} \sum_{\vec{l} \in \vec{L}} \operatorname{max}_{\vec{p} \in \vec{P}} \text { recall }(\vec{p}, \vec{l})
\end{equation}

The last measure is the \textbf{total recall}, here we consider a hit if any of the pattern falls in a labeled region. In general we expect this to be higher than the recall.

We report the summary performances in Figure \ref{fig:distributions}. As can be seen the algorithm achieves a very high $\mu$-consistency while still able to retrieve the majority of the ground truth patterns in both levels.

One can observe in Figure \ref{fig:superpixels} an inverse behaviour between recall and consistency as the number of superpixels retrieved grows. This is expected since less superpixels means bigger patterns, therefore it is more likely to retrieve more ground truth patterns. 

In order to study the robustness we repeated the same experiments  with an altered version of our dataset. In particular for each image we applied one of the following corruptions: Additive Gaussian Noise ($scale=0.1*255$), Gaussian Blur  ($\sigma = 3$), Spline Distortions (grid affine), Brightness ($+100$), and Linear Contrast ($1.5$).

\subsection*{Qualitative Validation}
Firstly we begin the comparison by commenting on \citep{DBLP:conf/cvpr/LiuL13}. One can observe that our approach has a significant advantage in terms of how the visual pattern is modeled. While the authors model visual repetitions as geometrical artifacts associating points, we output a higher order representation of the visual pattern. Indeed the capability to provide a segmentation mask of the repeated instance region together the ability to span over different levels unlocks a wider range of use cases and applications.

As qualitative comparison we also added the latest (and only) deep learning based methodology \citep{DBLP:conf/wacv/LettryPVG17} we found. This methodology is only able to find a single instance of visual pattern, namely the most frequent and most significant with respect to the filters weights. This means that the detection strongly depends from the training set of the CNN backbone, while our algorithm is fully unsupervised and data agnostic. 

\subsection*{Quantitative Validation}
We compared quantitatively our method against \citep{DBLP:conf/cvpr/LiuL13} that constitutes, to the best of our knowledge, the only work developed able to detect more than one visual pattern. We recreated the experimental settings of the authors by using the Face dataset \citep{DBLP:journals/cviu/Fei-FeiFP07} as benchmark achieving $1.00$ precision vs. $0.98$ of \citep{DBLP:conf/cvpr/LiuL13} and $0.77$ in recall vs. and $0.63$. We considered a miss on the object retrieval task, if more than 20\% of a pattern total area falls outside from the ground truth. The parameter used were $|\mathcal{C}|=9000$, $k=15$, $r=30$, $\tau=5$, $| \mathcal{P} |=150$. We also fixed the window of the gaussian vote to be $11 \times11$ pixels throughout all the experiments.

%%%%%%%%%%%%%%%%%%%%%
\section*{Conclusions}
\label{sec:conclusions}
%%%%%%%%%%%%%%%%%%%%%

With this study we introduced a fast and unsupervised method addressing the problem of finding semantic categories by detecting consistent visual pattern repetitions at a given scale. The proposed pipeline hierarchically detects self-similar regions represented by a segmentation mask.

As we demonstrated in the experimental evaluation, our approach retrieves more than one pattern and achieves better performances with respect to competitors methods. We also introduce the concept of \emph{semantic levels} endowed with a dedicated dataset and a new metric to provide to other researchers tools to evaluate the consistency of their approaches.

\subsection*{Acknowledgments}
We would like to express our gratitude to Alessandro Torcinovich and Filippo Bergamasco for their suggestions to improve the work. We also thank Mattia Mantoan for his work to produce the dataset labeling.

%% Comparison table
%\begin{table}[t!]
%\centering
%\normalsize
%\begin{tabular}{@{}lllll@{}}
%\toprule
%\multicolumn{5}{c}{Face Dataset %\citep{DBLP:journals/cviu/Fei-FeiFP07}}                                                    \\ \midrule
%                & \multicolumn{2}{l}{Precision} & \multicolumn{2}{l}{Recall}   \\
%GRASP \citep{DBLP:conf/cvpr/LiuL13}   & \multicolumn{2}{l}{0.98}    & \multicolumn{2}{l}{0.63} \\ 
%\textbf{Ours}      & \multicolumn{2}{l}{\textbf{1.00}}    & \multicolumn{2}{l}{\textbf{0.77}}  \\ \bottomrule
%\\
%\end{tabular}
%\caption{\label{tab:grasp}Object-Level precision/recall rates. Parameters used: $|\mathcal{C}|=9000$, $k=15$, $r=30$, $\tau=5$, $| \mathcal{P} |=150$}
%\vspace{-1mm}
%\end{table}

%\begin{table}[h]
%\centering
%\normalsize
%\begin{tabular}{@{}lllll@{}}
%\toprule
%\multicolumn{5}{c}{Face Dataset \citep{DBLP:journals/cviu/Fei-FeiFP07}}                                                    \\ \midrule
%& \multicolumn{2}{l}{\citep{DBLP:conf/cvpr/LiuL13}} & %\multicolumn{2}{l}{\textbf{Ours}}   \\
%Precision (\%)   & \multicolumn{2}{l}{0.98}    & \multicolumn{2}{l}{\textbf{1.00}} \\ 
%Recall (\%)      & \multicolumn{2}{l}{0.63}    & \multicolumn{2}{l}{\textbf{0.77}}  \\ %\bottomrule
%\\
%\end{tabular}
%\caption{\label{tab:grasp}Object-Level precision/recall rates. Parameters used: $|\mathcal{C}|=9000$, $k=15$, $r=30$, $\tau=5$, $| \mathcal{P} |=150$}
%\vspace{-1mm}
%\end{table}
\include{chapters/papers/unsup/related_works}
\include{chapters/papers/unsup/method}
\include{chapters/papers/unsup/experiments}
\include{chapters/papers/unsup/conclusions}

\chapter{Conclusions}
In this thesis, we contributed spanned the dissection of continual learning by providing several structural and data analyses. First we provide a gentle introduction to the topic of continual learning starting by highlighting the difference between natural and artificial models. Among the differences we stress the importance of time, which is an essential component for developing lifelong learning machines. Then, we informally introduce the main challenges that continual learning systems must tackle. In particular, catastrophic forgetting and the stability plasticity dilemma. To better provide an intuition about these topics, we provided a visual example of catastrophic forgetting in an autoencoder model, showing how distributional shifts in the subsequent tasks result in the abrupt damage of past knowledge.

Later, we move on by giving a more formal definition of continual learning settings prominently adopted in literature. We introduced the notions of class-incremental, task-incremental, online/offline learning along with a specification on other common settings in the field. Before moving on the contributions we provided a small literature review on the state-of-the-art by describing the main categories under which continual learning methods have been grouped. 

Finally, we move on the main contributions. First, we introduced a study on the quality/quantity trade-off in rehearsal-based continual learning. Here, we selected one of the most performant baselines, that is GDumb, and analyzed several compression techniques when applied to the replay buffer. We highlighted that the quantity of data is a far more important factor when storing examplars in the replay buffer. We do so by considering different compression schemes with extreme rates. Then, we moved into the second major contribution which considers Visual Transformers in an incremental setting. Here, besides being one of the first works on visual transformers for continual learning, we provided a surgical investigation on regularization methods for ViTs in the challenging setting of rehearsal-free CL. We compared functional, weight and attentional regularizations, with the latter being a regularization in the matrix of the self-attention mechanism. Attentional regularizations provide comparable performance with respect to the other methods. As second contribution we also introduced a loss inspired by a method nowadays in vogue (PODNet) and devised an asymmetric variant. We show that the introduction of the asymmetric variant allows achieving more plasticity to the model when applied to different part of the mechanism of self-attention. Then, we proposed a study on off-the-shelf continual learning exploiting fully pretrained networks and, in particular, we proposed a simple baseline. The baseline is composed by a feature extractor and a knn-like prototype memory. The baseline is crafted to be performant in practical scenarios achieving optimal results with a memory overhead of few KBytes. Moreover we discussed its possible extension to the realm of unsupervised continual learning. We then linked this preliminary discussion with the exploration of visual categories. To do so we introduce another work tackling unsupervised pattern discovery. In fact, the notion of \textit{discovery} is naturally included into the notion of lifelong learning: an agent capable of lifelong learning, surely should possess the ability to autonomously discover new knowledge. We do so by introducing a new unsupervised algorithm to perform unsupervised semantic segmentation at different semantic scales. 

\subsection*{Further Developings}
With the several studies proposed, we want to highlight the directions where it might be more fruitful to investigate further to build better Continual Learning agents. 

\textit{A first warning we raised regards the dataset usage to assess the performance of CL algorithms}. In particular, with Section \ref{wrk:smaller_is_better} we see that extreme levels of buffer data resize still provide good results in rehearsal systems, suggesting that, perhaps, more realistic datasets should be included to devise more useful solutions. This finding is also supported by Section \ref{wrk:off-the-shelf} which shows that tackling these benchmarks with a pretrained backbone is sufficient to overcome quasi-optimally continual learning scenarios on 5 different datasets. This also suggests that pretraining could be a great advantage, in the generalization ability of the model, when building new CL algorithms.

To tackle the aforementioned point, the community can focus more on unsupervised continual learning which is a natural and more challenging problem extension. While keeping the same datasets we can now also leverage pretrained backbones. While being appealing on its own, following this line is also greatly encouraged by the fact that there are virtually no works on such a topic. 

With the study proposed in Section \ref{wrk:cl_vit} we show that ViTs are naturally inclined continual learners. We suspect that the less inductive bias carried by such models might be the key that allows such models to perform better in incremental scenarios. On another side, we see that the results obtained without pretraining have difficulty achieving CNN performances so easily (we can compare the results of Section \ref{wrk:smaller_is_better} and Section \ref{wrk:cl_vit}). This calls for the need to build less data-hungry models in line with the world's fast-paced data generation. Within Section \ref{wrk:cl_vit} we also propose a new way to assess Continual Learning methods. We think that \textit{the community still lacks of a principled way to measure the stability-plasticity trade-off}. With our introduction of the two curves, we proposed an initial tentative to monitor the performance of a system.

Last but not least, with the work of Section \ref{wrk:unsup} we stress that autonomously discovering new patterns should be a core ability of an intelligent system. In fact, if an agent can explore the real world and find hierarchies of knowledge without help, all it has to do to incrementally learn is to store such knowledge in some kind of long-term memory repository which translates into a \textit{compression problem}.

\cleardoublepage

\fancyfoot[LO,RE]{Bibliography}
\bibliography{biblios}
%\cleardoublepage

%\appendix
%\fancyfoot[LO,RE]{Appendix \thechapter}
%\chapter{Appendix Title}
%\input{chapters/appendix}

\end{document}